\documentclass[two-column]{article}

\usepackage[affil-it]{authblk}

\usepackage{sectsty,relsize}
\sectionfont{\Large\selectfont}
\subsectionfont{\relsize{1.2}\normalsize\selectfont}
\subsubsectionfont{\normalsize\selectfont}

\usepackage{geometry}
\usepackage{url}
\usepackage{hyperref}
\usepackage{natbib}
\usepackage{graphicx}
\usepackage[subrefformat=parens]{subcaption}

\usepackage{xifthen}
\usepackage{amsmath}
\usepackage{amsfonts}

\renewcommand{\vec}[1]{{\boldsymbol{#1}}}

\usepackage{color}

\newcommand{\textsub}[1]{\textsc{\tiny #1}} 

\usepackage{pifont}
\makeatletter
\newcommand{\colorlabel}[1]{%
	\global\tag@true%
	\nonumber%
	\refstepcounter{equation}%
	\gdef\df@tag{\maketag@@@{{\color{#1}(\theequation)}}\def\@currentlabel{\theequation}}}
\makeatother

\newcommand{\st}{\text{s.t.}\space\:} 
\newcommand{\bigmid}{\;\ifnum\currentgrouptype=16 \middle\fi|\;} 
\newcommand{\mytilde}[0]{\mathds{\raise.17ex\hbox{$\scriptstyle\sim$}}} 

\DeclareMathOperator{\diff}{\mathrm{d}\!} 
\DeclareMathOperator{\T}{{\mathsf{T}}} 

\DeclareMathOperator{\EV}{\mathbb{E}}
\newcommand{\EVV}[2]{\EV_{#1}\!\left[{#2}\right]}

\DeclareMathOperator{\var}{\mathrm{Var}}

\newcommand{\gaussian}{\mathcal N}
\newcommand{\realspace}{\mathbb R}

\newcommand{\action}{a}
\newcommand{\state}{s}
\newcommand{\context}{c}
\newcommand{\params}{\theta}

\newcommand{\statespace}{\mathcal S}
\newcommand{\actionspace}{\mathcal A}

\newcommand{\dstate}{d_{\textsub{\state}}}
\newcommand{\daction}{d_{\textsub{\action}}}

\providecommand{\rwd}{\mathcal{R}}
\providecommand{\prob}{\mathcal{P}}

\newcommand{\rmodel}[1][]{%
	\ifthenelse{\equal{#1}{}}{\rwd\left(\state,\action\right)}{\rwd\left(\state^{[#1]},\action^{[#1]}\right)}%
}
\newcommand{\rmodelt}[1][]{%
	\ifthenelse{\equal{#1}{}}{\rwd\left(\state_t,\action_t\right)}{\rwd\left(\state_t^{[#1]},\action_t^{[#1]}\right)}%
}

\newcommand{\pmodel}[1][]{%
	\ifthenelse{\equal{#1}{}}{\prob\left(\state'|\state,\action\right)}{\prob\left(\state^{[#1+1]}|\state^{[#1]},\action^{[#1]}\right)}%
}
\newcommand{\pmodelt}[1][]{%
	\ifthenelse{\equal{#1}{}}{\prob\left(\state_{t+1}|\state_t,\action_t\right)}{\prob\left(\state_{t+1}^{[#1]}|\state_t^{[#1]},\action_t^{[#1]}\right)}%
}

\newcommand{\rmodelctx}[1][]{%
	\ifthenelse{\equal{#1}{}}{\rwd\left(\context,\params\right)}{\rwd\left(\context^{[#1]},\params^{[#1]}\right)}%
}
\newcommand{\berror}[1][]{%
	\ifthenelse{\equal{#1}{}}{\delta\left(\state,\action\right)}{\delta_#1\left(\state,\action\right)}%
}

\newcommand{\q}{q}
\usepackage{relsize}

\newcommand{\paramsQ}{\omega}
\newcommand{\tdcoeff}{\eta}
\newcommand{\elicoeff}{\lambda}

\renewcommand{\paramsQ}{\vec\omega}
\renewcommand{\params}{\vec\theta}

\begin{document}

\title{TD-Regularized Actor-Critic Methods}

\author[1]{Simone Parisi\thanks{Corresponding author: \texttt{parisi@ias.tu-darmstadt.de}}}
\author[2]{Voot Tangkaratt}
\author[1,3]{Jan Peters}
\author[2]{Mohammad Emtiyaz Khan}
\date{\small Machine Learning Journal (in press)}

\affil[1]{\small Technische Universit{\"a}t Darmstadt, Hochschulstr. 10, 64289 Darmstadt, Germany}
\affil[2]{RIKEN Center for Advanced Intelligence Project
	\\ 1-4-1 Nihonbashi, Chuo-ku, Tokyo 103-0027, Japan}
\affil[3]{Max Planck Institute for Intelligent Systems, Spemannstr. 41, 72076 T\"ubingen, Germany}

\maketitle

\hrule
\subsubsection*{Abstract}
Actor-critic methods can achieve incredible performance on difficult reinforcement learning problems, but they are also prone to instability. This is partly due to the interaction between the actor and the critic during learning, e.g., an inaccurate step taken by one of them might adversely affect the other and destabilize the learning.
To avoid such issues, we propose to regularize the learning objective of the actor by penalizing the temporal difference (TD) error of the critic. This improves stability by avoiding large steps in the actor update whenever the critic is highly inaccurate.
The resulting method, which we call the \emph{TD-regularized} actor-critic method, is a simple plug-and-play approach to improve stability and overall performance of the actor-critic methods. Evaluations on standard benchmarks confirm this.\\
Source code can be found at \url{https://github.com/sparisi/td-reg}

\bigskip
\noindent\emph{Keywords:} reinforcement learning, actor-critic, temporal difference
\bigskip
\hrule


\section{Introduction}
\label{sec:intro}
\begin{figure}[t]
	\centering
	\includegraphics[width=\linewidth]{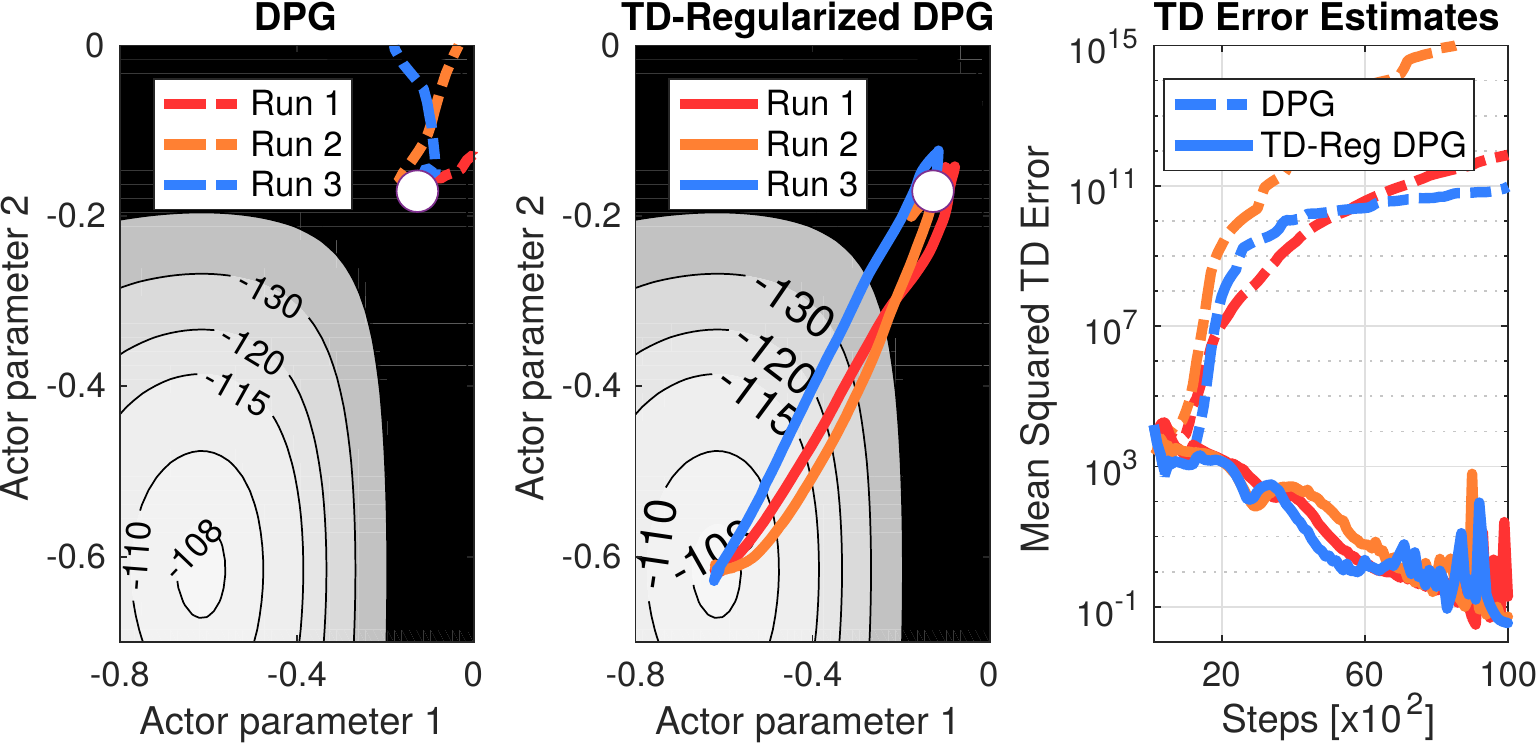}
	\caption{\label{fig:figure1} Left figure shows three runs that failed to converge out of ten runs for an actor-critic method called deterministic policy gradient (DPG). The contour lines show the true expected return for the two parameters of the actor, while the white circle shows the starting parameter vector. For DPG, we approximate the value function by an incompatible linear function (details in Section \ref{sec:lqr}). None of the three runs make it to the maximum which is located at the bottom-left corner. By contrast, as shown in the middle figure, adding the TD-regularization fixes the instability and all the runs converge. The rightmost figure shows the estimated TD error for the two methods. We clearly see that TD-regularization reduces the error over time and improves not only stability and convergence but also the overall performance.}
\end{figure}

Actor-critic methods have achieved incredible results, showing super-human skills in complex real tasks such as playing Atari games and the game of Go~\citep{silver2016mastering, mnih2016asynchronous}.
Unfortunately, these methods can be extremely difficult to train and, in many cases, exhibit unstable behavior during learning.
One of the reasons behind their instability is the interplay between the actor and the critic during learning, e.g., a wrong step taken by one of them might adversely affect the other and can destabilize the learning \citep{dai2018boosting}.
This behavior is more common when nonlinear approximators, such as neural networks, are employed, but it could also arise even when simple linear functions are used\footnote{Convergence is assured when special types of linear functions known as \emph{compatible} functions are used to model the critic \citep{sutton1999policy, peters2008natural}. Convergence for other types of approximators is assured only for some algorithms and under some assumptions \citep{baird1995residual,konda2000actor,castro2008temporal}.}.
Figure~\ref{fig:figure1} (left) shows such an example where a linear function is used to model the critic but the method fails in three out of ten learning trajectories. Such behavior is only amplified when deep neural networks are used to model the critic. 

In this paper, we focus on developing methods to improve the stability of actor-critic methods.
Most of the existing methods have focused on stabilizing either the actor or the critic. For example, some recent works improve the stability of the critic by using a slowly-changing critic \citep{lillicrap2015continuous, mnih2015human, hessel2018rainbow}, a low-variance critic \citep{munos2016retrace, audrunas2018reactor}, or two separate critics to reduce their bias~\citep{hasselt2010double,fujimoto2018addressing}.
Others have proposed to stabilize the actor instead, e.g., by constraining its update using entropy or the Kullback-Leibler (KL) divergence~\citep{peters2010relative, schulman2015trust, akrour2016model, achiam2017constrained, nachum2017trust,haarnoja2018soft}.
In contrast to these approaches that focus on stabilizing either the actor or the critic, we focus on stabilizing the \emph{interaction} between them.

Our proposal is to stabilize the actor by penalizing its learning objective whenever the critic's estimate of the value function is highly inaccurate.
We focus on critic's inaccuracies that are caused by severe violation of the Bellman equation, as well as large temporal difference (TD) error.
We penalize for such inaccuracies by adding the critic's TD error as a regularization term in the actor's objective.
The actor is updated using the usual gradient update, giving us a simple yet powerful method which we call the \textit{TD-regularized actor-critic method}.
Due to this simplicity, our method can be used as a plug-and-play method to improve stability of existing actor-critic methods together with other critic-stabilizing methods. In this paper, we show its application to stochastic and deterministic actor-critic methods~\citep{sutton1999policy,silver2014deterministic}, trust-region policy optimization~\citep{schulman2015trust} and proximal policy optimization~\citep{schulman2017proximal}, together with Retrace~\citep{munos2016retrace} and double-critic methods~\citep{hasselt2010double,fujimoto2018addressing}.
Through evaluations on benchmark tasks, we show that our method is complementary to existing actor-critic methods, improving not only their stability but also their performance and data efficiency.

\subsection{Related Work}
Instability is a well-known issue in actor-critic methods, and many approaches have addressed it. The first set of methods do so by stabilizing the critic.
For instance, the so-called target networks have been regularly used in deep reinforcement learning to improve stability of TD-based critic learning methods~\citep{mnih2015human, lillicrap2015continuous, hasselt2010double, gu2016continuous}.
These target networks are critics whose parameters are slowly updated and are used to provide stable TD targets that do not change abruptly.
Similarly, \citet{fujimoto2018addressing} proposed to take the minimum value between a pair of critics to limit overestimation, and delay to update the policy parameters so that the per-update error is reduced. 
Along the same line, \citet{munos2016retrace} proposed to use truncated importance weighting to compute low-variance TD targets to stabilize critic learning. 
Instead, to avoid sudden changes in the critic, \citet{schulman2016high} proposed to constrain the learning of the value function such that the average Kullback-Leibler divergence between the previous and the current value function is sufficiently small. 
All of these methods can be categorized as methods that improve the stability by stabilizing the critic.

An alternative approach is to stabilize the actor by forcing it to not change abruptly.
This is often done by incorporating a Kullback-Leibler divergence constraint to the actor learning objective.
This constraint ensures that the actor does not take a large update step, ensuring safe and stable actor learning~\citep{peters2010relative,schulman2015trust,akrour2016model,achiam2017constrained,nachum2017trust}.

Our approach differs from both these approaches. Instead of stabilizing either the actor or the critic, we focus on stabilizing the interaction between the two. We do so by penalizing the mistakes made by the critic during the learning of the actor. Our approach directly addresses the instability  arising due to the interplay between the actor and the critic. 

\citet{prokhorov1997adaptive} proposed a method in a spirit similar to our approach where they only update the actor when the critic is sufficiently accurate. This \emph{delayed} update can stabilize the actor, but it might require many more samples to ensure an accurate critic, which could be time consuming and make the method very slow. Our approach does not have this issue.
Another recent approach proposed by \citet{dai2018boosting} uses a dual method to address the instability due to the interplay between the actor and the critic.
In their framework the actor and the critic have competing objectives, while ours encourages cooperation between them.

\section{Actor-Critic Methods and Their Instability}
\label{sec:rl_ac}
We start with a description of the reinforcement learning (RL) framework, and then review actor-critic methods. Finally, we discuss the sources of instability considered in this paper for actor-critic methods.

\subsection{Reinforcement Learning and Policy Search}
We consider RL in an environment governed by a Markov Decision Process (MDP). An MDP is described by the tuple $\langle \statespace, \actionspace, \prob, \rwd, \mu_1 \rangle$, where $\statespace \subseteq \realspace^{\dstate}$ is the state space, $\actionspace \subseteq \realspace^{\daction}$ is the action space, $\pmodel$ defines a Markovian transition probability density between the current $\state$ and the next state $\state'$ under action
$\action$, $\rmodel$ is the reward function, and $\mu_1$ is initial distribution for state $s_1$.
Given such an environment, the goal of RL is to learn to act. Formally, we want to find a \emph{policy} $\pi(a|s)$ to take an appropriate action when the environment is in state $s$. By following such a policy starting at initial state $s_1$, we obtain a sequence of states, actions and rewards $(\state_t, \action_t, r_t)_{t=1\ldots T}$, where $r_t = \rwd(s_t, a_t)$ is the reward at time $t$ ($T$ is the total timesteps). 
We refer to such sequences as \emph{trajectories} or \emph{episodes}.
Our goal is to find a policy that maximizes the expected return of such trajectories,
\begin{align}
	\max_\pi & \:\: \EVV{\mu_\pi(\state) \pi(\action|\state)}{Q^{\pi}(\state,\action)},
	\\
	\textrm{where} & \:\:\:
	Q^{\pi}(\state_t,\action_t) := \EVV{\prod_{i=t+1}^T \pi(a_{i}|s_{i})\prob(s_{i+1}|s_{i},a_{i})}{\sum_{i=t}^T \gamma^{i-t} r_{i+1}},  \label{eq:Q}
\end{align}
where $\mu_\pi(\state)$ is the state distribution under $\pi$, i.e., the probability of visiting state $\state$ under $\pi$, $Q^{\pi}(\state,\action)$ is the action-state value function (or Q-function) which is the expected return obtained by executing $a$ in state $s$ and then following $\pi$, and, finally, $\gamma \in [0,1)$ is the discount factor which assigns weights to rewards at different timesteps.  

One way to solve the optimization problem of Eq.~\eqref{eq:Q} is to use policy search~\citep{deisenroth2013survey}, e.g., we can use a parameterized policy function $\pi(\action|\state;\params)$ with the parameter $\params$ and take the following gradients steps
\begin{align}
	\params_{i+1} = \params_i + \alpha_\params \left. \nabla_{\params} \EVV{\mu_\pi(s) \pi(a|s;\params)}{ Q^{\pi}(s,a)} \right\vert_{\params = \params_i}, \label{eq:policy_grad}
\end{align}
where $\alpha_\params > 0$ is the stepsize and $i$ is a learning iteration. There are many ways to compute a stochastic estimate of the above gradient, e.g., we can first collect one trajectory starting from $s_1\sim \mu_1(s)$, compute Monte Carlo estimates $\widehat{Q}^{\pi}(s_t,a_t)$ of $Q^{\pi}(s_t,a_t)$, and then compute the gradient using REINFORCE~\citep{williams1992simple} as
\begin{align}
	&\nabla_{\params} \EVV{\mu_\pi(s) \pi(a|s;\params)}{ Q^{\pi}(s,a)} \approx  \sum_{t = 1}^T \left[ \nabla_{\params} \log \pi(a_t|s_t;\params) \right] \widehat{Q}^{\pi}(s_t,a_t) , \label{eq:reinforce}\\
	&\quad\quad\quad\textrm{where} \quad \widehat{Q}^{\pi}(s_t,a_t) = \rwd(s_t,a_t) + \gamma \widehat{Q}^{\pi}(s_{t+1},a_{t+1}),\quad  \forall t=1,2,\ldots, T-1, \nonumber
\end{align}
and $\widehat{Q}^{\pi}(s_T,a_T) := \rwd(s_T,a_T) = r_T$.
The recursive update to estimate $\widehat{Q}^{\pi}$ is due to the definition of $Q^{\pi}$ shown in Eq. \eqref{eq:Q}.
The above stochastic gradient algorithm is guaranteed to converge to a locally optimal policy when the stepsize are chosen according to Robbins-Monro conditions~\citep{robbins1985stochastic}.
However, in practice, using one trajectory might have high variance and the method requires averaging over many trajectories which could be inefficient. Many solutions have been proposed to solve this problem, e.g., baseline substraction methods~\citep{greensmith2004variance, gu2015muprop, wu2018variance}. Actor-critic methods are one of the most popular methods for this purpose.

\subsection{Instability of Actor-Critic Methods}
In actor-critic methods, the value function is estimated using a parameterized function approximator, i.e., $Q^{\pi}(s,a) \approx \widehat{Q}(s,a;\paramsQ)$, where $\paramsQ$ are the parameters of the approximator such as a linear model or a neural network. This estimator is called the \emph{critic} and can have much lower variance than traditional Monte Carlo estimators. Critic's estimate are used to optimize the policy $\pi(a|s;\params)$, also called the \emph{actor}. 

Actor-critic methods alternate between updating the parameters of the actor and the critic. Given the critic parameters $\paramsQ_i$ at iteration $i$ and its value function estimate $\widehat{Q}(s,a; \paramsQ_i)$, the actor can be updated using a policy search step similar to Eq. \eqref{eq:policy_grad}, 
\begin{align}
	\params_{i+1} &= \params_i + \alpha_\params \,\, \left. \nabla_{\params} \EVV{\mu_\pi(s) \pi(a|s;\params)}{\widehat{Q}(s,a; \paramsQ_i)} \right|_{\params=\params_i}. \label{eq:actor_update}
\end{align}
The parameters $\paramsQ$ are updated next by using a gradient method, e.g., we can minimize the \emph{temporal difference} (TD) error $\delta_Q$ using the following update: 
{
	\begin{gather}
		\paramsQ_{i+1} = \paramsQ_i + \alpha_\paramsQ \EVV{\mu_\pi(\state),\pi(\action|\state;\params),\pmodel}{ {\color{blue} \delta_Q(\state,\action,\state'; \params_{i+1}, \paramsQ_i)}  \nabla_{\paramsQ} \widehat{Q}(\state, \action; \paramsQ) \vert_{\paramsQ=\paramsQ_i} }, \nonumber \\ 
		\textrm{where} \quad {\color{blue} \delta_{Q}(\state,\action,\state';\params,\paramsQ)} := \rwd(s,a) + \gamma \EVV{{\pi(\action'|\state',\params)}} {\widehat{Q}(\state',\action'; {\paramsQ})} - \widehat{Q}(\state,\action;\paramsQ). \label{eq:td_q}
	\end{gather}
}%
The above update is approximately equivalent to minimizing the mean square of the TD error \citep{baird1995residual}. The updates \eqref{eq:actor_update} and \eqref{eq:td_q} together constitute a type of actor-critic method. The actor's goal is to optimize the expected return shown in Eq.~\eqref{eq:Q}, while the critic's goal is to provide an accurate estimate of the value function.

A variety of options are available for the actor and the critic, e.g., stochastic and deterministic actor-critic methods~\citep{sutton1999policy,silver2014deterministic}, trust-region policy optimization methods~\citep{schulman2015trust}, and proximal policy optimization methods~\citep{schulman2017proximal}.
Flexible approximators, such as deep neural networks, can be used for the actor and the critic. Actor-critic methods exhibit lower variance than the policy gradient methods that use Monte Carlo methods to estimate of the Q-function. They are also more sample efficient. Overall, these methods, when tuned well, can perform extremely well and achieved state-of-the-art performances on many difficult RL problems~\citep{mnih2015human, silver2016mastering}.

However, one issue with actor-critic methods is that they can be unstable, and may require careful tuning and engineering to work well~\citep{lillicrap2015continuous,dai2018boosting,henderson2017deep}.
For example, deep deterministic policy gradient (DDPG)~\citep{lillicrap2015continuous} requires implementation tricks such as target networks, and it is known to be highly sensitive to its hyperparameters~\citep{henderson2017deep}.
Furthermore, convergence is guaranteed only when the critic accurately estimates the value function~\citep{sutton1999policy}, which could be prohibitively expensive.
In general, stabilizing actor-critic methods is an active area of research.

One source of instability, among many others, is the interaction between the actor and the critic. The algorithm alternates between the update of the actor and the critic, so inaccuracies in one update might affect the other adversely. For example, the actor relies on the value function estimates provided by the critic. This estimate can have lower variance than the Monte Carlo estimates used in Eq.~\eqref{eq:reinforce}. However, Monte Carlo estimates are unbiased, because they maintain the recursive relationship between $\widehat{Q}^\pi(s_t,a_t)$ and $\widehat{Q}^\pi(s_{t+1},a_{t+1})$ which ensures that the expected value of $\widehat{Q}^\pi(s_t, a_t)$ is equal to the true value function (the expectation is taken with respect to the trajectories).
When we use function approximators, it is difficult to satisfy such recursive properties of the value function estimates. Due to this reason, critic estimates are often biased. At times, such inaccuracies might push the actor into wrong directions, from which the actor may never recover. In this paper, we propose a new method to address instability caused by such bad steps.

\section{TD-Regularized Actor-Critic}
\label{sec:tdreg}
As discussed in the previous section, the critic's estimate of the value function $\widehat{Q}(s,a;\paramsQ)$ might be biased, while Monte Carlo estimates can be unbiased. In general, we can ensure the unbiased property of an estimator if it satisfies the Bellman equation. 
This is because the following recursion ensures that each $\widehat{Q}(s,a;\paramsQ)$ is equal to the true value function in expectation, as shown below
\begin{align}
	\setlength{\abovedisplayskip}{2pt}
	\setlength{\belowdisplayskip}{2pt}
	\widehat{Q}(\state,\action; \paramsQ) = \mathcal{R}(s,a) + \gamma \EVV{\pmodel, \pi(a'|s'; \params)}{\widehat{Q}(\state',\action'; \paramsQ)}, \quad \forall s \in \mathcal{S}, \forall a \in \mathcal{A}. \label{eq:constraint_critic}
\end{align}
If $\widehat{Q}(\state',\action'; \paramsQ)$ is unbiased, $\widehat{Q}(\state,\action; \paramsQ)$ will also be unbiased. Therefore, by induction, all estimates are unbiased. Using this property, we modify actor's learning goal (Eq.~\eqref{eq:Q}) as the following constrained optimization problem
\begin{align}
	\max_\params \:\: & \EVV{\mu_\pi(\state), \pi(\action|\state; \params)}{\widehat{Q}(\state,\action; \paramsQ)}, \\
	\text{s.t.} \:\: & \widehat{Q}(\state,\action; \paramsQ) = \mathcal{R}(s,a) + \gamma \EVV{\pmodel, \pi(a'|s'; \params)}{\widehat{Q}(\state',\action'; \paramsQ)}, \forall s \in \mathcal{S}, \forall a \in \mathcal{A}. \label{eq:constraint_critic_modify}
\end{align}
We refer to this problem as the \emph{Bellman-constrained policy search}. At the optimum, when $\widehat{Q} = Q^{\pi}$, the constraint is satisfied, therefore the optimal solution of this problem is equal to the original problem of Eq. \eqref{eq:policy_grad}. For a suboptimal critic, the constraint is not satisfied and constrains the maximization of the expected return proportionally to the deviation in the Bellman equation. We expect this to prevent a large update in the actor when the critic is highly inaccurate for some state-action pairs.
The constrained formulation is attractive but computationally difficult due to the large number of constraints, e.g., for continuous state and action space this number might be infinite. In what follows, we make three modifications to this problem to obtain a practical method.

\subsection{Modification 1: Unconstrained Formulation Using the Quadratic Penalty Method}
\label{subsec:mod1}
Our first modification is to reformulate the constrained problem as an unconstrained one by using the quadratic penalty method~\citep{nocedal2006numerical}. In this method, given an optimization problem with equality constraints
\begin{equation}
	\setlength{\abovedisplayskip}{3pt}
	\setlength{\belowdisplayskip}{3pt}
	\max_\params f(\params), \quad \st h_j(\params) = 0, \quad j=1,2,\ldots,M
\end{equation}
we optimize the following function
\begin{equation}
	\setlength{\abovedisplayskip}{-1pt}
	\setlength{\belowdisplayskip}{0pt}
	\widehat{f}(\params, \tdcoeff) := f(\params) - \tdcoeff \sum_{j=1}^M b_j h_j^2(\params),
\end{equation}
where $b_j$ are the weights of the equality constraints and can be used to trade-off the effect of each constraint, and $\tdcoeff > 0$ is the parameter controlling the trade-off between the original objective function and the penalty function. When $\tdcoeff = 0$, the constraint does not matter, while when $\tdcoeff \to \infty$, the objective function does not matter.
Assuming that $\paramsQ$ is fixed, we propose to optimize the following quadratic-penalized version of the Bellman-constrained objective for a given $\tdcoeff$
{
	\begin{align}
		L(\params, \tdcoeff) &:=  \EVV{\mu_\pi(\state), \pi(\action|\state; \params)}{\widehat{Q}(\state,\action; \paramsQ)} \label{eq:bellman_penalize_int}\\
		&\phantom{:=}- \tdcoeff \iint b(s,a) \left(\mathcal{R}(s,a) + \gamma \EVV{\pmodel, \pi(a'|s'; \params)}{\widehat{Q}(\state',\action'; \paramsQ)} - \widehat{Q}(\state,\action; \paramsQ) \right)^2 \mathrm{d}s \mathrm{d}a,  \notag
	\end{align}
}%
where $b(s,a)$ is the weight of the constraint corresponding to the pair $(s,a)$.
%

\subsection{Modification 2: Reducing the Number of Constraints}
\label{subsec:mod2}
The integration over the entire state-action space is still computationally infeasible.
Our second modification is to focus on few constraints by appropriately choosing the weights $b(s,a)$. A natural choice is to use the state distribution $\mu_{\pi}(s)$ and the current policy $\pi(a|s;\params)$ to sample the candidate state-action pairs whose constraints we will focus on. In this way, we get the following objective
{
	\begin{align}
		L(\params, \tdcoeff) &:= \EVV{\mu_\pi(\state), \pi(\action|\state; \params)}{\widehat{Q}(\state,\action; \paramsQ)} \label{eq:bellman_penalty}\\
		&\phantom{:=} - \tdcoeff \EVV{\mu_\pi(\state), \pi(\action|\state;\params)} {\left(\mathcal{R}(s,a) + \gamma \EVV{{\color{blue} \pmodel}, \pi(a'|s'; \params)}{\widehat{Q}(\state',\action'; \paramsQ)} - \widehat{Q}(\state,\action; \paramsQ) \right)^2 } \notag.
	\end{align}
}%
The expectations in the penalty term in Eq. \eqref{eq:bellman_penalty} can be approximated using the observed state-action pairs. We can also use the same samples for both the original objective and the penalty term.
This can be regarded as a \emph{local} approximation where only a subset of the infinitely many constraint are penalized, and the subset is chosen based on its influence to the original objective function.

\subsection{Modification 3: Approximation Using the TD Error}
\label{subsec:mod3}
The final difficulty is that the expectation over $\mathcal{P}(s'|s,a)$ is inside the square function. This gives rise to a well-known issue in RL, called the \emph{double-sampling} issue~\citep{baird1995residual}. In order to compute an unbiased estimate of this squared expectation over $\mathcal{P}(s'|s,a)$, we require two sets of independent samples of $s'$ sampled from $\mathcal{P}(s'|s,a)$. The independence condition means that we need to independently sample many of the next states $s'$ from an identical state $s$.
This requires the ability to reset the environment back to state $s$ after each transition, which is typically impossible for many real-world systems.
Notice that the squared expectation over $\pi(a'|s';\params)$ is less problematic since we can always internally sample many actions from the policy without actually executing those actions on the environment.

To address this issue, we propose a final modification where we pull the expectation over $\mathcal{P}(s'|s,a)$ outside the square
{
	\begin{align}
		L(\params, \tdcoeff) &:= \EVV{\mu_\pi(\state), \pi(\action|\state; \params)}{\widehat{Q}(\state,\action; \paramsQ)} \label{eq:td_reg}\\
		&\phantom{:=}- \tdcoeff \mathbb{E}_{\mu_\pi(\state), \pi(\action|\state;\params), \pmodel} \Big[ \big(\underbrace{ \mathcal{R}(s,a) + \gamma \EVV{\pi(a'|s'; \params)}{\widehat{Q}(\state',\action'; \paramsQ)} - \widehat{Q}(\state,\action; \paramsQ) }_{:=\delta_Q(s,a,s';\params, \paramsQ)} \big)^2 \Big] \notag.
	\end{align}
}%
This step replaces the Bellman constraint of a pair $(s,a)$ by the \emph{temporal difference} (TD) error $\delta_Q(s,a,s';\params, \paramsQ)$ defined over the tuple $(s,a,s')$. We can estimate the TD error by using TD(0) or a batch version of it, thereby resolving the double-sampling issue. To further reduce the bias of TD error estimates, we can also rely on TD($\elicoeff$) (more details in Section \ref{subsec:gae-reg}).
Note that this final modification only approximately satisfies the Bellman constraint.

\subsection{Final Algorithm: the TD-Regularized Actor-Critic Method}
\label{subsec:final_alg}
Eq. \eqref{eq:td_reg} is the final objective we will use to update the actor. For the ease of notation, in the rest of the paper, we will refer to the two terms in $L(\params,\tdcoeff)$ using the following notation
\begin{align}
	L(\params, \tdcoeff) &:= J(\params) - \tdcoeff G(\params), \quad \textrm{where}  \label{eq:final1} \\
	J(\params) &:= \mathbb{E}_{\mu_\pi(\state), \pi(\action|\state; \params)}[\widehat{Q}(\state,\action; \paramsQ)],  \label{eq:final2} \\
	G(\params) &:= \mathbb{E}_{\mu_\pi(\state), \pi(\action|\state;\params), \pmodel} [\delta_Q(s,a,s';\params, \paramsQ)^2 ]. \label{eq:final3}
\end{align}
We propose to replace the usual policy search step (Eq. \eqref{eq:actor_update}) in the actor-critic method by a step that optimizes the TD-regularized objective for a given $\tdcoeff_i$ in iteration $i$,
\begin{align}
	\params_{i+1} &= \params_i + \alpha_\params \,\, \left( \left. \nabla_{\params} \EVV{\mu_\pi(s) \pi(a|s;\params)}{\widehat{Q}(s,a; \paramsQ_i)} \right|_{\params=\params_i} {\color{blue} - \tdcoeff_i \nabla_\params G(\params)\Big\vert_{\params=\params_i}  } \right). \label{eq:actor_update_tdreg}
\end{align}
The blue term is the extra penalty term involving the TD error, where we allow $\tdcoeff_i$ to change with the iteration. We can alternate between the above update of the actor and the update of the critic, e.g., by using Eq. \eqref{eq:td_q} or any other method.
We call this method the \textit{TD-regularized} actor-critic. We use the terminology ``regularization'' instead of ``penalization'' since it is more common in the RL and machine learning communities. 

A simple interpretation of Eq. \eqref{eq:td_reg} is that the actor is penalized for increasing the squared TD error, implying that the update of the actor favors policies achieving a small TD error. The main objective of the actor is still to maximize the estimated expected returns, but the penalty term helps to avoid bad updates whenever the critic's estimate has a large TD error. Because the TD error is an approximation to the deviation from the Bellman equation, we expect that the proposed method helps in stabilizing learning whenever the critic's estimate incurs a large TD error. 

In practice, the choice of penalty parameter is extremely important to enable a good trade-off between maximizing the expected return and avoiding the bias in the critic's estimate. In a typical optimization problem where the constraints are only functions of $\params$, it is recommended to slowly increase $\tdcoeff_i$ with $i$. This way, as the optimization progresses, the constraints become more and more important. However, in our case, the constraints also depend on $\paramsQ$
which changes with iterations, therefore the constraints also change with $i$. As long as the overall TD error of the critic decreases as the number of iterations increase, the overall penalty due the constraint will eventually decrease too. Therefore, we do not need to artificially make the constraints more important by increasing $\tdcoeff_i$. In practice, we found that if the TD error decreases over time, then $\tdcoeff_i$ can, in fact, be decreased with $i$. In Section \ref{sec:evaluation}, we use a simple decaying rule  $\tdcoeff_{i+1} = \kappa \tdcoeff_i$ where $0 < \kappa<1$ is a decay factor.

\section{Application of TD-Regularization to Actor-Critic Methods}
\label{sec:applications}
Our TD-regularization method is a general plug-and-play method that can be applied to any actor-critic method that performs policy gradient for actor learning.
In this section, we first demonstrate its applications to popular actor-critic methods including DPG~\citep{silver2014deterministic}, TRPO~\citep{schulman2015trust} and PPO~\citep{schulman2017proximal}. Subsequently, building upon the TD($\elicoeff$) error, we present a second regularization that can be used by actor-critic methods doing advantage learning, such as GAE~\citep{schulman2016high}.
For all algorithms, we show that our method only slightly increases computation time. The required gradients can be easily computed using automatic differentiation, making it very easy to apply our method to existing actor-critic methods.
Empirical comparison on these methods are given in Section~\ref{sec:evaluation}.

\subsection{TD-Regularized Stochastic Policy Gradient (SPG)}
\label{subsec:spg}
For a stochastic policy $\pi(a|s;\params)$, the gradient of Eq. \eqref{eq:td_reg} can be computed  using the chain rule and log-likelihood ratio trick~\citep{williams1992simple, sutton1998reinforcement}.
Specifically, the gradients of TD-regularized stochastic actor-critic are given by
\begin{align}
	\nabla_{\params} J(\params) 
	&= \EVV{\mu_\pi(\state), \pi(\action|\state; \params)}{ \nabla_{\params} \log \pi(a|s;\params) \widehat{Q}(\state,\action; \paramsQ) }, \label{eq:spg_j}
	\\
	\nabla_{\params} G(\params) 
	&= \mathbb{E}_{\mu_\pi(\state), \pi(\action|\state;\params), \pmodel} \Big[ \nabla_{\params} \log \pi(a|s;\params) \delta_Q(s,a,s';\params, \paramsQ)^2 \big.  \notag \\
	&\phantom{=} + 2\gamma \mathbb{E}_{\pi(\action'|\state';\params)} \left. \left[ \nabla_\params \log\pi(\action'|\state';\params) \delta_Q(\state,\action,\state';\params, \paramsQ) \widehat{Q}(\state',\action';{\paramsQ}) \right]\right].\label{eq:spg_g}
\end{align}
When compared to the standard SPG method, TD-regularized SPG requires only extra computations to compute $\nabla_{\params} \log\pi(\action'|\state';\params)$ and $\delta_{Q}(s,a,s';\params,\paramsQ)$.

\subsection{TD-Regularized Deterministic Policy Gradient (DPG)}
\label{subsec:dpg}
DPG~\citep{silver2014deterministic} is similar to SPG but learns a deterministic policy ${a=\pi(s;\params)}$.
To allow exploration and collect samples, DPG uses a behavior policy $\beta(a|s)$\footnote{\citet{silver2014deterministic} showed that DPG can be more advantageous than SPG as deterministic policies have lower variance. However, the behavior policy has to be chosen appropriately.}. Common examples are $\epsilon$-greedy policies or Gaussian policies. Consequently, the state distribution $\mu_{\pi}(s)$ is replaced by $\mu_{\beta}(s)$ and the expectation over the policy $\pi(a|s; \params)$ in the regularization term is replaced by an expectation over a behavior policy $\beta(a|s)$. The TD error does not change, but the expectation over $\pi(a'|s',\params)$ disappears.
TD-regularized DPG components are
\begin{align}
	J_\textsub{dpg}(\params) &:= \EVV{\mu_\beta(s)}{ \widehat{Q}(s,\pi(s;\params); \paramsQ) }, \label{eq:dpg_j}
	\\
	G_\textsub{dpg}(\params) &:= \EVV{\mu_\beta(\state), \beta(a|s), \prob(s'|s,a)} {\delta_Q(s,a,s';\params, \paramsQ)^2},\label{eq:dpg_g}
	\\
	\delta_{Q}(\state,\action,\state';\params,\paramsQ) &:= \rwd(s,a) + \gamma \widehat{Q}(\state',\pi(\state';\params); {\paramsQ}) - \widehat{Q}(\state, a;\paramsQ).
\end{align}
Their gradients can be computed by the chain rule and are given by
{
	\begin{align}
		\nabla_{\params} J_\textsub{dpg}(\params) 
		&= \EVV{\mu_\beta(\state)}{\nabla_\params \pi(\state;\params) \nabla_\action \widehat{Q}(\state,\action; \paramsQ)\big\vert_{\action=\pi(s;\params)} } , \label{eq:dpg_tdreg}
		\\
		\nabla_{\params} G_\textsub{dpg}(\params) 
		&= 2\gamma\EV_{\mu_\beta(\state),\beta(a|s),\prob(\state'|\state,\action)} \Big[\delta_{Q}(\state,\action,\state';\params, {\paramsQ}) \nabla_\params \pi(\state';\params) \nabla_{\action'}\widehat{Q}(\state',a';{\paramsQ})\big\vert_{\action'=\pi(s';\params)}\Big].
	\end{align}
}%
The gradient of the regularization term requires extra computations to compute $\delta_{Q}(\state,\action,\state';\params, {\paramsQ})$, $\nabla_\params \pi(\state';\params)$, and $\nabla_{\action'}\widehat{Q}(\state',\action';{\paramsQ})$.

\subsection{TD-Regularized Trust-Region Policy Optimization (TRPO)}
\label{subsec:trpo}
In the previous two examples, the critic estimates the Q-function. In this section, we demonstrate an application to a case where the critic estimates the V-function.
The V-function of $\pi$ is defined as $V^\pi(s) := \mathbb{E}_{\pi(a|s)}[ Q^{\pi}(s,a) ]$, and satisfies the following Bellman equation
\begin{align}
	V^\pi(s) = \EVV{\pi(a|s)}{ \mathcal{R}(s,a) } + \gamma \EVV{\pi(a|s), \pmodel} {V(s')}, \qquad \forall \state \in \statespace.
\end{align}
The TD error for a critic $\widehat{V}(s;\paramsQ)$ with parameters $\paramsQ$ is
\begin{align}
	\delta_{V}(s,s';\paramsQ) := \mathcal{R}(s,a) + \gamma \widehat{V}(s'; \paramsQ) - \widehat{V}(s; \paramsQ). \label{eq:td_v}
\end{align}
One difference compared to previous two sections is that $\delta_V(s,s'; \paramsQ)$ does not directly contain $\pi(a|s;\params)$. We will see that this greatly simplifies the update.
Nonetheless, the TD error still depends on $\pi(a|s;\params)$ as it requires to sample the action $a$ to reach the next state $s'$.
Therefore, the TD-regularization can still be applied to stabilize actor-critic methods that use a V-function critic.

In this section, we regularize TRPO~\citep{schulman2015trust}, which uses a V-function critic and solves the following optimization problem
\begin{align}
	\max_{\params} \: L_\textsub{trpo}(\params,\tdcoeff) &:= J_\textsub{trpo}(\params) - \tdcoeff G_\textsub{trpo}(\params), \qquad \mathrm{s.t.}~\EVV{\mu_\pi(s)} {\mathrm{KL}(\pi || \pi_{\mathrm{old}})} \leq \epsilon , \label{eq:trpo1}
	\\[-4pt]
	\textrm{ where } \:\:\:\:\: J_\textsub{trpo}(\params) &:= \EVV   {\mu_\pi(s),\pi(a|s;\params)}{\rho(\params) \widehat{A}(s,a; \paramsQ)} , \label{eq:trpo2}
	\\ 
	G_\textsub{trpo}(\params) &:= \EVV{\mu_\pi(s), \pi(\action|\state;\params),\prob(\state'|\state,\action)}{\rho(\params) \delta_{V}(\state,\state'; \paramsQ)^2} , \label{eq:trpo3}
\end{align}
where $\rho(\params) = {\pi(\action|\state;\params)}/{\pi(\action|\state; \params_{\mathrm{old}})}$ are importance weights. $\mathrm{KL}$ is the Kullback-Leibler divergence between the new learned policy $\pi(\action|\state; \params)$ and the old one $\pi(\action|\state; \params_{\mathrm{old}})$, and helps in ensuring small policy updates.  $\widehat{A}(s,a ; \paramsQ)$ is an estimate of the advantage function ${A^\pi(\state,\action) := Q^\pi(\state,\action) - V^\pi(\state)}$ computed by learning a V-function critic $\smash{\widehat{V}(s;\paramsQ)}$ and approximating $Q^\pi(\state,\action)$ either by Monte Carlo estimates or from $\smash{\widehat{V}(s;\paramsQ)}$ as well. We will come back to this in Section~\ref{subsec:gae-reg}. 
The gradients of $J_\textsub{trpo}(\params)$ and $G_\textsub{trpo}(\params)$ are
\begin{align}
	\nabla_{\params} J_\textsub{trpo}(\params) 
	&= \EVV{\mu_{\pi}(s), \pi(\action|\state; \params)}{ \rho(\params) \nabla_{\params} \log \pi(a|s;\params) \widehat{A}(\state,\action; \paramsQ) }, \label{eq:trpo_grad1}
	\\
	\nabla_{\params} G_\textsub{trpo}(\params) 
	&= \EVV{\mu_{\pi}(s), \pi(\action|\state; \params), \pmodel}{ \rho(\params) \nabla_{\params} \log \pi(a|s;\params) \delta_{V}(\state, \state'; \paramsQ)^2}.\label{eq:trpo_grad2}
\end{align}
The extra computation for TD-regularized TRPO only comes from computing the square of the TD error $\delta_{V}(s,s';\paramsQ)^2$. 

Notice that, due to linearity of expectations, TD-regularized TRPO can be understood as performing the standard TRPO with a TD-regularized advantage $\widehat{A}_{\tdcoeff}(s,a;\paramsQ) := \widehat{A}(s,a;\paramsQ) - \tdcoeff\mathbb{E}_{\pmodel}[\delta_V(s,s';\paramsQ)^2]$. 
This greatly simplifies implementation of our TD-regularization method.
In particular, TRPO performs natural gradient ascent to approximately solve the KL constraint optimization problem\footnote{Natural gradient ascent on a function $f(\params)$ updates the function parameters $\params$ by ${\params \leftarrow \params + \alpha_\params \vec{F}^{-1}(\params) \vec{g}(\params)}$, where $g(\params)$ is the gradient and $ \vec{F}(\params) $ is the Fisher information matrix.}.
By viewing TD-regularized TRPO as TRPO with regularized advantage, we can use the same natural gradient procedure for TD-regularized TRPO.

\subsection{TD-Regularized Proximal Policy Optimization (PPO)}
\label{subsec:ppo}
PPO~\citep{schulman2017proximal} simplifies the optimization problem of TRPO by removing the KL constraint, and instead uses clipped importance weights and a pessimistic bound on the advantage function
\begin{align}
	\underset{\params}{\max} \: & \EVV{\mu_{\pi}(s), \pi(\action | \state; \params)}{ \min \{ \rho(\params) \widehat A(\state,\action; \paramsQ), \: \rho_\varepsilon(\params)\widehat{A}(\state,\action; \paramsQ)\} } \label{eq:ppo},
\end{align}
where the $\rho_\varepsilon(\params)$ is the importance ratio $\rho(\params)$ clipped between $[1-\varepsilon, 1+\varepsilon]$ and $0 < \varepsilon < 1$ represents the update stepsize (the smaller $\varepsilon$, the more conservative the update is).
By clipping the importance ratio, we remove the incentive for moving $\rho(\params)$ outside of the interval $[1-\varepsilon, 1 + \varepsilon]$, i.e., for moving the new policy far from the old one. By taking the minimum between the clipped and the unclipped advantage, the final objective is a lower bound (i.e., a pessimistic bound) on the unclipped objective. 

Similarly to TRPO, the advantage function $\widehat{A}(s,a ; \paramsQ)$ is computed using a V-function critic $\widehat{V}(s; \paramsQ)$, thus we could simply use the regularization in Eq.~\eqref{eq:trpo3}. However, the TD-regularization would not benefit from neither importance clipping nor the pessimistic bound, which together provide a way of performing small safe policy updates. For this reason, we propose to modify the TD-regularization as follows
{
	\begin{flalign}
		&\max_{\params} L_\textsub{ppo}(\params,\tdcoeff) := J_\textsub{ppo}(\params) - \tdcoeff G_\textsub{ppo}(\params), \qquad \textrm{ where }  \label{eq:ppo1}
		\\[-4pt] 
		&\:J_\textsub{ppo}(\params) := \EVV{\mu_{\pi}(s), \pi(\action | \state; \params)}{ \min \{ \rho(\params) \widehat A(\state,\action; \paramsQ), \:  \rho_\varepsilon(\params)\widehat{A}(\state,\action; \paramsQ)\} }, \label{eq:ppo2}
		\\ 
		&G_\textsub{ppo}(\params) := \EVV{\mu_{\pi}(s), \pi(\action | \state; \params), \pmodel}{ \max \{ \rho(\params) \delta_V(\state,\state'; \paramsQ)^2, \: \rho_\varepsilon(\params)\delta_V(\state,\state'; \paramsQ)^2\} }, \label{eq:ppo3}
	\end{flalign}
}%
i.e., we apply importance clipping and the pessimistic bound also to the TD-regularization. The gradients of $J_\textsub{ppo}(\params)$ and $G_\textsub{ppo}(\params)$ can be computed as
{
	\begin{align}
		\nabla_\params J_\textsub{ppo}(\params) &\phantom{:}= \EVV{\mu_{\pi}(s), \pi(\action | \state; \params)}{ f(\params) } \qquad \textrm{ where } 
		\\
		f(\params) &:= \begin{cases}
			\rho(\params)\nabla_{\params} \log \pi(a|s;\params)\widehat{A}(\state,\action; \paramsQ) & \hspace*{10pt} \text{if $\rho(\params)\widehat{A}(\state,\action; \paramsQ) < \rho_\varepsilon(\params)\widehat{A}(\state,\action; \paramsQ)$}
			\\
			0 & \hspace*{10pt} \text{otherwise,}
		\end{cases} \nonumber
		\\
		\nabla_\params G_\textsub{ppo}(\params) &\phantom{:}= \EVV{\mu_{\pi}(s), \pi(\action | \state; \params)}{ g(\params) } \qquad \textrm{ where } 
		\\
		g(\params) &:= \begin{cases}
			\rho(\params)\nabla_{\params} \log \pi(a|s;\params)\delta_V(\state,\state'; \paramsQ)^2 & \text{if $\rho(\params)\delta_V(\state,\state'; \paramsQ)^2 > \rho_\varepsilon(\params)\delta_V(\state,\state'; \paramsQ)^2$}
			\\
			0 & \text{otherwise.}
		\end{cases} \nonumber
	\end{align}
}%

\subsection{GAE-Regularization}
\label{subsec:gae-reg}
In Sections \ref{subsec:trpo} and \ref{subsec:ppo}, we have discussed how to apply the TD-regularization when a V-function critic is learned. The algorithms discussed, TRPO and PPO, maximize the advantage function $\widehat{A}(\state,\action;\paramsQ)$ estimated using a V-function critic $\widehat{V}(s,\paramsQ)$. Advantage learning has a long history in RL literature~\citep{baird1993advantage} and one of the most used and successful advantage estimator is the generalized advantage estimator (GAE)~\citep{schulman2016high}. 
In this section, we build a connection between GAE and the well-known TD($\elicoeff$) method \citep{sutton1998reinforcement} to propose a different regularization, which we call the \textit{GAE-regularization}.
We show that this regularization is very convenient for algorithms already using GAE, as it does not introduce any computational cost, and has interesting connections with other RL methods.

Let the $n$-step return $R_{t:t+n}$ be the sum of the first $n$ discounted rewards plus the estimated value of the state reached in $n$ steps, i.e.,
\begin{align}
	R_{t:t+n} &:= r_{t} + \gamma r_{t+1} + \ldots + \gamma^{n-1}r_{t+n-1} + \gamma^{n} \widehat{V}(\state_{t+n}; \paramsQ), \quad 0 \leq t \leq T - n, \label{eq:n_return}
	\\
	R_{t:T+1} &:= {\sum_{i=t}^T \gamma^{i-t} r_{i}}. \label{eq:n_return_T}
\end{align}
The full-episode return $R_{t:T+1}$ is a Monte Carlo estimate of the value function.
The idea behind TD($\elicoeff$) is to replace the TD error target $r_t + \gamma \widehat{V}(\state_{t+1},\paramsQ)$ with the average of the $n$-step returns, each weighted by $\elicoeff^{n-1}$, where $\elicoeff \in [0,1]$ is a decay rate. Each $n$-step return is also normalized by $1-\elicoeff$ to ensure that the weights sum to 1. The resulting TD($\elicoeff$) targets are the so-called $\elicoeff$-returns
\begin{align}
	R^\elicoeff_t &:= (1-\elicoeff)\sum_{i=t}^{T-1} \elicoeff^{i-t} R_{t:i+1} + \elicoeff^{T-t} R_{t:T+1}, \label{eq:lambda_return}
\end{align}
and the corresponding TD($\elicoeff$) error is
\begin{equation}
	\delta_V^\elicoeff(\state_t,\state_{t+1};\paramsQ) := R_t^\elicoeff - \widehat V(\state_t;\paramsQ). \label{eq:tde_lambda}
\end{equation}
From the above equations, we see that if $\elicoeff = 0$, then the $\elicoeff$-return is the TD target $\smash{R^0_t = r_t + \gamma\widehat{V}(\state_{t+1};\paramsQ)}$. If $\elicoeff = 1$, then $R^1_t = R_{t:T+1}$ as in Monte Carlo methods. In between are intermediate methods that control the bias-variance trade-off between TD and Monte Carlo estimators by varying $\elicoeff$. As discussed in Section~\ref{sec:rl_ac}, in fact, TD estimators are biased, while Monte Carlo are not. The latter, however, have higher variance.

Motivated by the same bias-variance trade-off, we propose to replace $\delta_V$ with $\delta_V^\elicoeff$ in Eq.~\eqref{eq:trpo3} and~\eqref{eq:ppo3}, i.e., to perform TD($\elicoeff$)-regularization.
Interestingly, this regularization is equivalent to regularize with the GAE advantage estimator, as shown in the following.
Let $\delta_V$ be an approximation of the advantage function \citep{schulman2016high}.
Similarly to the $\elicoeff$-return, we can define the $n$-step advantage estimator
\begin{align}
	A_{t:t+n} &:= \delta_V(\state_t,\state_{t+1};\paramsQ) + \gamma \delta_V(\state_{t+1},\state_{t+2};\paramsQ) + \ldots + \gamma^{n-1}\delta_V(\state_{t+n-1},\state_{t+n};\paramsQ), \nonumber
	\\
	&\phantom{:}= R_{t:t+n} - \widehat{V}(\state_t;\paramsQ) \label{eq:n_adv}, \hspace*{4.cm} 0 \leq t \leq T-n,
	\\
	A_{t:T+1} &:= R_{t:T+1} - \widehat{V}(\state_t;\paramsQ). \label{eq:n__T}
\end{align}
Following the same approach of TD($\elicoeff$), GAE advantage estimator uses exponentially weighted averages of $n$-step advantage estimators
\begin{align}
	\widehat{A}^\elicoeff(\state_t,\action_t;\paramsQ) &:= (1-\elicoeff)\sum_{i=t}^{T-1} \elicoeff^{i-t} A_{t:i+1} + \elicoeff^{T-t} A_{t:T+1}, \label{eq:gae}
\end{align}
From the above equation, we see that GAE estimators are discounted sums of TD errors. Similarly to TD($\elicoeff$), if $\elicoeff = 0$ then the advantage function estimate is just the TD error estimate, i.e.,  $\widehat{A}^0(\state_t,\action_t; \paramsQ) = \allowbreak r_t + \widehat{V}(\state_{t+1};\paramsQ) - \widehat{V}(\state_t;\paramsQ)$. If $\elicoeff = 1$ then the advantage function estimate is the difference between the Monte Carlo estimate of the return and the V-function estimate, i.e., $\smash{\widehat{A}^1(\state_t,\action_t; \paramsQ) = R_{t:T+1} - \widehat V(\state_t; \paramsQ)}$. 
Finally, plugging Eq.~\eqref{eq:n_adv} into Eq.~\eqref{eq:gae}, we can rewrite the GAE estimator as
\begin{align}
	\widehat{A}^\elicoeff(\state_t,\action_t;\paramsQ) &:= (1-\elicoeff)\sum_{i=t}^{T-1} \elicoeff^{i-t} (R_{t:i+1}-\widehat V(\state_t;\paramsQ)) + \elicoeff^{T-t}(R_{t:T+1} - \widehat{V}(\state_t;\paramsQ))\nonumber
	\\
	&\phantom{:}= (1-\elicoeff)\sum_{i=t}^{T-1} \elicoeff^{i-t} R_{t:i+1} + \elicoeff^{T-t} R_{t:T+1} - \widehat V(\state_t;\paramsQ) \nonumber
	\\
	& \phantom{:}= R_t^\elicoeff - \widehat V(\state_t;\paramsQ) = \delta_V^\elicoeff(\state_t,\state_{t+1}; \paramsQ), \label{eq:gae_eq_tdl}
\end{align}
i.e., the GAE advantage estimator is equivalent to the TD($\elicoeff$) error estimator.
Therefore, using the TD($\elicoeff$) error to regularize actor-critic methods is equivalent to regularize with the GAE estimator, yielding the following quadratic penalty
\begin{equation}
	G_\textsub{gae}(\params) = \EVV{\mu_\pi(s), \pi(\action|\state;\params)}{\widehat{A}^\elicoeff(\state,\action; \paramsQ)^2} , \label{eq:gae_reg}
\end{equation}
which we call the \textit{GAE-regularization}. 
The GAE-regularization is very convenient for methods which already use GAE, such as TRPO and PPO\footnote{For PPO, we also apply the importance clipping and pessimistic bound proposed in Eq.~\eqref{eq:ppo3}.}, as it does not introduce any computational cost. Furthermore, the decay rate $\elicoeff$ allows to tune the bias-variance trade-off between TD and Monte Carlo methods\footnote{We recall that, since GAE approximates $A^\pi(\state,\action)$ with the TD($\elicoeff$) error, we are performing the same approximation presented in Section \ref{subsec:mod3}, i.e., we are still approximately satisfying the Bellman constraint.}. In Section \ref{sec:evaluation} we present an empirical comparison between the TD- and GAE-regularization.

Finally, the GAE-regularization has some interesting interpretations.
As shown by \citet{belousov2017f}, minimizing the squared advantage function is equivalent to maximizing the average reward with a penalty over the Pearson divergence between the new and old state-action distribution $\mu_{\pi}(\state)\pi(\action|\state;\params)$, and a hard constraint to satisfy the stationarity condition $\iint \mu_\pi(\state)\pi(\action|\state;\params)\pmodel \diff\state\diff\action = \mu_\pi(\state'), \forall \state'$. The former is to avoid overconfident policy update steps, while the latter is the dual of the Bellman equation (Eq.~\eqref{eq:constraint_critic}). Recalling that the GAE-regularization approximates the Bellman equation constraint with the TD($\elicoeff$) error, the two methods are very similar. The difference in the policy update is that the GAE-regularization replaces the stationarity condition with a soft constraint, i.e., the penalty\footnote{For the critic update, instead, \citet{belousov2017f} learn the V-function parameters together with the policy rather than separately as in actor-critic methods.}.
\\
Interestingly, Eq.~\eqref{eq:gae_reg} is also equivalent to minimizing the variance of the centered GAE estimator, i.e., $\smash{\EV[{(\widehat{A}^\elicoeff(\state,\action; \paramsQ) - \mu_{\hat A})^2}] = \var[\widehat{A}^\elicoeff(\state,\action; \paramsQ)]}$. Maximizing the mean of the value function estimator and penalizing its variance is a common approach in risk-averse RL called \textit{mean-variance} optimization~\citep{tamar2012policy}. Similarly to our method, this can be interpreted as a way to avoid overconfident policy updates when the variance of the critic is high. By definition, in fact, the expectation of the true advantage function of any policy is zero\footnote{{$\EVV{\pi(\action|\state)}{A^{\pi}(\state,\action)} = \EVV{\pi(\action|\state)}{Q^{\pi}(\state,\action) - V^\pi(\state)} = \EVV{\pi(\action|\state)}{Q^{\pi}(\state,\action)} - V^\pi(\state) = V^\pi(\state) -V^\pi(\state) = 0$.}}, thus high-variance is a sign of an inaccurate critic.

\section{Evaluation}
\label{sec:evaluation}
We propose three evaluations. First, we study the benefits of the TD-regularization in the 2-dimensional linear-quadratic regulator (LQR). In this domain we can compute the true Q-function, expected return, and TD error in closed form, and we can visualize the policy parameter space. 
We begin this evaluation by setting the initial penalty coefficient to $\tdcoeff_0 = 0.1$ and then decaying it according to $\tdcoeff_{i+1} = \kappa\tdcoeff_i$ where $\kappa=0.999$. 
We then investigate different decaying factors $\kappa$ and the behavior of our approach in the presence of non-uniform noise.
The algorithms tested are DPG and SPG. For DPG, we also compare to the twin delayed version proposed by~\citet{fujimoto2018addressing}, which achieved state-of-the-art results. 

The second evaluation is performed on the single- and double-pendulum swing-up tasks~\citep{yoshikawa1990foundations,brockman2016openai}. 
Here, we apply the proposed TD- and GAE-regularization to TRPO together with and against Retrace~\citep{munos2016retrace} and double-critic learning~\citep{hasselt2010double}, both state-of-the-art techniques to stabilize the learning of the critic.

The third evaluation is performed on OpenAI Gym \citep{brockman2016openai} continuous control benchmark tasks with the MuJoCo physics simulator \citep{todorov2012mujoco} and compares TRPO and PPO with their TD- and GAE-regularized counterparts. Due to time constraints, we were not able to evaluate Retrace on this tasks as well.
Details of the hyperparameter settings are given in Appendix \ref{app:mujoco}.

For the LQR and the pendulum swing-up tasks, we tested each algorithm over 50 trials with fixed random seeds. At each iteration, we turned the exploration off and evaluated the policy over several episodes.
Due to limited computational resources, we tested MuJoCo experiments over five trials with fixed random seeds. For TRPO, the policy was evaluated over 20 episodes without exploration noise. For PPO, we used the same samples collected during learning, i.e., including exploration noise.

\subsection{2D Linear Quadratic Regulator (LQR)}
\label{sec:lqr}
The LQR problem is defined by the following discrete-time dynamics
\begin{equation*}
	\state' = A \state +B \action + \gaussian(0,0.1^2),\quad\quad\quad\quad
	a = K\state, \quad\quad\quad\quad
	\rmodel = -\state^{\T} X\state - {\action}^{\T} Y \action,
\end{equation*}
where $A,B,X,Y \in \realspace^{d \times d}$, $X$ is a symmetric positive semidefinite matrix, $Y$ is a symmetric positive definite matrix, and $K \in \realspace^{d \times d}$ is the control matrix. The policy parameters we want to learn are $\params = \text{vec}(K)$. 
Although low-dimensional, this problem presents some challenges. First, the policy can easily make the system unstable. The LQR, in fact, is stable only if the matrix $(A+BK)$ has eigenvalues of magnitude smaller than one. Therefore, small stable steps have to be applied when updating the policy parameters, in order to prevent divergence.
Second, the reward is unbounded and the expected negative return can be extremely large, especially at the beginning with an initial random policy. As a consequence, with a common zero-initialization of the Q-function, the initial TD error can be arbitrarily large. Third, states and actions are unbounded and cannot be normalized in [0,1], a common practice in RL.

Furthermore, the LQR is particularly interesting because we can compute in closed form both the expected return and the Q-function, being able to easily assess the quality of the evaluated algorithms. More specifically, the Q-function is quadratic in the state and in the action, i.e.,
\begin{equation*}
	Q^\pi(\state,\action) = Q_0 + \state^{\T} Q_{\state\state}\state + \action^{\T} Q_{\action\action}\action + \state^{\T} Q_{\state\action}\action,
\end{equation*}
where $Q_0, Q_{\state\state}, Q_{\action\action}, Q_{\state\action}$ are matrices computed in closed form given the MDP characteristics and the control matrix $K$.
To show that actor-critic algorithms are prone to instability in the presence of function approximation error, we approximate the Q-function linearly in the parameters $\widehat Q(\state,\action;\paramsQ) = \phi(\state,\action)^{\T}\paramsQ$, where $\phi(\state,\action)$ includes linear, quadratic and cubic features.

Along with the expected return, we show the trend of two mean squared TD errors (MSTDE): one is estimated using the currently learned $\widehat Q(\state,\action;\paramsQ)$, the other is computed in closed form using the true $Q^\pi(\state,\action)$ defined above. It should be noticed that $Q^\pi(\state,\action)$ is not the optimal Q-function (i.e., of the optimal policy), but the true Q-function with respect to the current policy. 
For details of the hyperparameters and an in-depth analysis, including an evaluation of different Q-function approximators, we refer to Appendix~\ref{app:lqr}.

\subsubsection{Evaluation of DPG and SPG}
DPG and TD-regularized DPG (DPG TD-REG) follow the equations presented in Section \ref{subsec:dpg}.
The difference is that DPG maximizes only Eq. \eqref{eq:dpg_j}, while DPG TD-REG objective includes Eq. \eqref{eq:dpg_g}.
TD3 is the twin delayed version of DPG presented by~\citet{fujimoto2018addressing}, which uses two critics and delays policy updates. TD3 TD-REG is its TD-regularized counterpart. 
For all algorithms, all gradients are optimized by ADAM~\citep{kingma2014adam}. 
After 150 steps, the state is reset and a new trajectory begins.

\begin{figure}[t]
	\begin{minipage}[t]{\textwidth}
		\begin{subfigure}[t]{.325\linewidth}
			\centering
			\includegraphics[width=\textwidth]{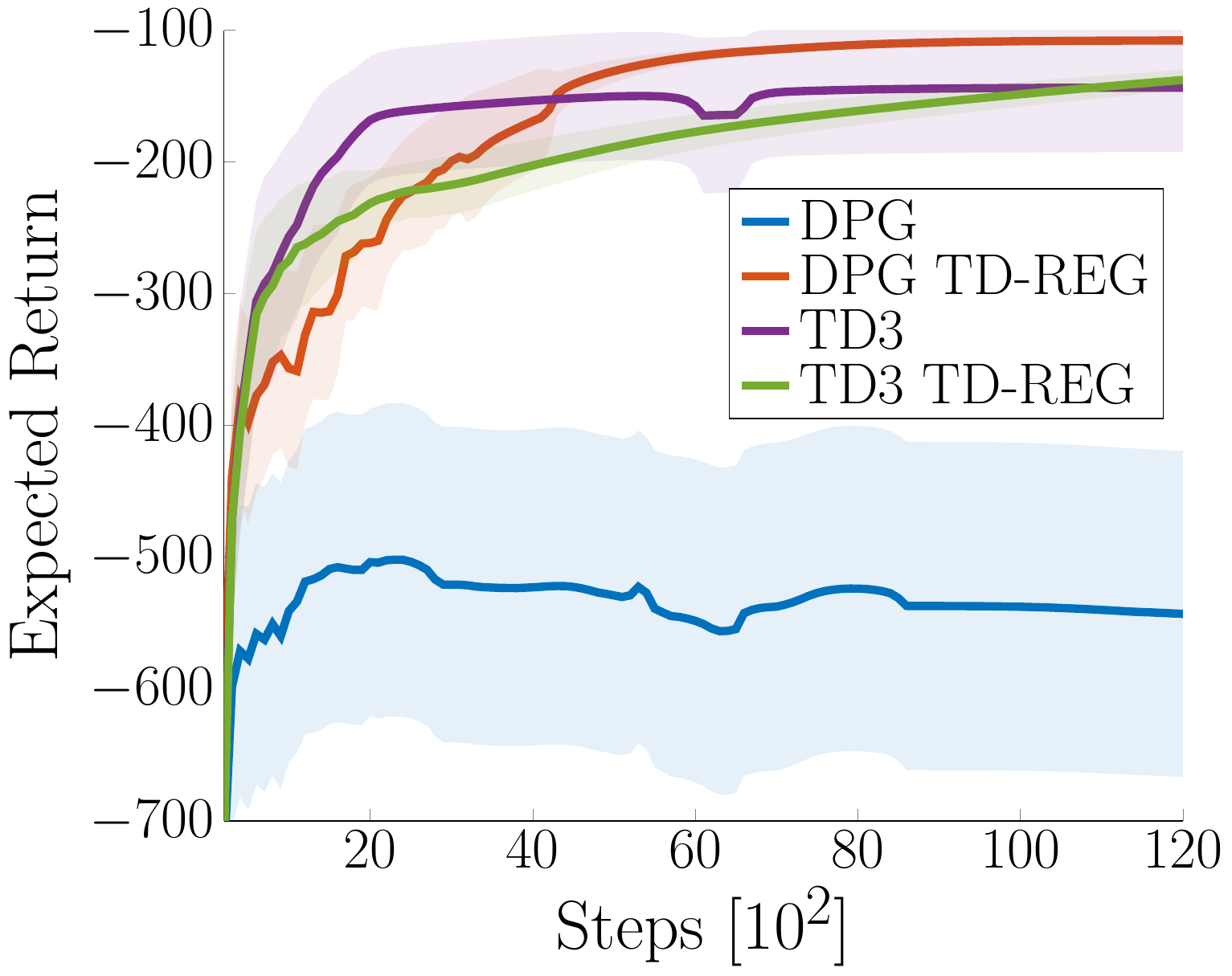}
			\caption{\label{fig:dpg_lqr_ret}}
		\end{subfigure}
		\hfill
		\begin{subfigure}[t]{.325\linewidth}
			\centering
			\includegraphics[width=\textwidth]{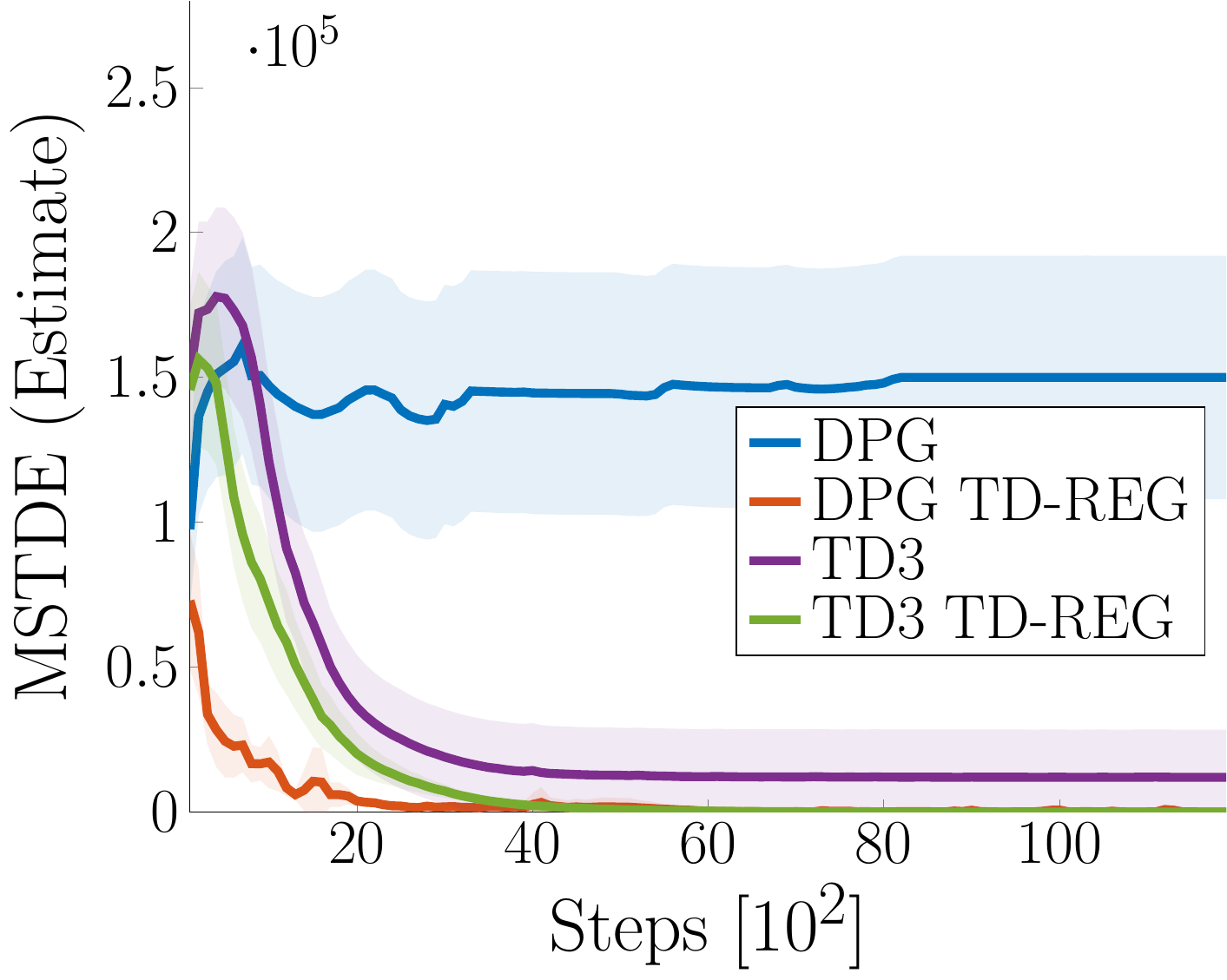}
			\caption{\label{fig:dpg_lqr_tdlearn}}
		\end{subfigure}
		\hfill
		\begin{subfigure}[t]{.325\linewidth}
			\centering
			\includegraphics[width=\textwidth]{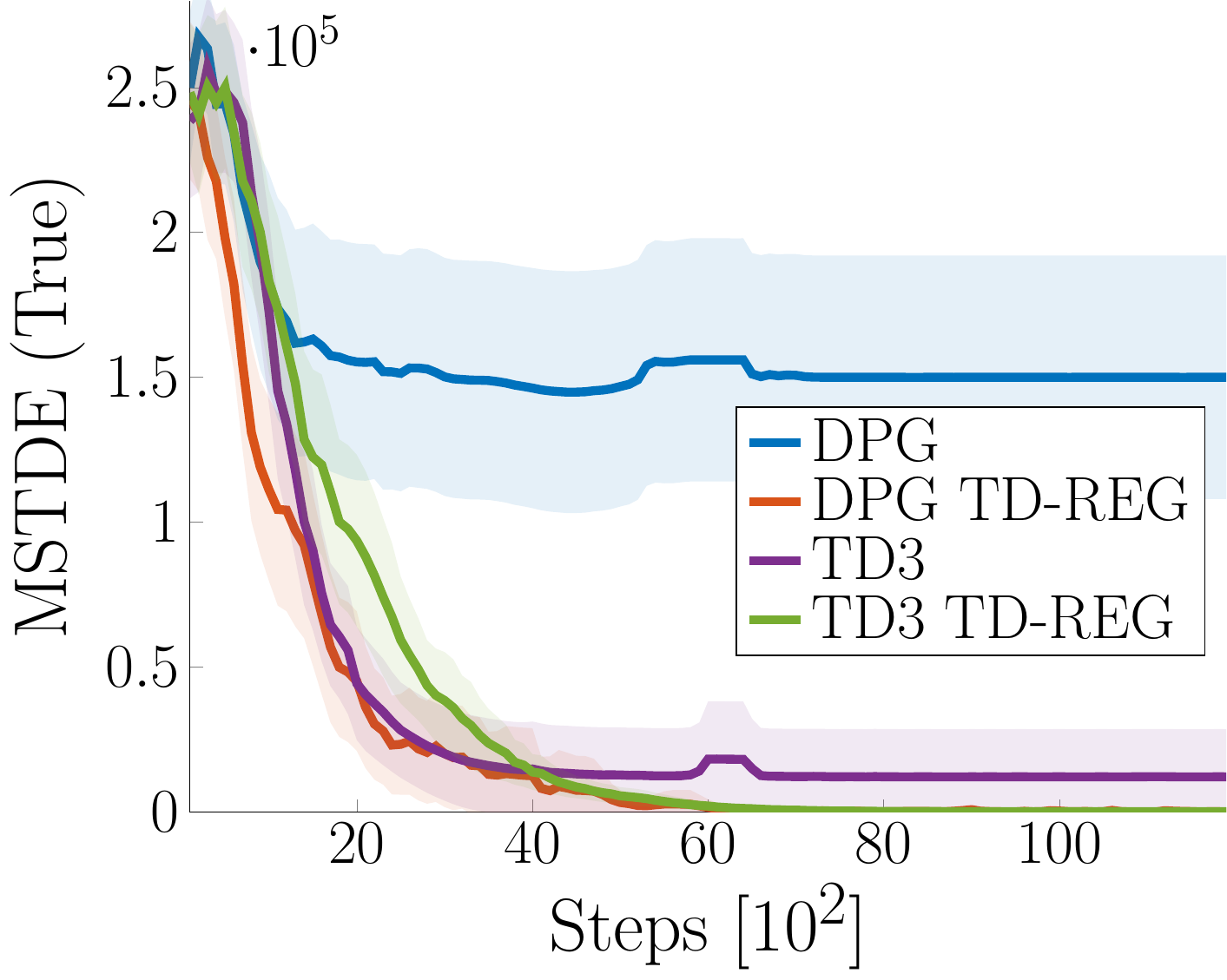}
			\caption{\label{fig:dpg_lqr_tdtrue}}
		\end{subfigure}
		\caption{\label{fig:dpg_lqr} DPG comparison on the LQR. Shaded areas denote 95\% confidence interval. DPG diverged 24 times out of 50, thus explaining its very large confidence interval. TD3 diverged twice, while TD-regularized algorithms never diverged. Only DPG TD-REG, though, always learned the optimal policy within the time limit.}
	\end{minipage}
	\\[1.5em]
	\begin{minipage}[t]{\textwidth}
		\begin{subfigure}[t]{.325\linewidth}
			\centering
			\includegraphics[width=\textwidth]{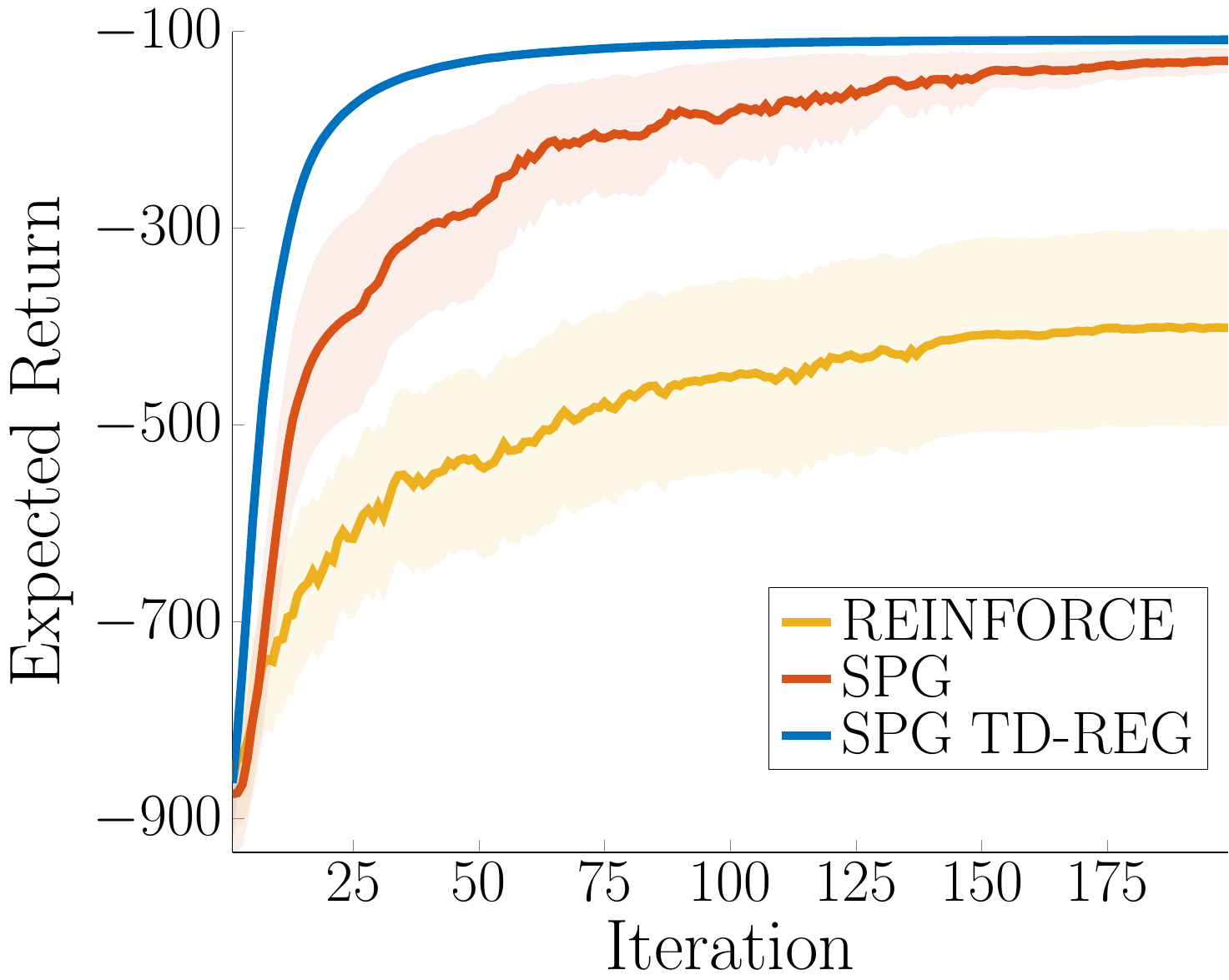}
			\caption{\label{fig:spg_lqr_ret}}
		\end{subfigure}
		\hfill
		\begin{subfigure}[t]{.325\linewidth}
			\centering
			\includegraphics[width=\textwidth]{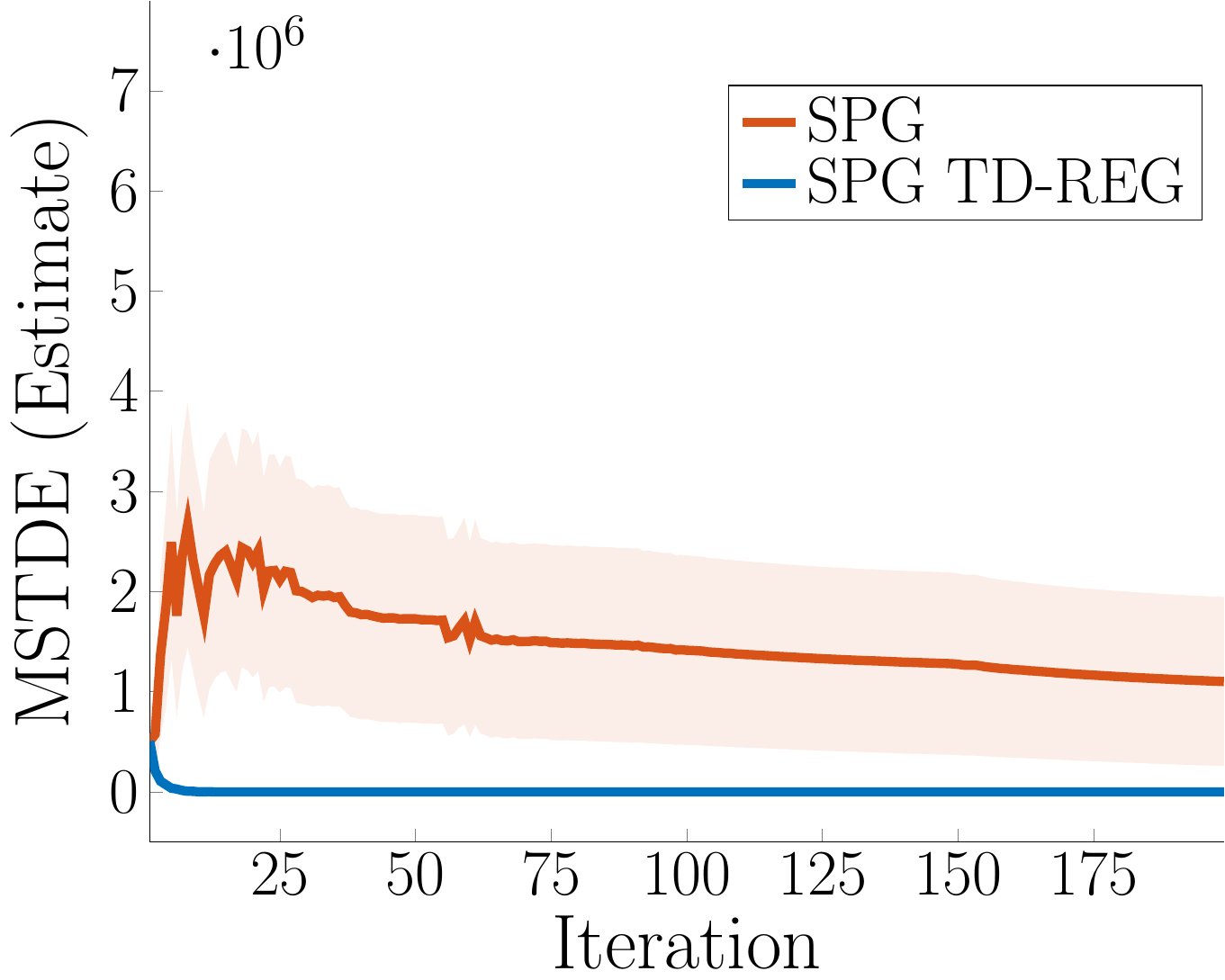}
			\caption{\label{fig:spg_lqr_td}}
		\end{subfigure}
		\hfill
		\begin{subfigure}[t]{.325\linewidth}
			\centering
			\includegraphics[width=\textwidth]{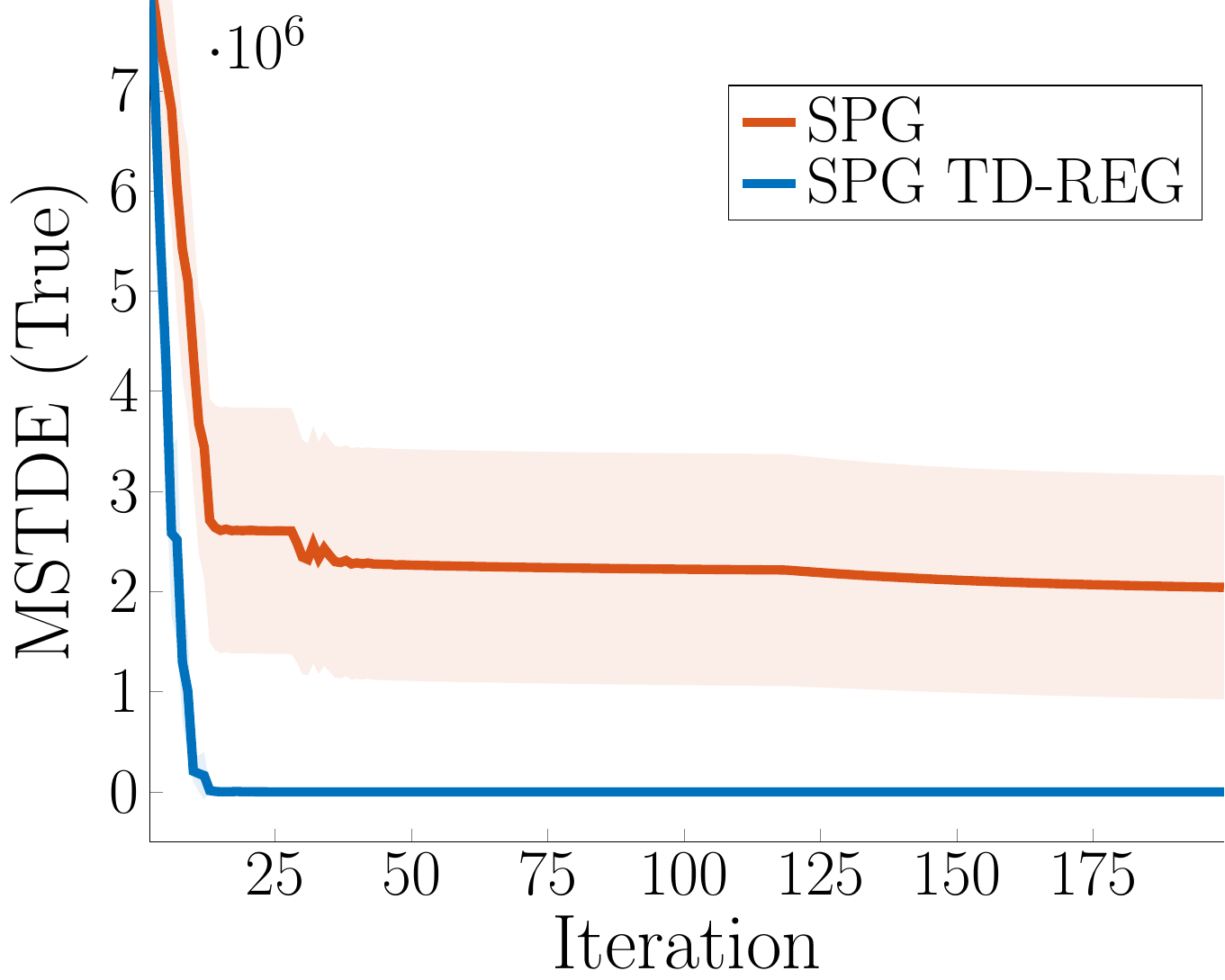}
			\caption{\label{fig:spg_lqr_tdtrue}}
		\end{subfigure}
		\caption{\label{fig:spg_lqr} SPG comparison on the LQR. One iteration corresponds to 150 steps. REINFORCE does not appear in the TD error plots as it does not learn any critic. SPG TD-REG shows an incredibly fast convergence in all runs. SPG-TD, instead, needs much more iterations to learn the optimal policy, as its critic has a much larger TD error. REINFORCE diverged 13 times out of 50, thus explaining its large confidence interval.}
	\end{minipage}
\end{figure}

As expected, because the Q-function is approximated with also cubic features, the critic is prone to overfit and the initial TD error is very large. 
Furthermore, the true TD error (Figure~\ref{fig:dpg_lqr_tdtrue}) is more than twice the one estimated by the critic (Figure~\ref{fig:dpg_lqr_tdlearn}), meaning that the critic underestimates the true TD error.
Because of the incorrect estimation of the Q-function, vanilla DPG diverges 24 times out of 50. TD3 performs substantially better, but still diverges two times. 
By contrast, TD-REG algorithms never diverges. Interestingly, only DPG TD-REG always converges to the true critic and to the optimal policy within the time limit, while TD3 TD-REG improves more slowly. Figure~\ref{fig:dpg_lqr} hints that this ``slow learning'' behavior may be due to the delayed policy update, as both the estimated and the true TD error are already close to zero by mid-learning. In Appendix~\ref{app:lqr} we further investigate this behavior and show that TD3 policy update delay is unnecessary if TD-REG is used.
The benefits of TD-REG in the policy space can also be seen in Figure~\ref{fig:figure1}. Whereas vanilla DPG falls victim to the wrong critic estimates and diverges, DPG TD-REG enables more stable updates. 

The strength of the proposed TD-regularization is also confirmed by its application to SPG, as seen in Figure~\ref{fig:spg_lqr}.
Along with SPG and SPG TD-REG, we evaluated REINFORCE~\citep{williams1992simple}, which does not learn any critic and just maximizes Monte Carlo estimates of the Q-function, i.e., $\widehat Q^\pi(\state_t,\action_t) =  {\sum_{i=t}^T \gamma^{i-t} r_{i}}$.
For all three algorithms, at each iteration samples from only one trajectory of 150 steps are collected and used to compute the gradients, which are then normalized. For the sake of completeness, we also tried to collect more samples per iteration, increasing the number of trajectories from one to five. In this case, all algorithms performed better, but still neither SPG nor REINFORCE matched SPG TD-REG, as they both needed several samples more than SPG TD-REG. More details in Appendix~\ref{app:lqr_spg}.

\begin{figure}[t]
	\begin{minipage}[t]{\textwidth}
		\begin{subfigure}[t]{.5\linewidth}
			\centering
			\includegraphics[width=\textwidth]{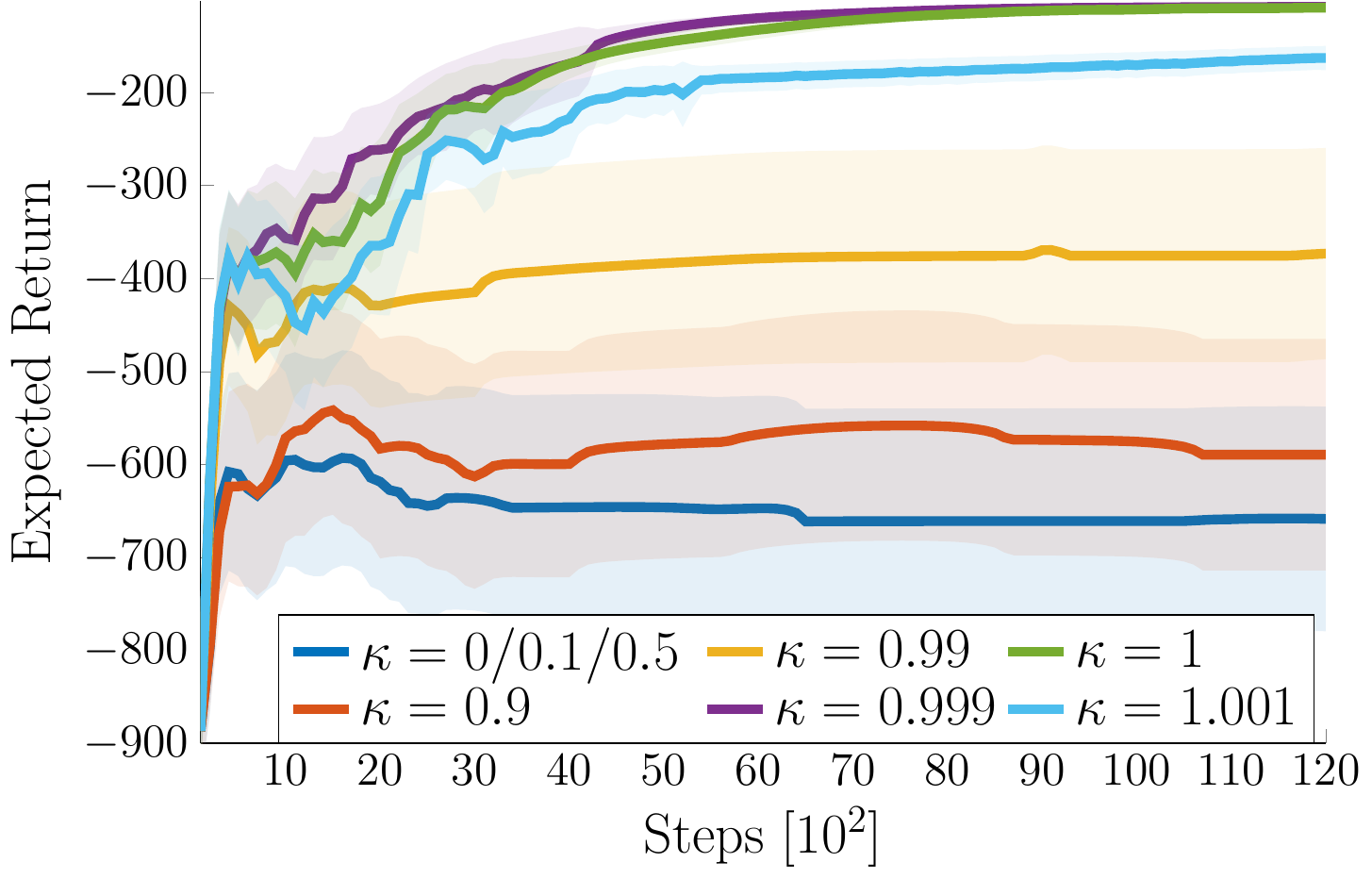}
			\subcaption{\label{fig:lambda_tuning_dpg}DPG}
		\end{subfigure}
		\hfill
		\begin{subfigure}[t]{.49\linewidth}
			\centering
			\includegraphics[width=\textwidth]{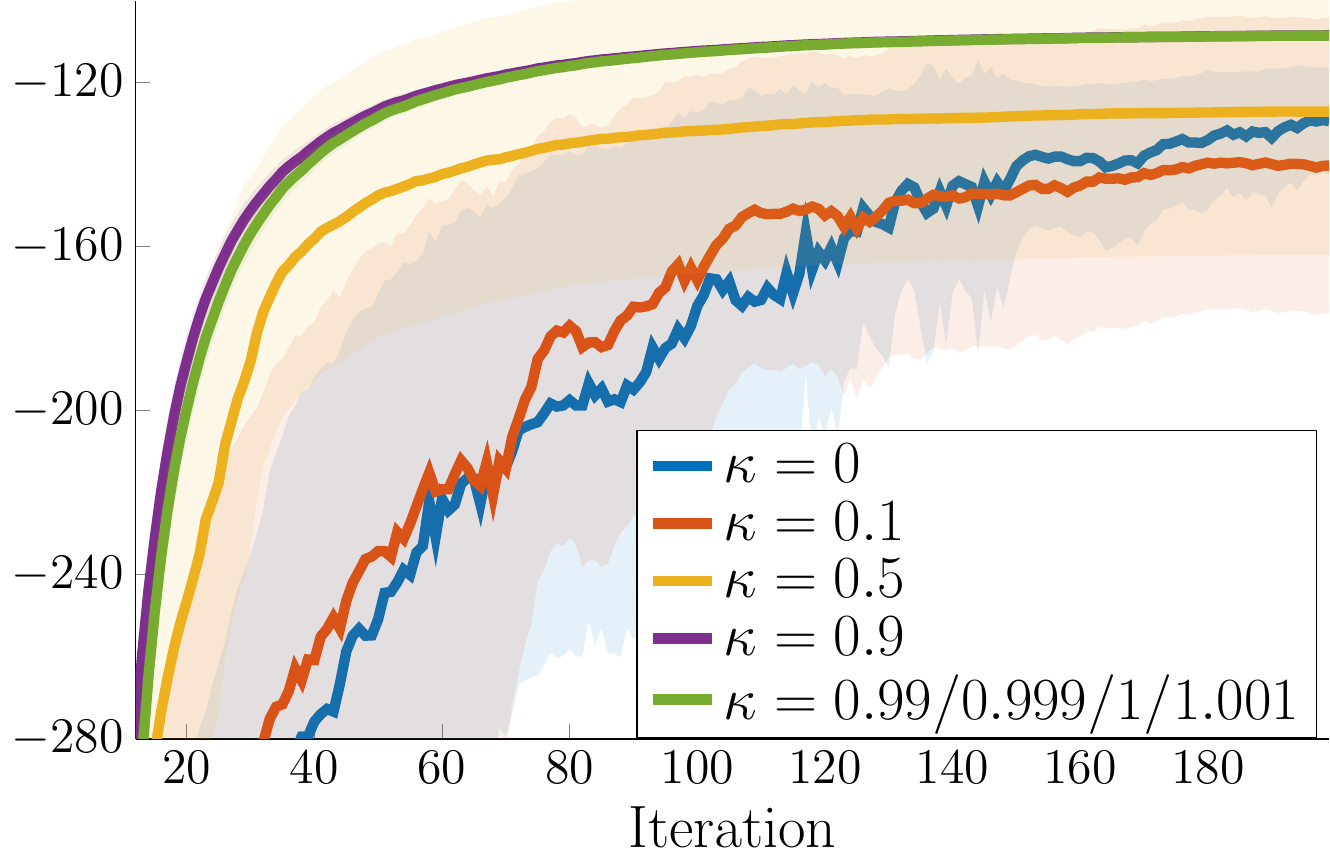}
			\subcaption{\label{fig:lambda_tuning_spg}SPG}
		\end{subfigure}
		\caption{\label{fig:lambda_tuning} Comparison of different values of $\kappa$. Shaded areas denote 95\% confidence interval. In order to provide enough regularization, $\kappa$ must be sufficiently large during the whole learning. With small values, in fact, $\tdcoeff$ vanishes and the TD-regularization is not in effect anymore.}
	\end{minipage}
\end{figure}

\subsubsection{Analysis of the TD-Regularization Coefficient $\tdcoeff$}
In Section \ref{subsec:final_alg} we have discussed that Eq. \eqref{eq:td_reg} is the result of solving a constrained optimization problem by penalty function methods. 
In optimization, we can distinguish two approaches to apply penalty functions \citep{boyd2004convex}.
\textit{Exterior} penalty methods start at optimal but infeasible points and iterate to feasibility as $\tdcoeff \to\infty$. By contrast, \textit{interior} penalty methods start at feasible but sub-optimal points and iterate to optimality as $\tdcoeff \to 0$.
In actor-critic, we usually start at infeasible points, as the critic is not learned and the TD error is very large. 
However, unlike classical constrained optimization, the constraint changes at each iteration, because the critic is updated to minimize the same penalty function. 
This trend emerged from the results presented in Figures \ref{fig:dpg_lqr} and \ref{fig:spg_lqr}, showing the change of the mean squared TD error, i.e., the penalty.

In the previous experiments we started with a penalty coefficient $\tdcoeff_0 = 0.1$ and decreased it at each policy update according to $\tdcoeff_{t+1} = \kappa\tdcoeff_t$, with $\kappa = 0.999$. 
In this section, we provide a comparison of different values of $\kappa$, both as decay and growth factor.
In all experiments we start again with $\tdcoeff_0 = 0.1$ and we test the following $\kappa$: 0, 0.1, 0.5, 0.9, 0.99, 0.999, 1, 1.001. 

As shown in Figures~\ref{fig:lambda_tuning_dpg} and~\ref{fig:lambda_tuning_spg}, results are different for DPG TD-REG and SPG TD-REG. 
In DPG TD-REG, 0.999 and 1 allowed to always converge to the optimal policy. Smaller $\kappa$ did not provide sufficient help, up to the point where 0.1 and 0.5 did not provide any help at all. However, it is not true that larger $\kappa$ yield better results, as with 1.001 performance decreases. This is expected, since by increasing $\tdcoeff$ we are also increasing the magnitude of the gradient, which then leads to excessively large and unstable updates. 

Results are, however, different for SPG TD-REG. First, 0.99, 0.999, 1, 1.001 all achieve the same performance. The reason is that gradients are normalized, thus the size of the update step cannot be excessively large and $\kappa > 1$ does not harm the learning. Second, 0.9, which was not able to help enough DPG, yields the best results with a slightly faster convergence. The reason is that DPG performs a policy update at each step of a trajectory, while SPG only at the end. Thus, in DPG $\tdcoeff$, which is updated after a policy update, will decay too quickly if a small $\kappa$ is used.

\begin{figure}[t]
	\centering
	\begin{minipage}[t]{\textwidth}
		\begin{subfigure}[t]{.325\linewidth}
			\centering
			\includegraphics[width=\textwidth]{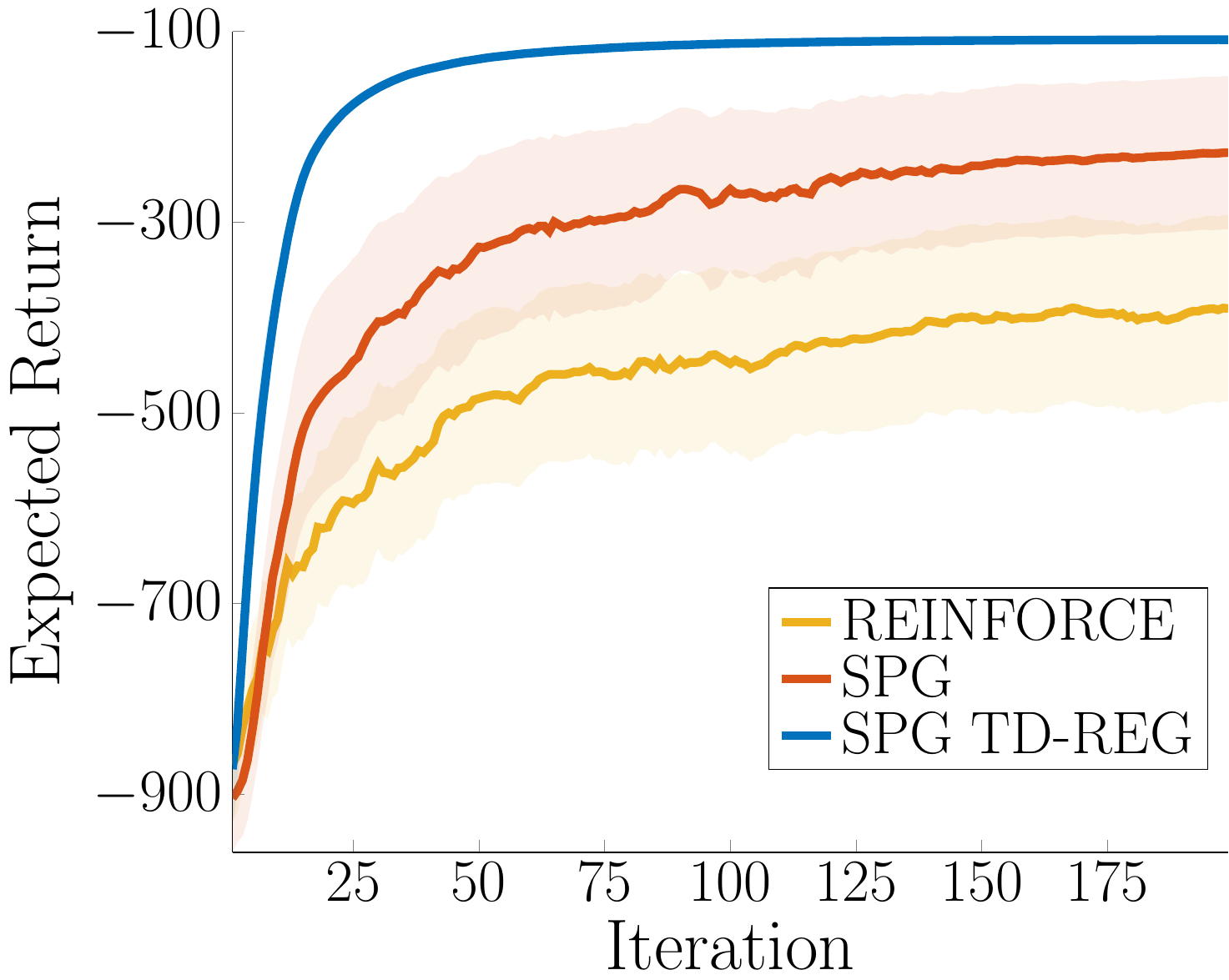}
			\caption{\label{fig:spg_noise_ret}}
		\end{subfigure}
		\hfill
		\begin{subfigure}[t]{.325\linewidth}
			\centering
			\includegraphics[width=\textwidth]{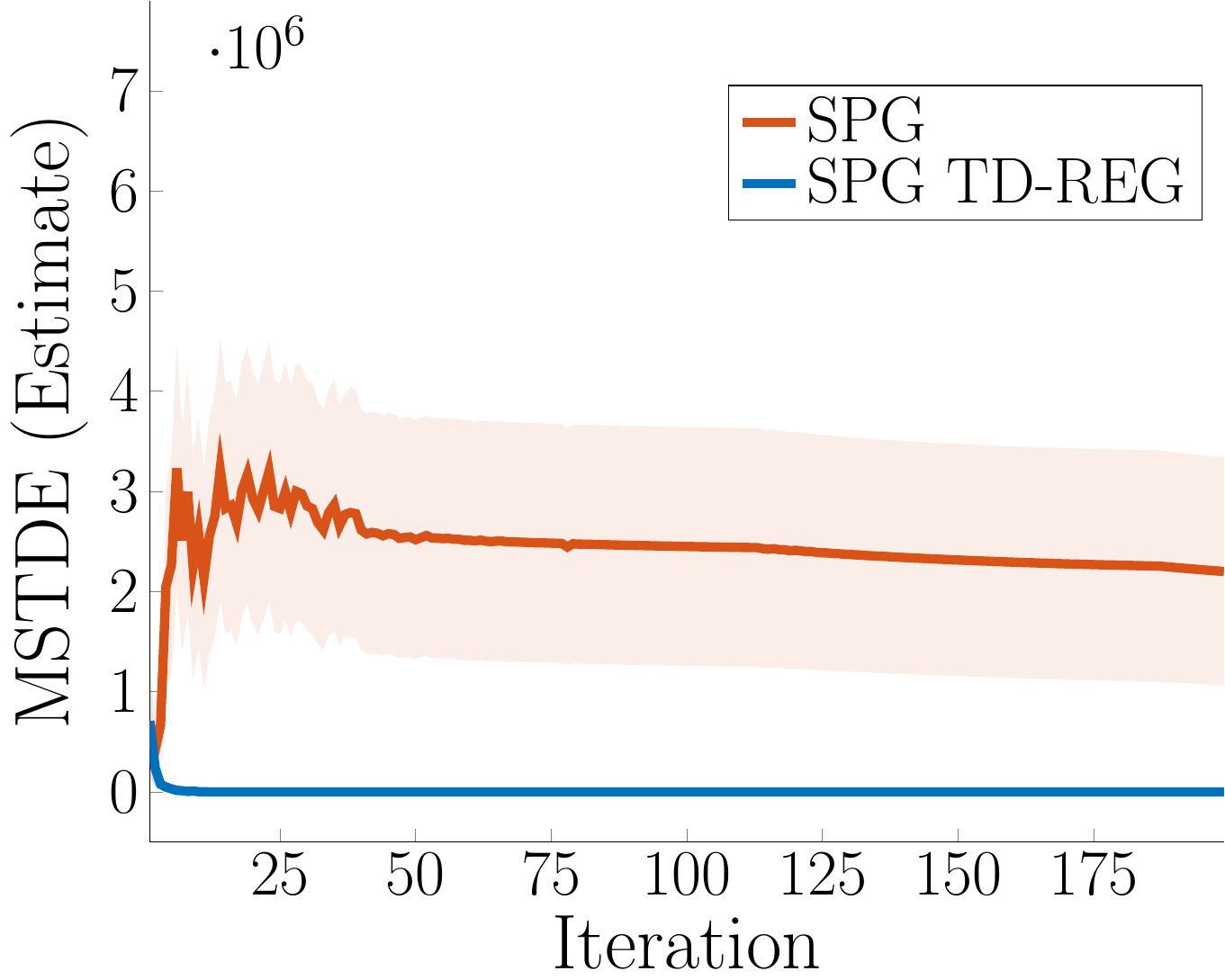}
			\caption{\label{fig:spg_noise_td}}
		\end{subfigure}
		\hfill
		\begin{subfigure}[t]{.325\linewidth}
			\centering
			\includegraphics[width=\textwidth]{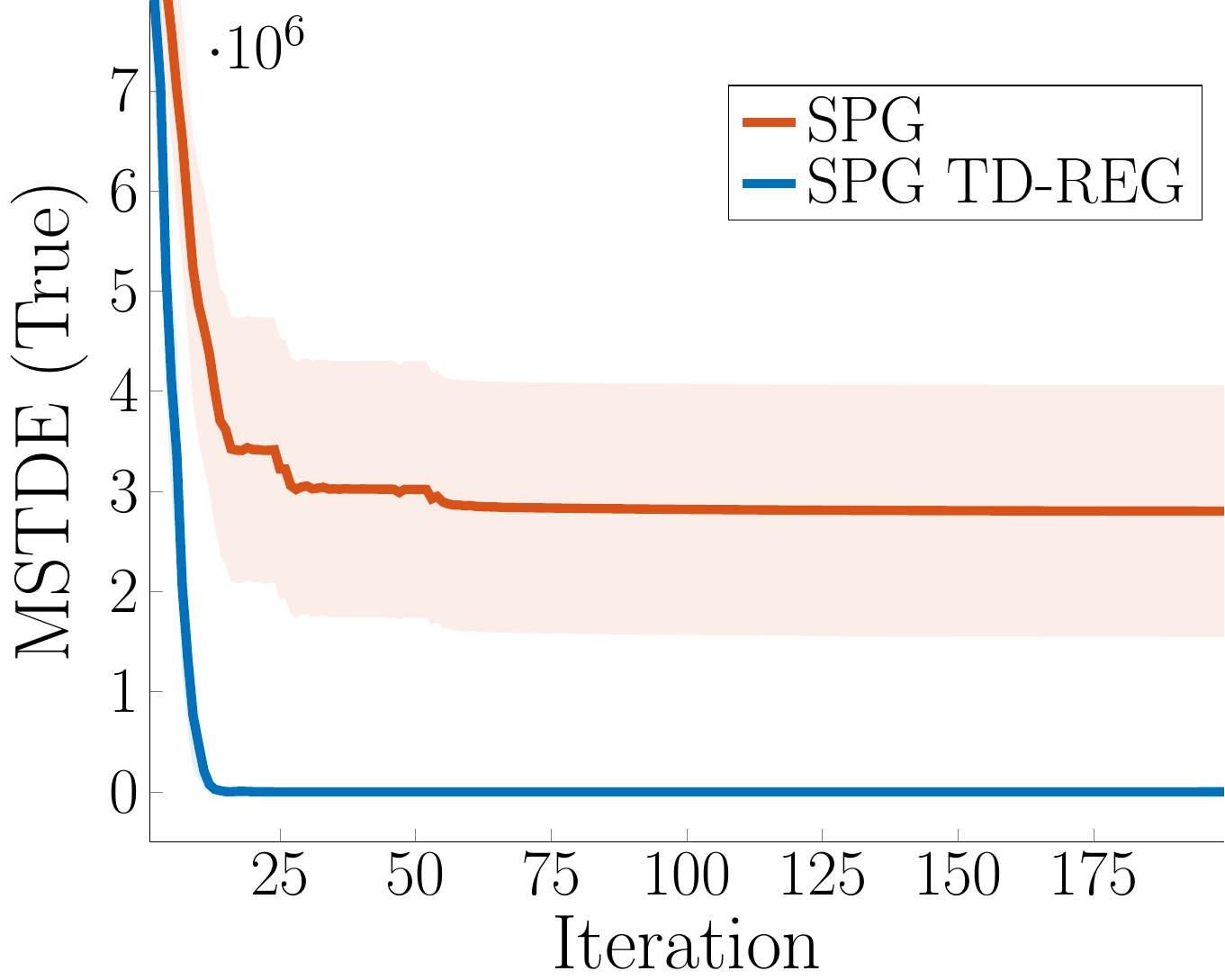}
			\caption{\label{fig:spg_noise_tdtrue}}
		\end{subfigure}
		\caption{\label{fig:spg_noise}SPG comparison on the LQR with non-uniform noise on the state observation. Shaded areas denote 95\% confidence interval. The TD error is not shown for REINFORCE as it does not learn any critic. Once again, SPG TD-REG performs the best and is not affected by the noise. Instead of being drawn to low-noise regions (which are far from the goal and correspond to low-reward regions), its actor successfully learns the optimal policy in all trials and its critic achieves a TD error of zero. Un-regularized SPG, which did not diverge in Figure~\ref{fig:spg_lqr_ret}, here diverges six times.}
	\end{minipage}
\end{figure}

\subsubsection{Analysis of Non-Uniform Observation Noise}
So far, we have considered the case of high TD error due to an overfitting critic and noisy transition function. However, the critic can be inaccurate also because of noisy or partially observable state.
Learning in the presence of noise is a long-studied problem in RL literature. To address every aspect of it and to provide a complete analysis of different noises is out of the scope of this paper. However, given the nature of our approach, it is of particular interest to analyze the effects of the TD-regularization in the presence of non-uniformly distributed noise in the state space. In fact, since the TD-regularization penalizes for high TD error, the algorithm could be drawn towards low-noise regions of the space in order to avoid high prediction errors. Intuitively, this may not be always a desirable behavior.
Therefore, in this section we evaluate SPG TD-REG when non-uniform noise is added to the observation of the state, i.e.,
\begin{equation*}
	\state_\textsub{obs} = \state_\textsub{true} + \frac{\gaussian(0,0.05)}{\texttt{clip}(\state_\textsub{true}, 0.1, 200)},
\end{equation*}
where $\state_\textsub{obs}$ is the state observed by the actor and the critic, and $\state_\textsub{true}$ is the true state. The clipping between $[0.1, 200]$ is for numerical stability.
The noise is Gaussian and inversely proportional to the state. Since the goal of the LQR is to reach $\state_\textsub{true} = 0$, near the goal the noise will be larger and, subsequently, the TD error as well. One may therefore expect that SPG TD-REG would lead the actor towards low-noise regions, i.e., away from the goal.
However, as shown in Figure \ref{fig:spg_noise}, SPG TD-REG is the only algorithm learning in all trials and whose TD error goes to zero. By contrast, SPG, which never diverged with exact observations (Figure~\ref{fig:spg_lqr_ret}), here diverged six times out of 50 (Figure \ref{fig:spg_noise_ret}). SPG TD-REG plots, instead, are the same in both Figures.
REINFORCE, instead, does not significantly suffer from the noisy observations, since it does not learn any critic.

\subsection{Pendulum Swing-up Tasks}
The pendulum swing-up tasks are common benchmarks in RL. Their goal is to swing-up and stabilize a single- and double-link pendulum from any starting position.
The agent observes the current joint position and velocity and acts applying torque on each joint. As the pendulum is underactuated, the agent cannot swing it up in a single step, but needs to gather momentum by making oscillatory movements. Compared to the LQR, these tasks are more challenging --especially the double-pendulum-- as both the transition and the value functions are nonlinear.

\begin{figure}[t]
	\centering
	\begin{subfigure}[t]{\textwidth}
		\centering
		\includegraphics[width=\textwidth]{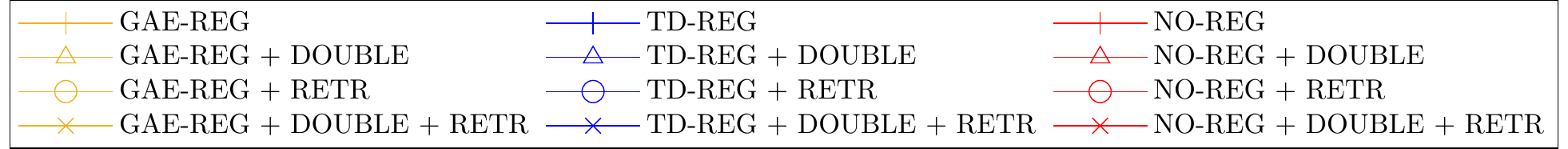}
	\end{subfigure}
	\\
	\begin{minipage}[t]{\textwidth}
		\begin{subfigure}[t]{.49\linewidth}
			\centering
			\includegraphics[width=\textwidth]{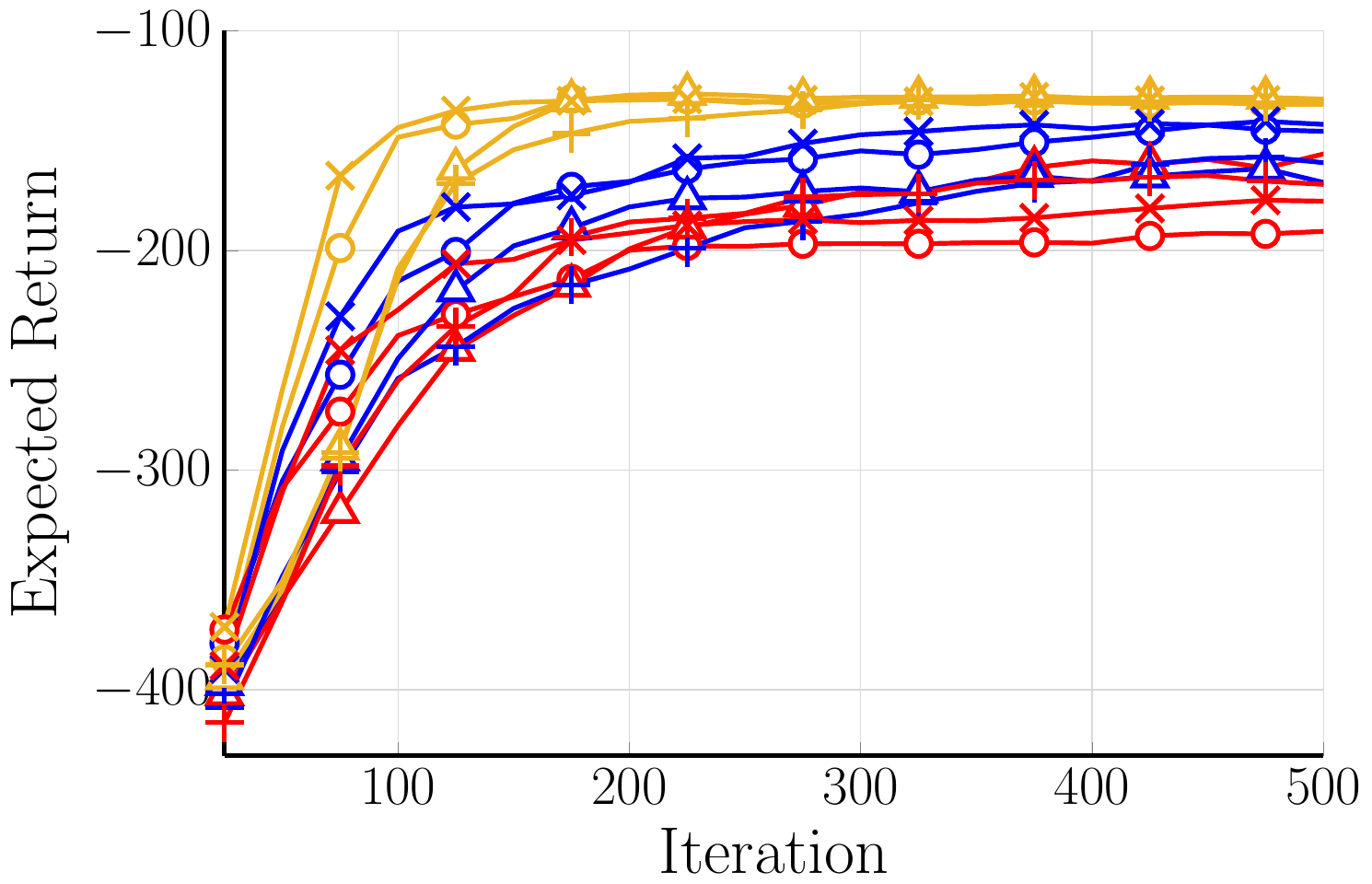}
			\caption{Single-pendulum.\label{fig:pend_ret}}
		\end{subfigure}
		\hfill
		\begin{subfigure}[t]{.49\linewidth}
			\centering
			\includegraphics[width=\textwidth]{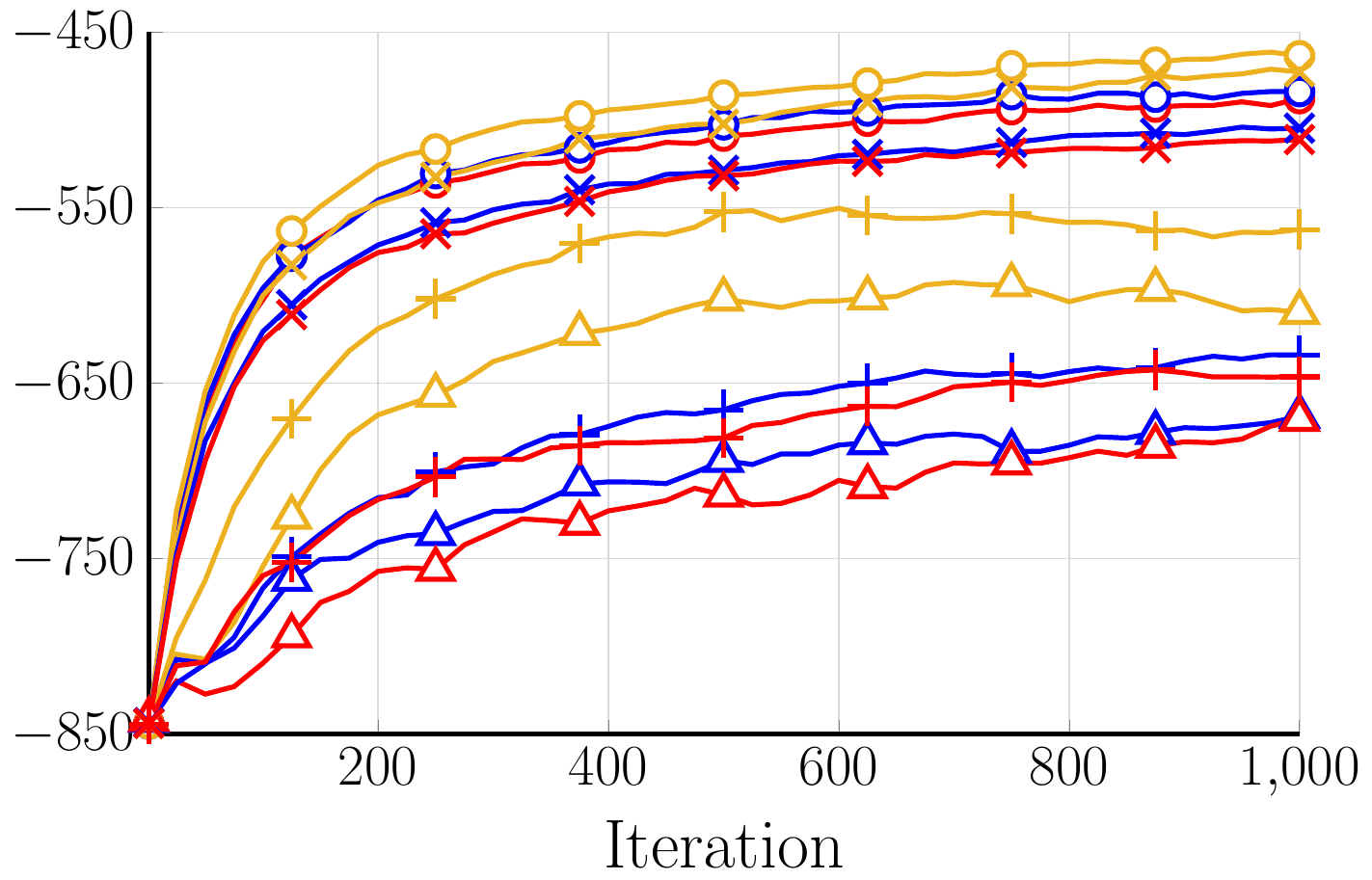}
			\caption{Double-pendulum.\label{fig:pend2_ret}}
		\end{subfigure}
		\\[1em]
		\begin{subfigure}[t]{.49\linewidth}
			\centering
			\includegraphics[width=\textwidth]{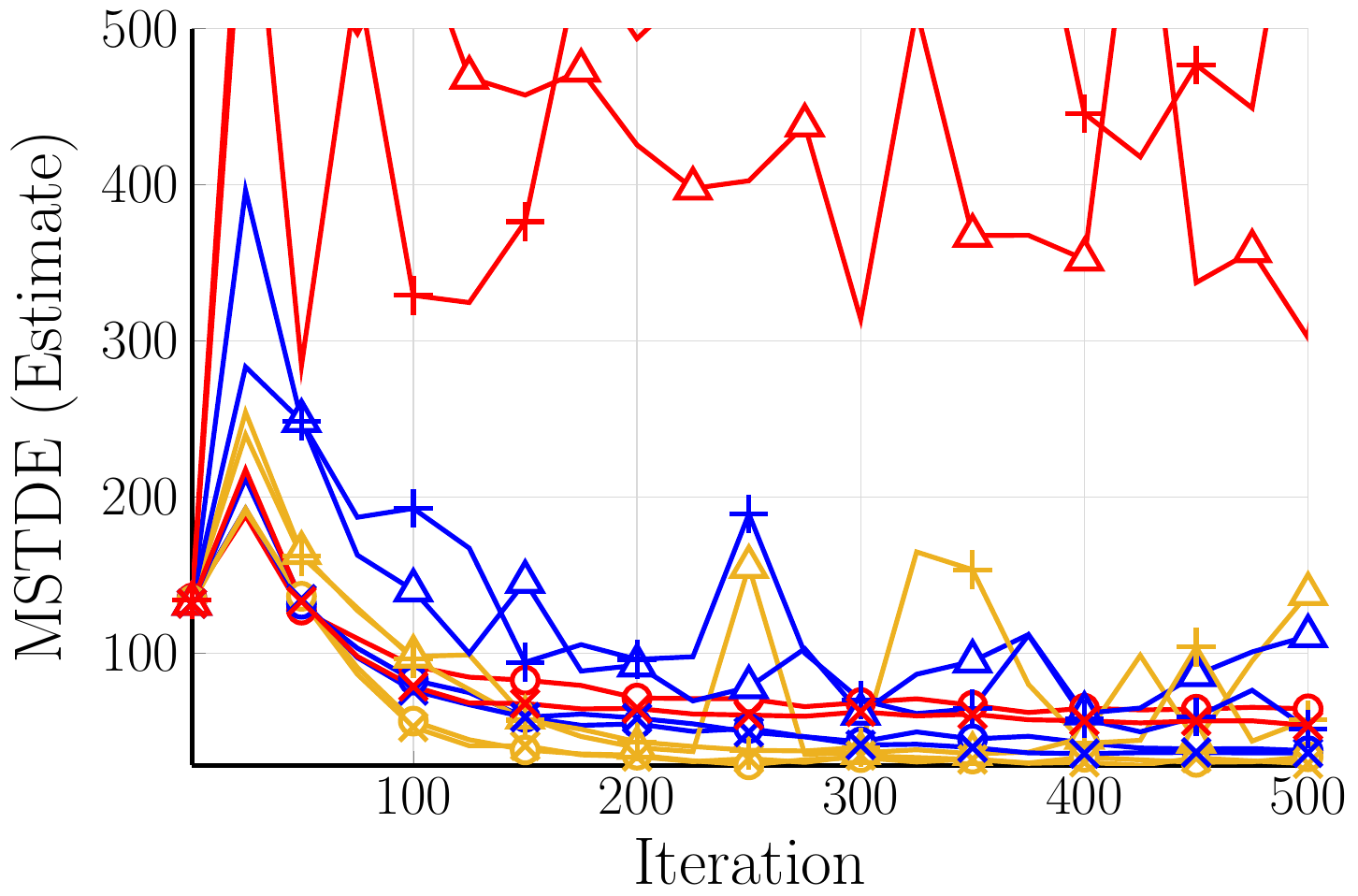}
			\caption{Single-pendulum.\label{fig:pend_td}}
		\end{subfigure}
		\hfill
		\begin{subfigure}[t]{.49\linewidth}
			\centering
			\includegraphics[width=\textwidth]{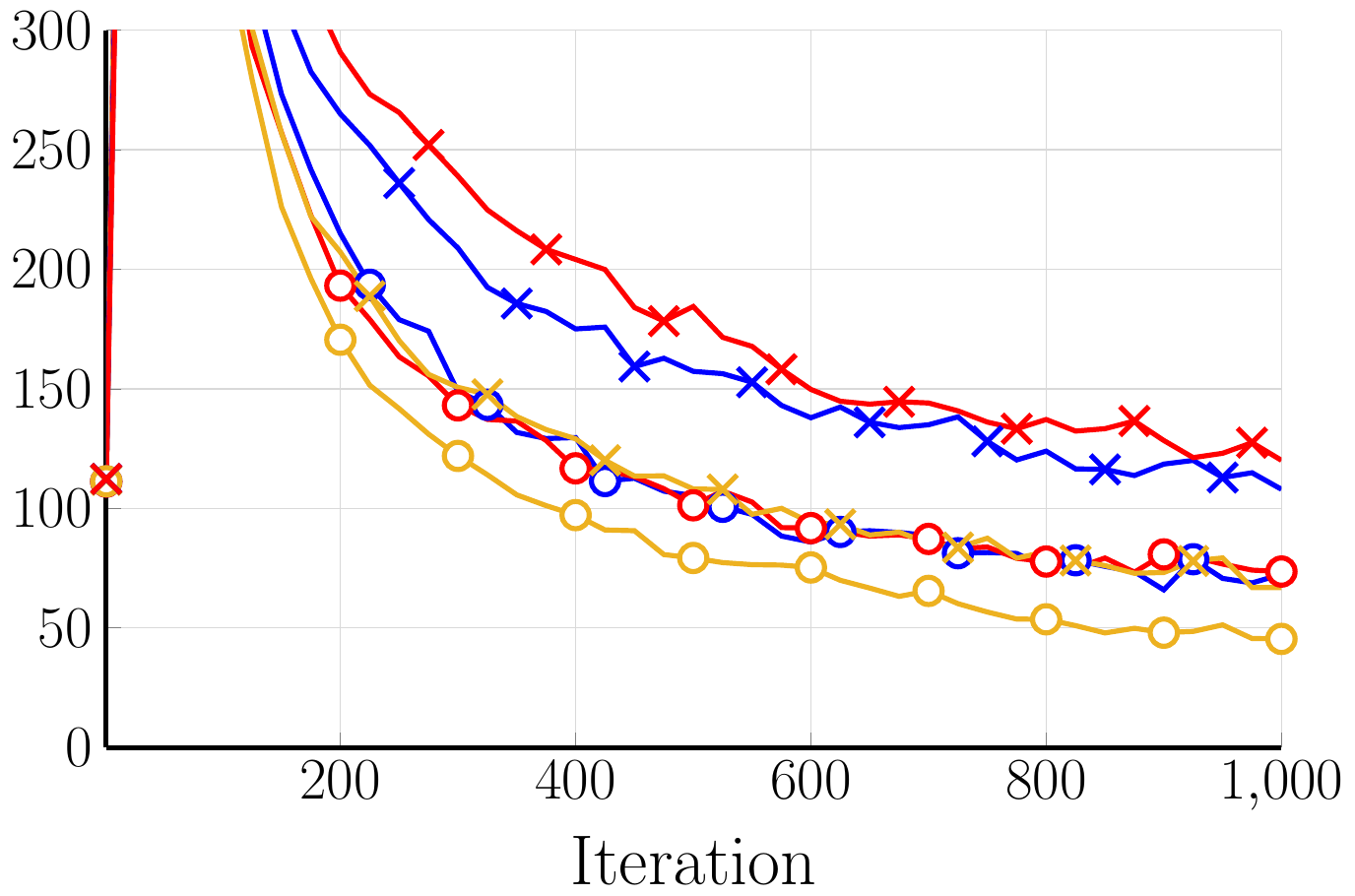}
			\caption{Double-pendulum.\label{fig:pend2_td}}
		\end{subfigure}
		\caption{\label{fig:pend} TRPO results on the pendulum swing-up tasks. In both tasks, GAE-REG + RETR yields the best results. In the single-pendulum, the biggest help is given by GAE-REG, which is the only version always converging to the optimal policy. In the double-pendulum, Retrace is the most important component, as without it all algorithms performed poorly (their TD error is not shown as too large). In all cases, NO-REG always performed worse than TD-REG and GAE-REG (red plots are always below blue and yellow plots with the same markers).}
	\end{minipage}
\end{figure}

In this section, we apply the proposed TD- and GAE-regularization to TRPO and compare to Retrace~\citep{munos2016retrace} and to double-critic learning~\citep{hasselt2010double}, both state-of-the-art techniques to stabilize the learning of the critic.
Similarly to GAE, Retrace replaces the advantage function estimator with the average of $n$-step advantage estimators, but it additionally employs importance sampling to use off-policy data
\begin{align}
	\widehat{A}^\elicoeff(\state_t,\action_t;\paramsQ) &:= (1-\elicoeff)\sum_{i=t}^{T-1} \elicoeff^{i-t} \left(\prod_{j=t}^{T-1}w_j\right) A_{t:i+1} + \elicoeff^{T-t} w_T A_{t:T+1}, \label{eq:retrace}
\end{align}
where the importance sampling ratio $w_j = \min(1, \pi(a_j|s_j;\params) / \beta(a_j|s_j))$ is truncated at 1 to prevent the ``variance explosion'' of the product of importance sampling ratios, and $\beta(a|s)$ is the behavior policy used to collect off-policy data. For example, we can reuse past data collected at the $i$-th iteration by having $\beta(a|s) = \pi(a|s;\params_i)$. 
\\
Double-critic learning, instead, employs two critics to reduce the overestimation bias, as we have seen with TD3 in the LQR task. However, TD3 builds upon DPG and modifies the target policy in the Q-function TD error target, which does not appear in the V-function TD error (compare Eq.~\eqref{eq:td_q} to Eq.~\eqref{eq:td_v}). Therefore, we decided to use the double-critic method proposed by~\citet{hasselt2010double}. In this case, at each iteration only one critic is randomly updated and used to train the policy. For each critic update, the TD targets are computed using estimates from the other critic, in order to reduce the overestimation bias.

In total, we present the results of 12 algorithms, as we tested all combinations of vanilla TRPO (NO-REG), TD-regularization (TD-REG), GAE-regularization (GAE-REG), Retrace (RETR) and double-critic learning (DOUBLE).
All results presented below are averaged over 50 trials. However, for the sake of clarity, in the plots we show only the mean of the expected return.
Both the actor and the critic are linear function with random Fourier features, as presented in~\citep{rajeswaran2017towards}. For the single-pendulum we used 100 features, while for the double-pendulum we used 300 features. In both tasks, we tried to collect as few samples as possible, i.e., 500 for the single-pendulum and 3,000 for the double-pendulum. All algorithms additionally reuse the samples collected in the past four iterations, effectively learning with 2,500 and 15,000 samples, respectively, at each iteration. The advantage is estimated with importance sampling as in Eq.~\eqref{eq:retrace}, but only Retrace uses truncated importance ratios. 
For the single-pendulum, the starting regularization coefficient is $\tdcoeff_0 = 1$. For the double-pendulum, $\tdcoeff_0 = 1$ for GAE-REG and $\tdcoeff_0 = 0.1$ for TD-REG, as the TD error was larger in the latter task (see Figure~\ref{fig:pend2_td}). In both tasks, it then decays according to $\tdcoeff_{t+1} = \kappa\tdcoeff_t$ with $\kappa = 0.999$. Finally, both the advantage and the TD error estimates are standardized for the policy update.
For more details about the tasks and the hyperparameters, we refer to Appendix \ref{app:pendulum}.

Figure~\ref{fig:pend} shows the expected return and the mean squared TD error estimated by the critic at each iteration. In both tasks, the combination of GAE-REG and Retrace performs the best. From Figure \ref{fig:pend_ret}, we can see that GAE-REG is the most important component in the single-pendulum task. First, because only yellow plots converge to the optimal policy. TD-REG also helps, but blue plots did not converge after 500 iterations. Second, because the worst performing versions are the ones without any regularization \textit{but} with Retrace. The reason why Retrace, if used alone, harms the learning can be seen in Figure \ref{fig:pend_td}. Here, the estimated TD error of NO-REG + RETR and of NO-REG + DOUBLE + RETR is rather small, but its poor performance in Figure \ref{fig:pend_ret} hints that the critic is affected by overestimation bias. This is not surprising, considering that Retrace addresses the variance and not the bias of the critic. 
\\
Results for the double-pendulum are similar but Retrace performs better. GAE-REG + RETR is still the best performing version, but this time Retrace is the most important component, given that all versions without Retrace performed poorly. We believe that this is due to the larger number of new samples collected per iteration. 

From this evaluation, we can conclude that the TD- and GAE-regularization are complementary to existing stabilization approaches. In particular, the combination of Retrace and GAE-REG yields very promising results.

\begin{figure}[t]
	\begin{center}
		\bfseries{OpenAI Gym Tasks with MuJoCo Physics - PPO}
	\end{center}
	\begin{minipage}[t]{\textwidth}
		\begin{subfigure}[t]{.5\linewidth}
			\centering
			\includegraphics[width=\textwidth]{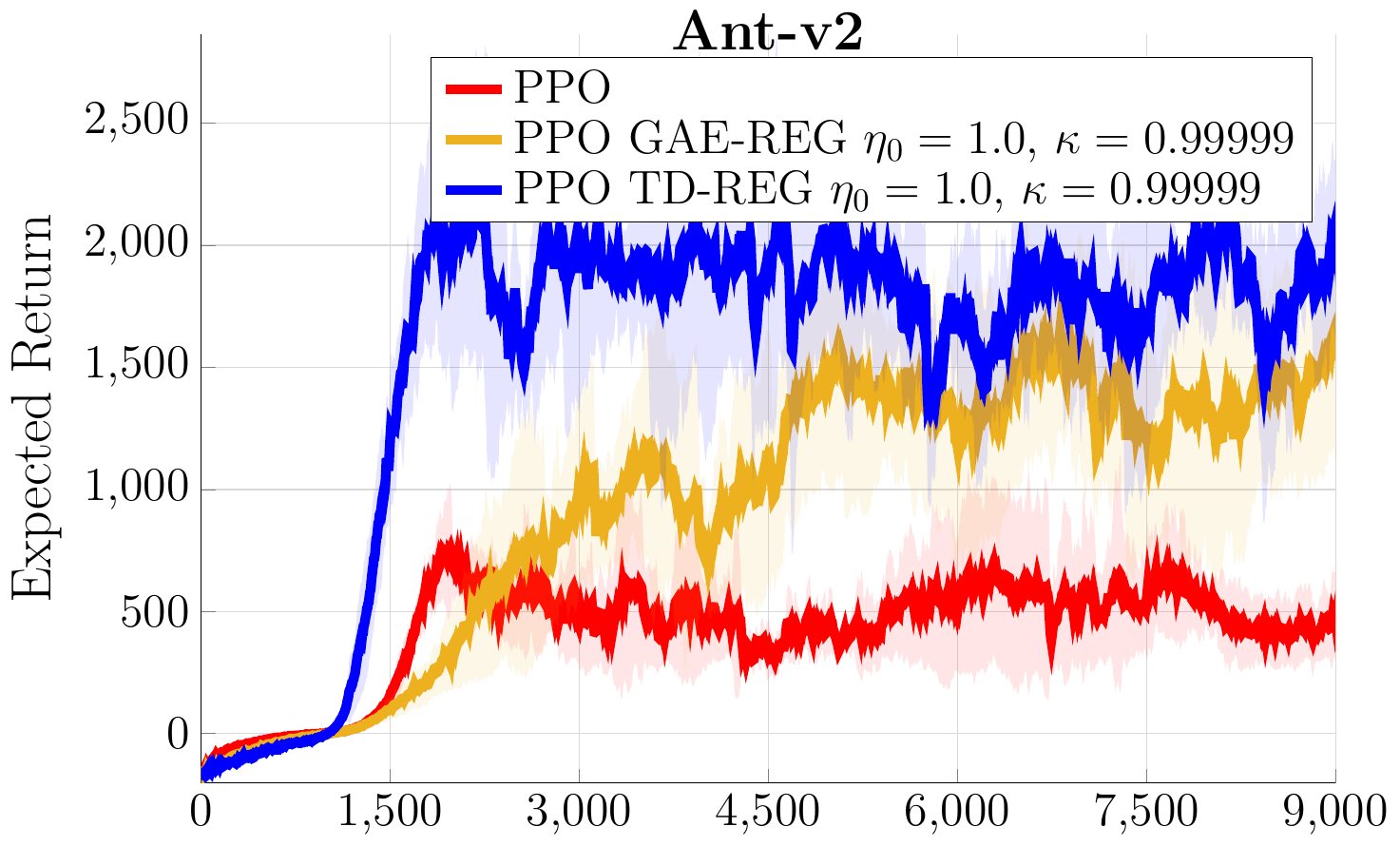}
		\end{subfigure}
		\hfill
		\begin{subfigure}[t]{.49\linewidth}
			\centering
			\includegraphics[width=\textwidth]{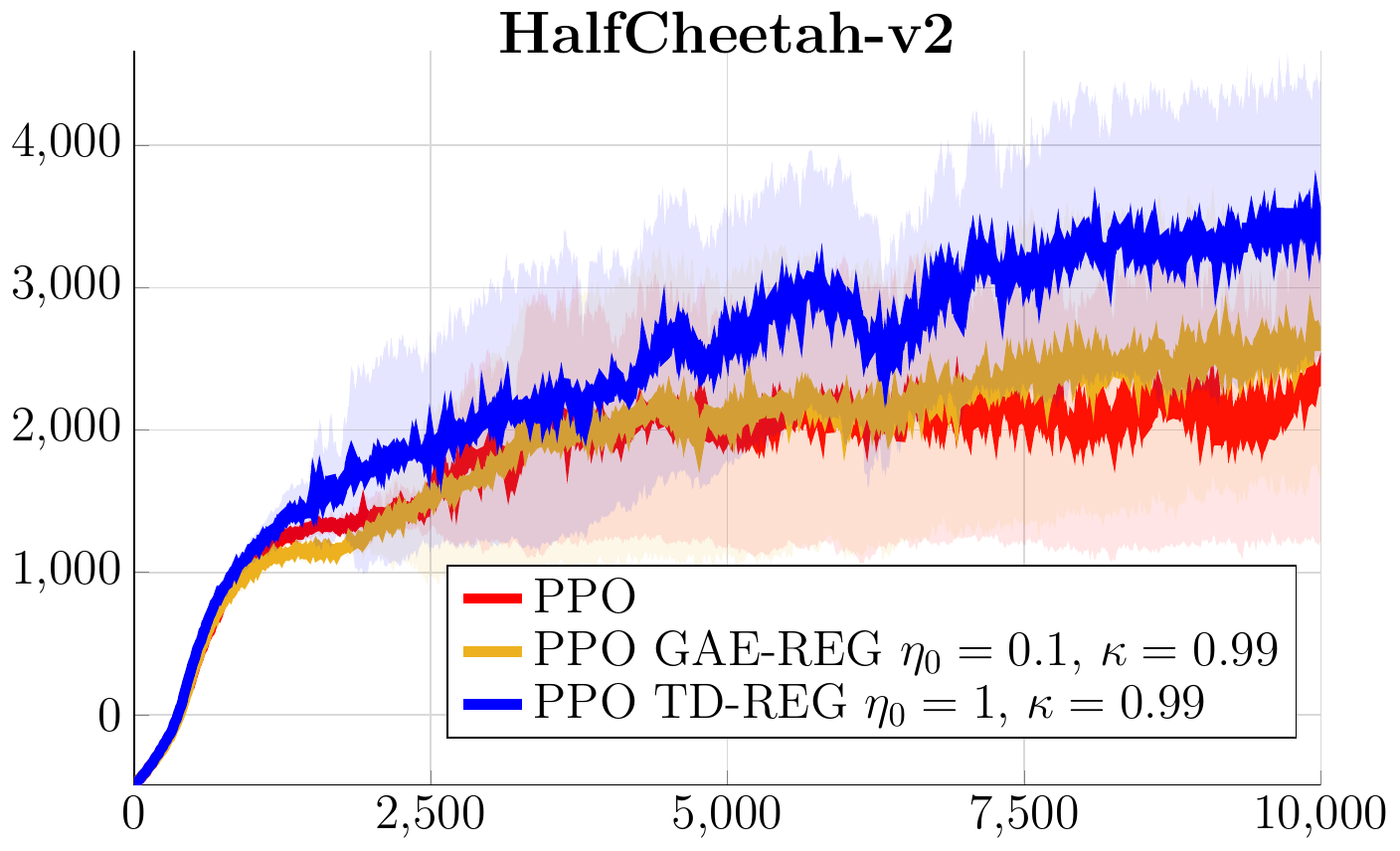}
		\end{subfigure}
		\\
		\begin{subfigure}[t]{.5\linewidth}
			\centering
			\includegraphics[width=\textwidth]{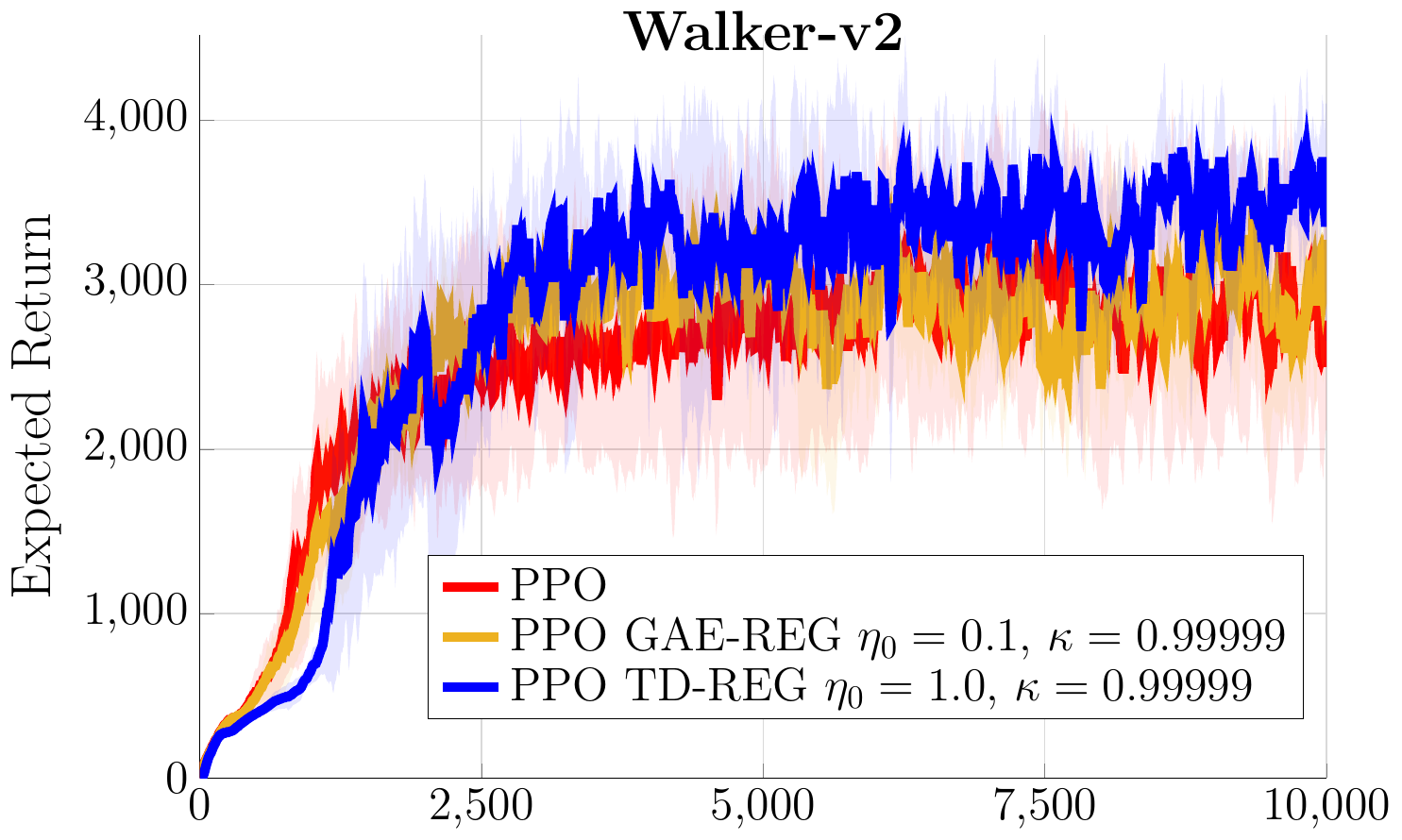}
		\end{subfigure}
		\hfill
		\begin{subfigure}[t]{.49\linewidth}
			\centering
			\includegraphics[width=\textwidth]{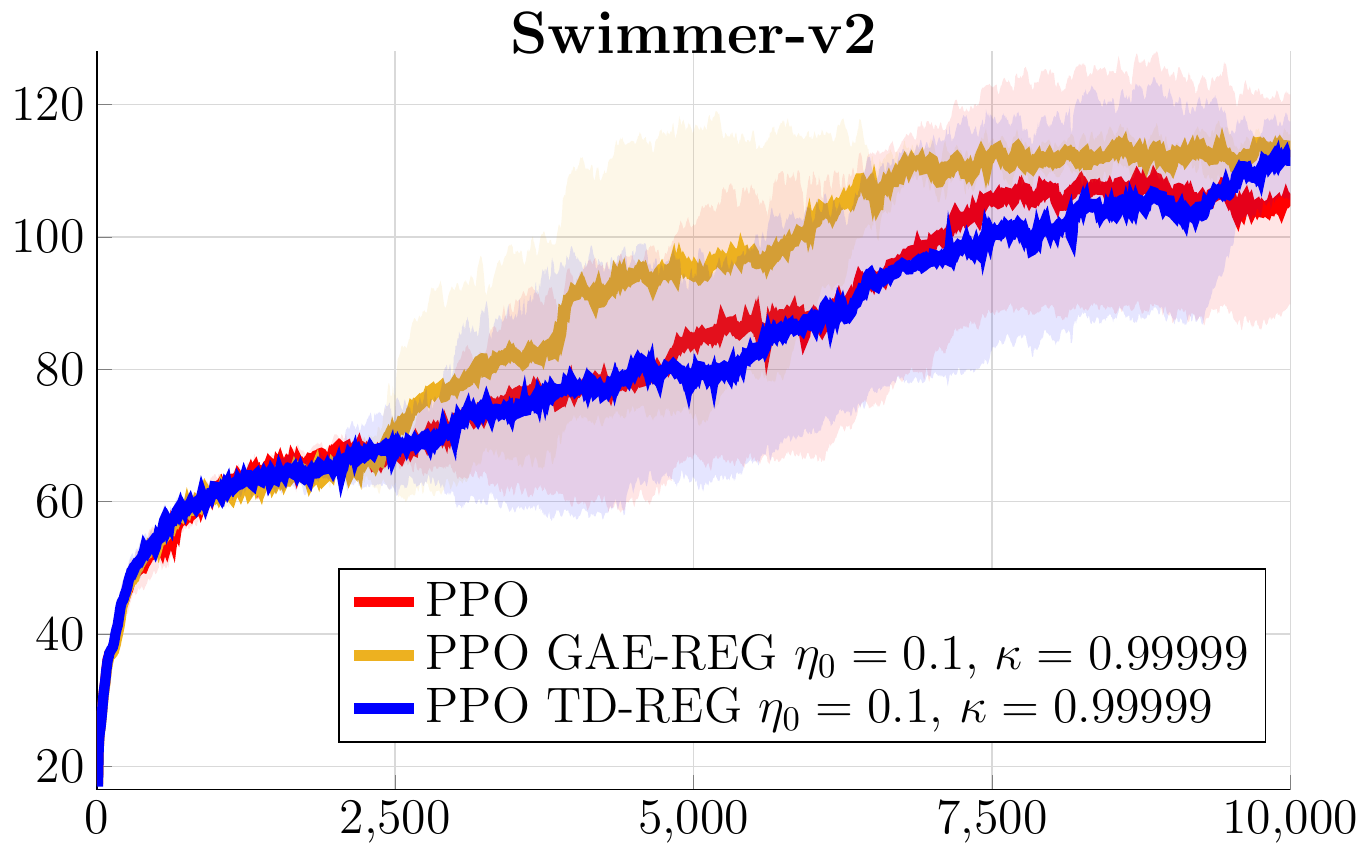}
		\end{subfigure}
		\\
		\begin{subfigure}[t]{.5\linewidth}
			\centering
			\includegraphics[width=\textwidth]{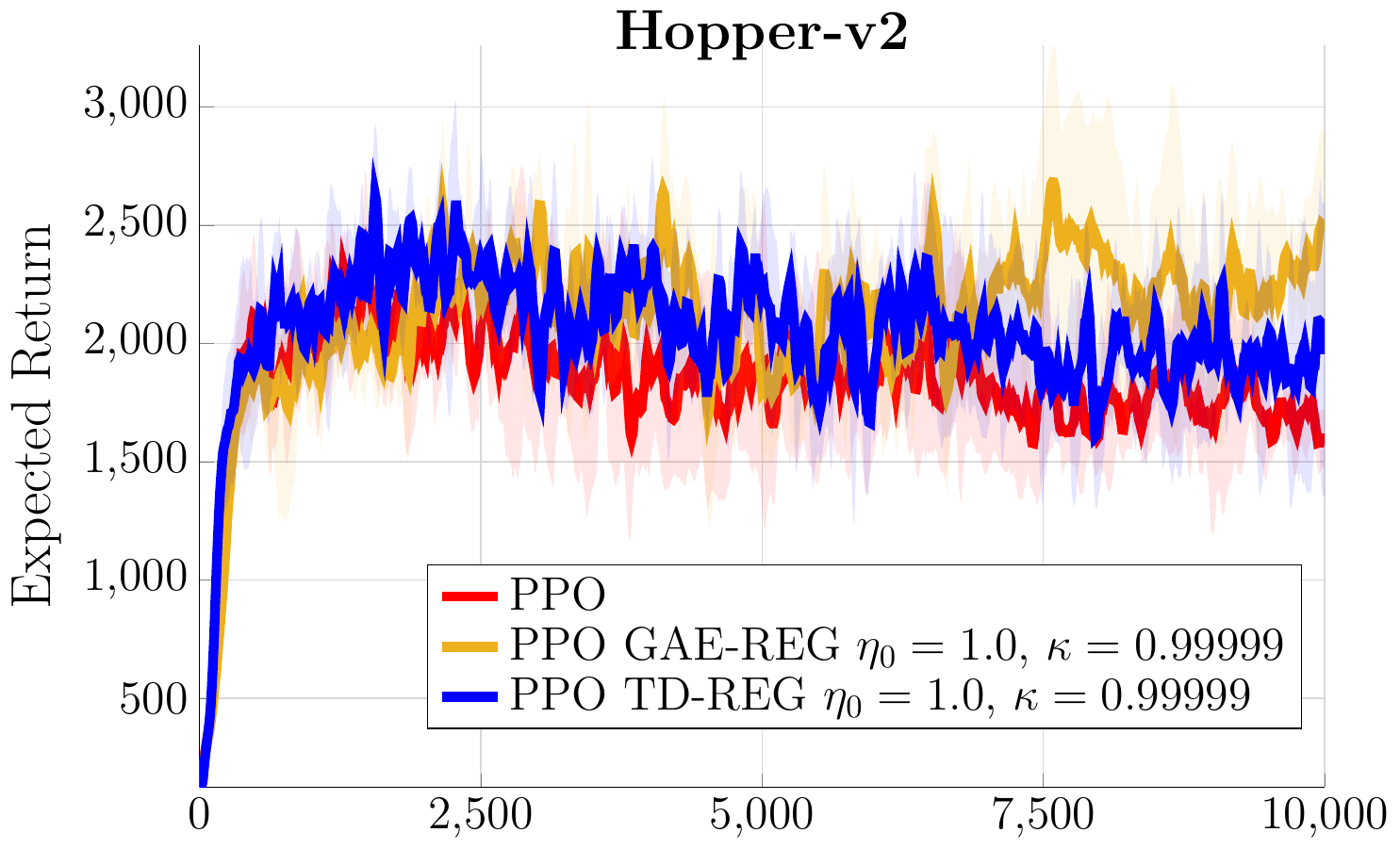}
		\end{subfigure}
		\hfill
		\begin{subfigure}[t]{.49\linewidth}
			\centering
			\includegraphics[width=\textwidth]{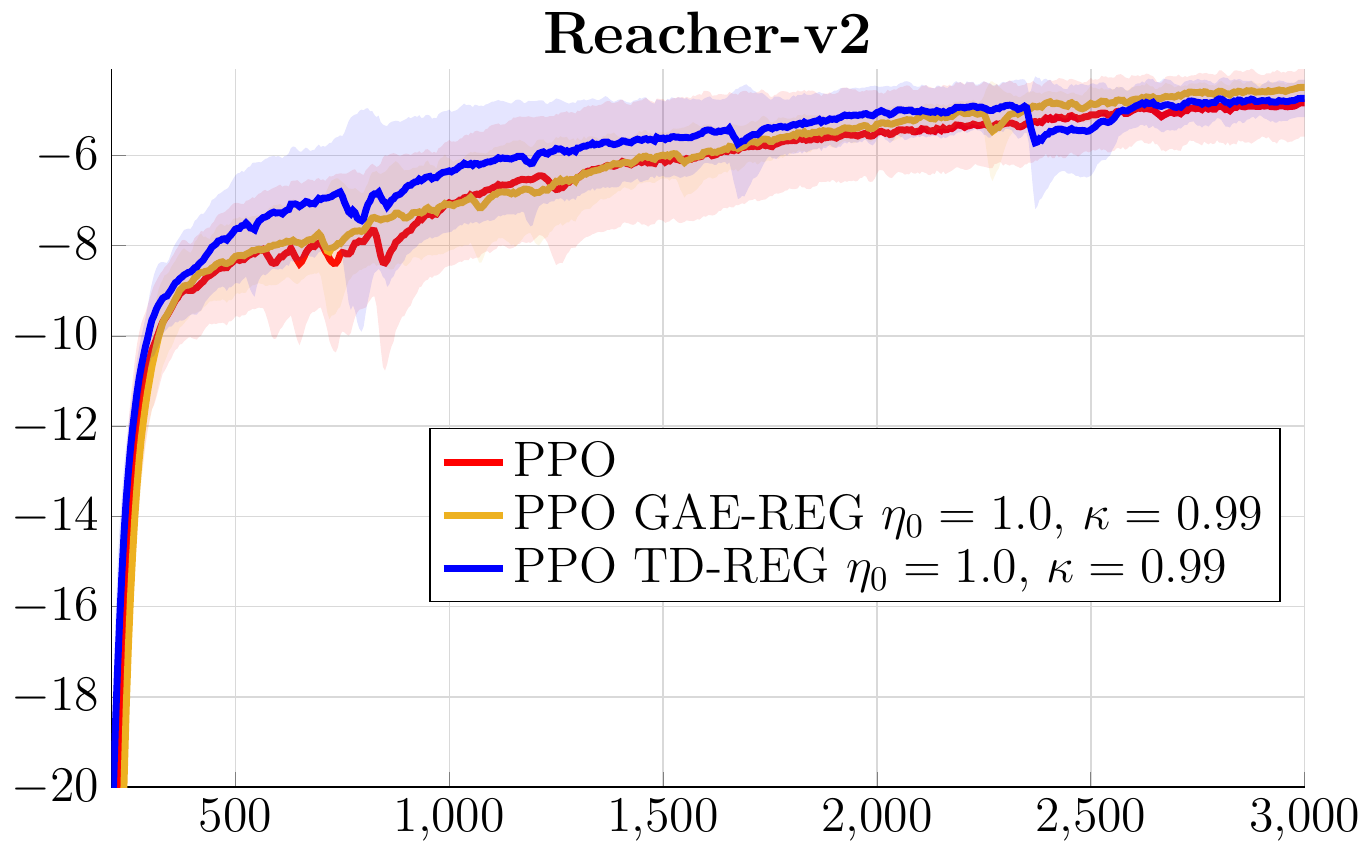}
		\end{subfigure}
		\\
		\begin{subfigure}[t]{.5\linewidth}
			\centering
			\includegraphics[width=\textwidth]{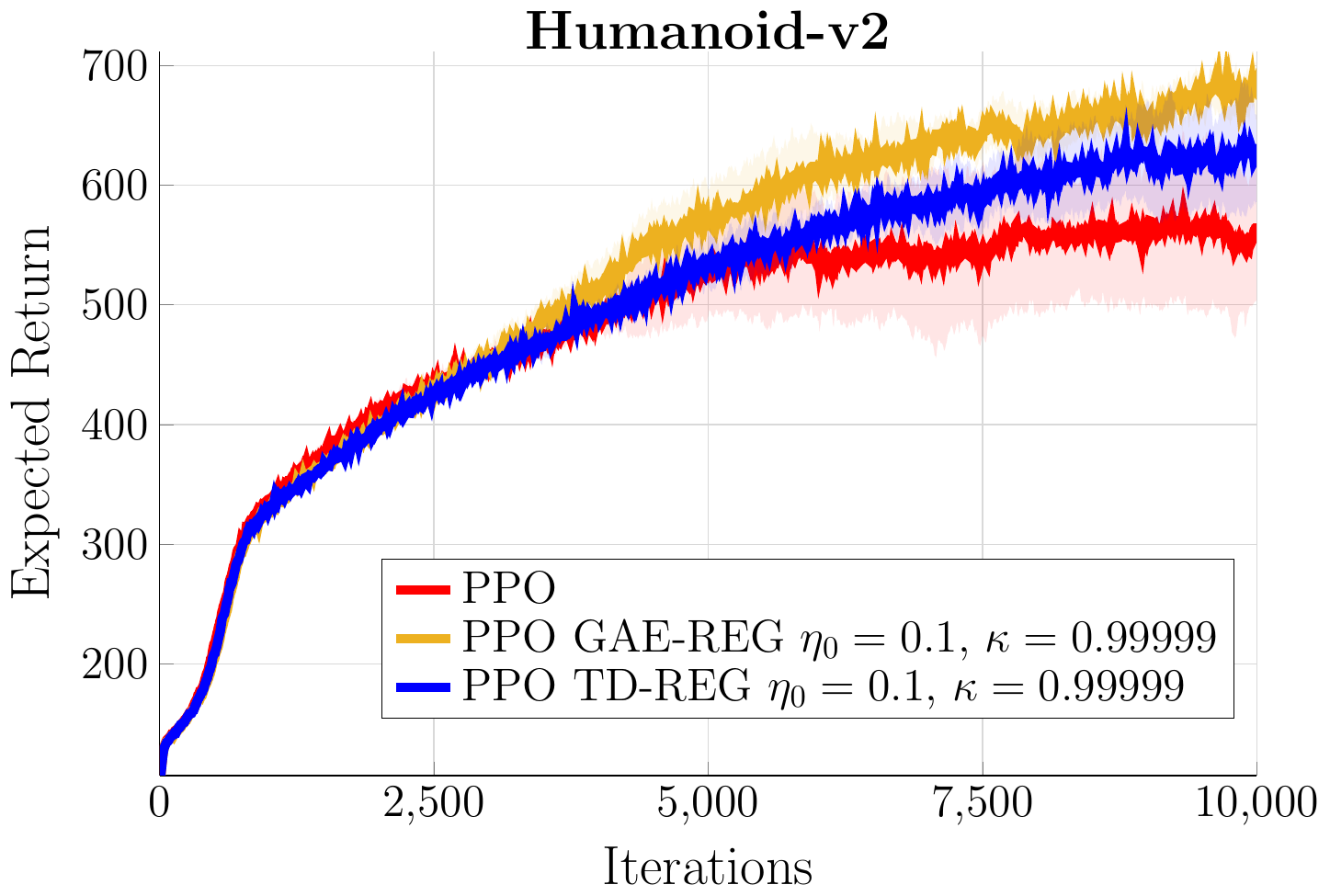}
		\end{subfigure}
		\hfill
		\begin{subfigure}[t]{.49\linewidth}
			\centering
			\includegraphics[width=\textwidth]{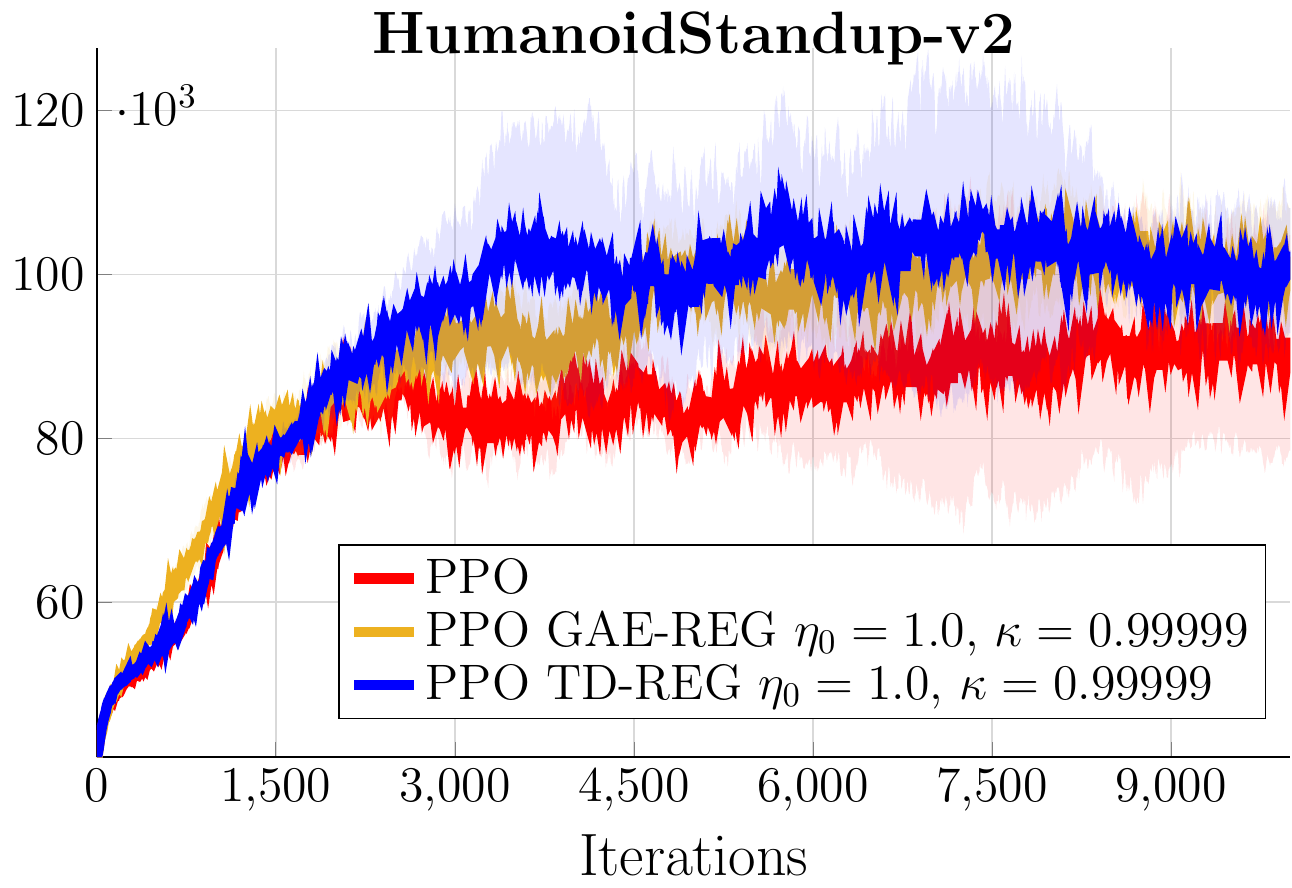}
		\end{subfigure}
		\caption{\label{fig:ppo} Results averaged over five runs, shaded areas denote 95\% confidence interval.}
	\end{minipage}
\end{figure}

\begin{figure}[t]
	\begin{center}
		\bfseries{OpenAI Gym Tasks with MuJoCo Physics - TRPO}
	\end{center}
	\begin{minipage}[t]{\textwidth}
		\begin{subfigure}[t]{.5\linewidth}
			\centering
			\includegraphics[width=\textwidth]{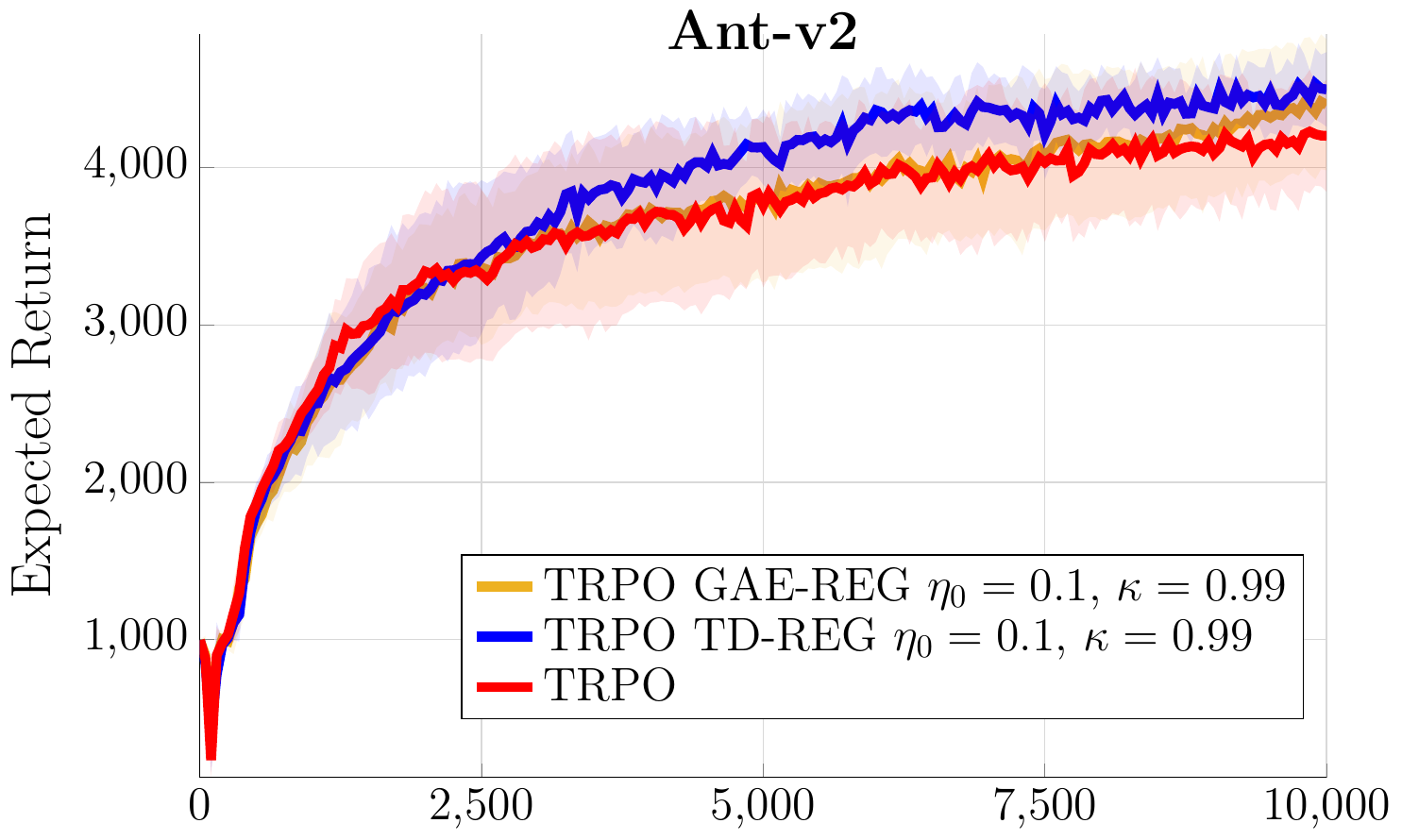}
		\end{subfigure}
		\hfill
		\begin{subfigure}[t]{.49\linewidth}
			\centering
			\includegraphics[width=\textwidth]{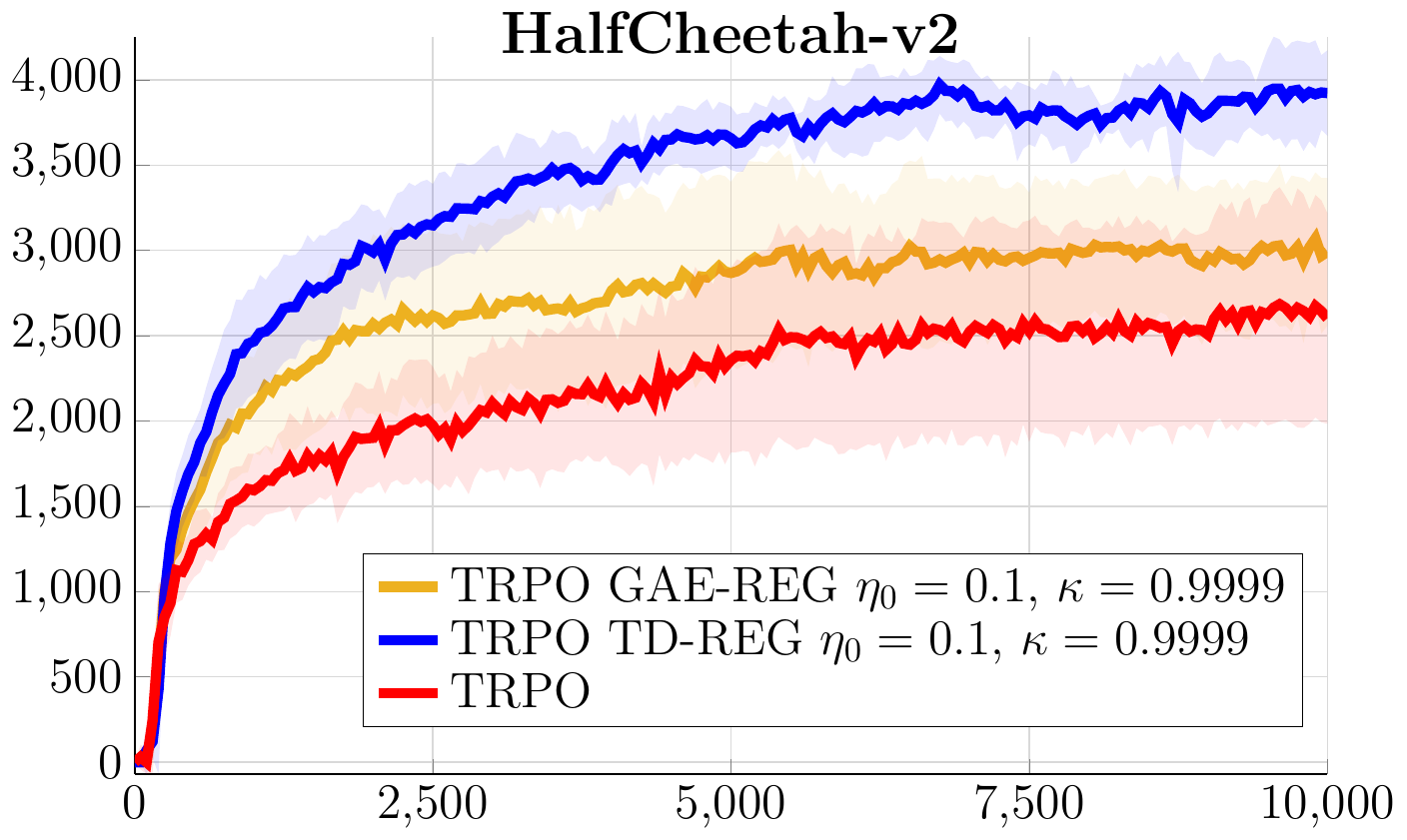}
		\end{subfigure}
		\\
		\begin{subfigure}[t]{.5\linewidth}
			\centering
			\includegraphics[width=\textwidth]{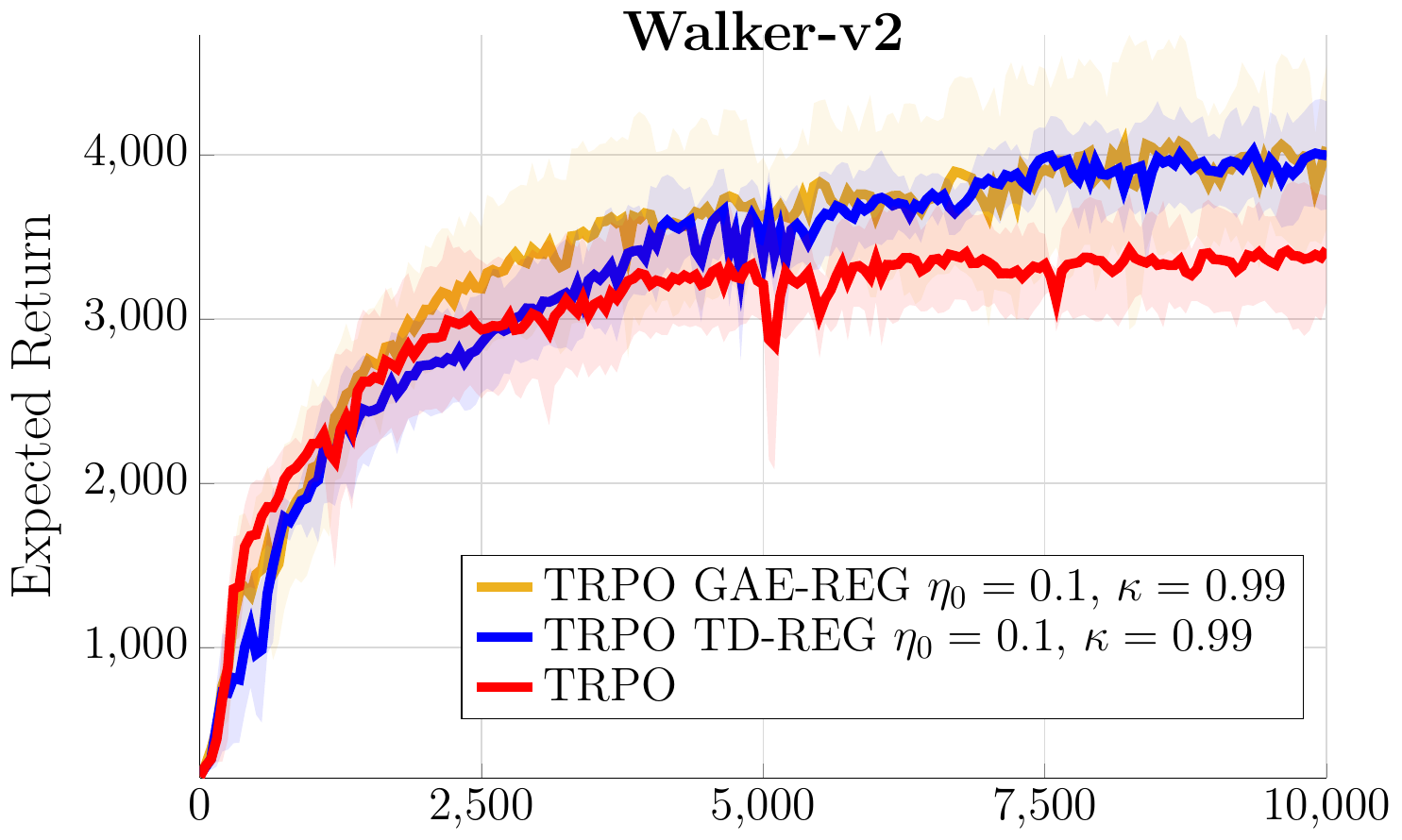}
		\end{subfigure}
		\hfill
		\begin{subfigure}[t]{.49\linewidth}
			\centering
			\includegraphics[width=\textwidth]{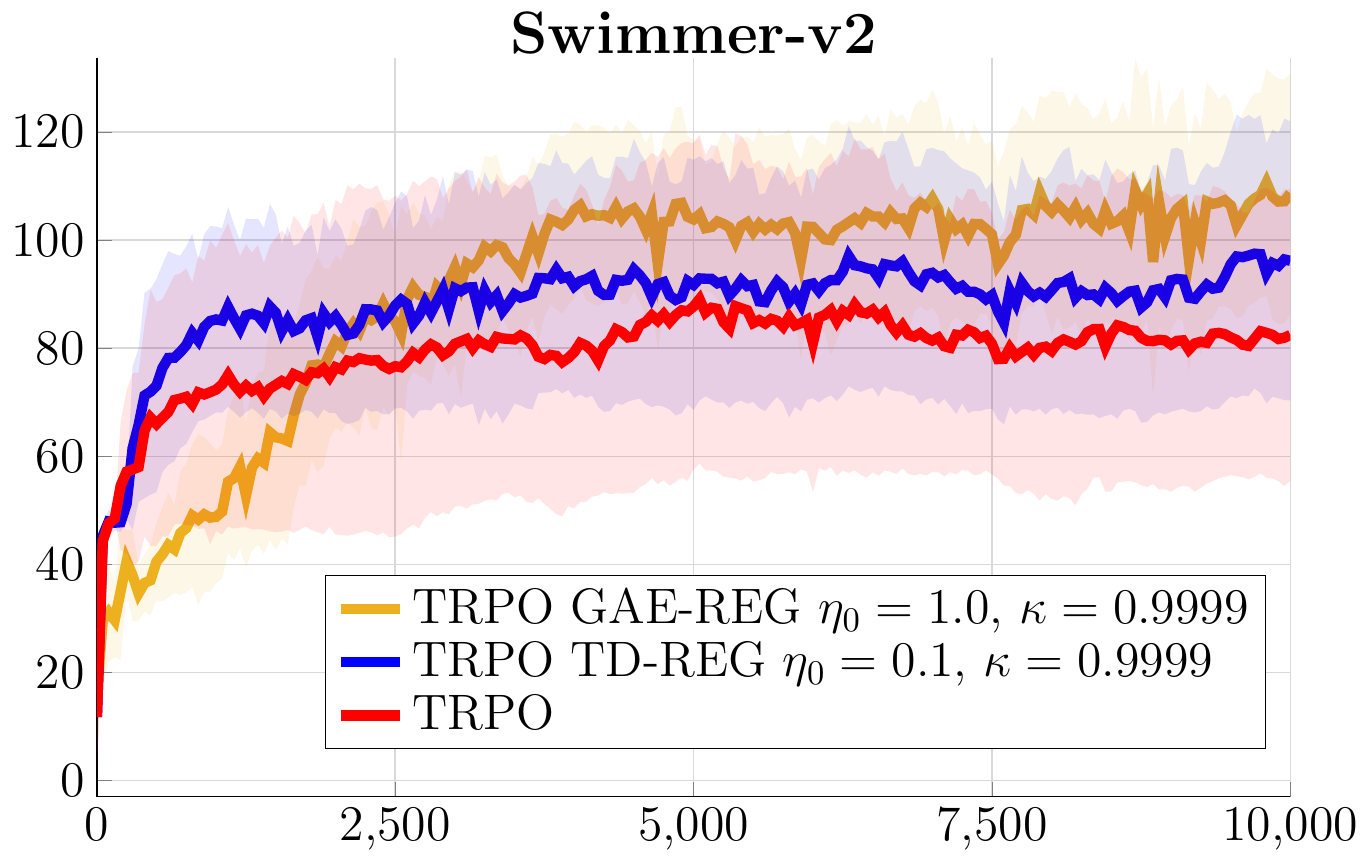}
		\end{subfigure}
		\\
		\begin{subfigure}[t]{.5\linewidth}
			\centering
			\includegraphics[width=\textwidth]{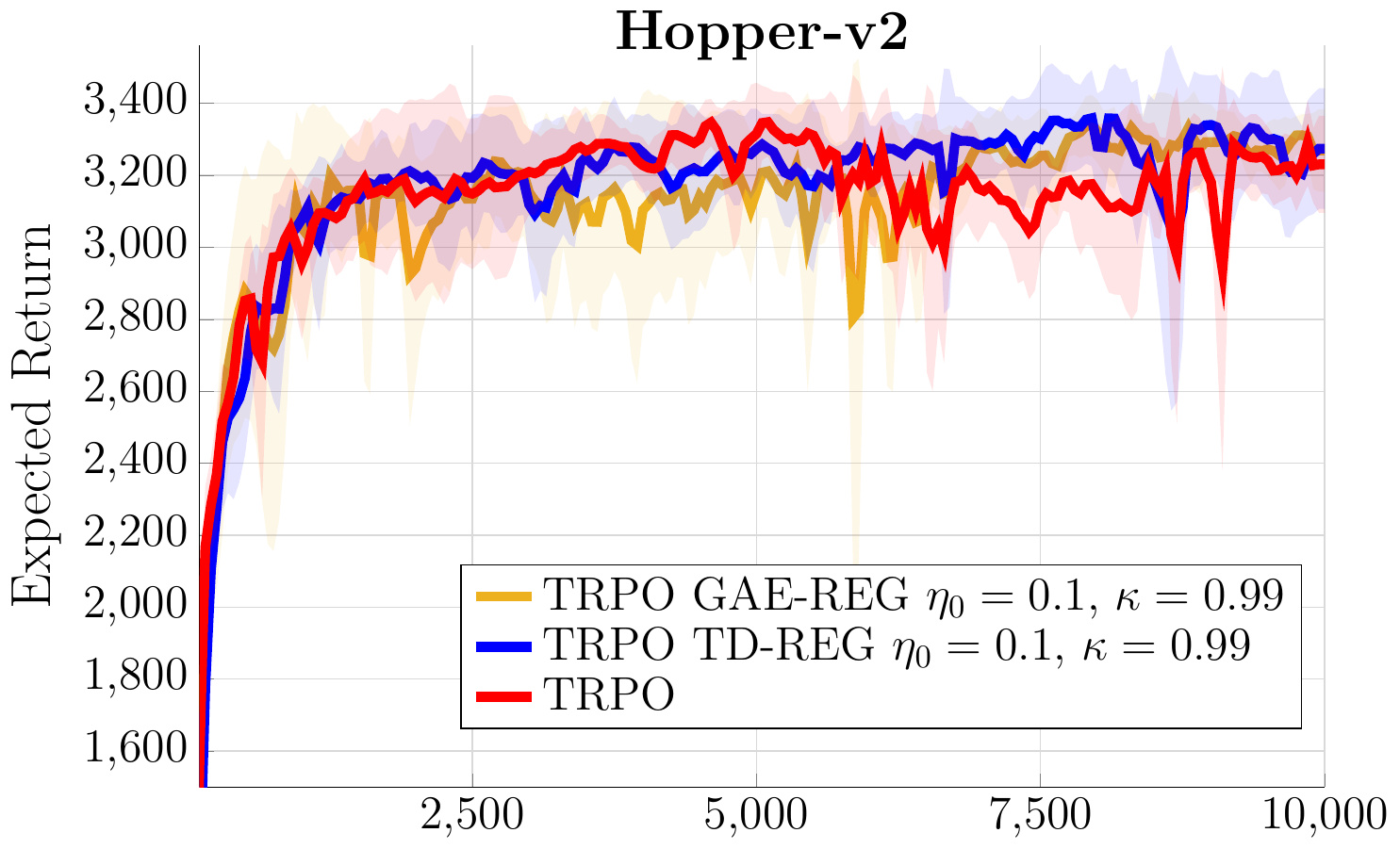}
		\end{subfigure}
		\hfill
		\begin{subfigure}[t]{.49\linewidth}
			\centering
			\includegraphics[width=\textwidth]{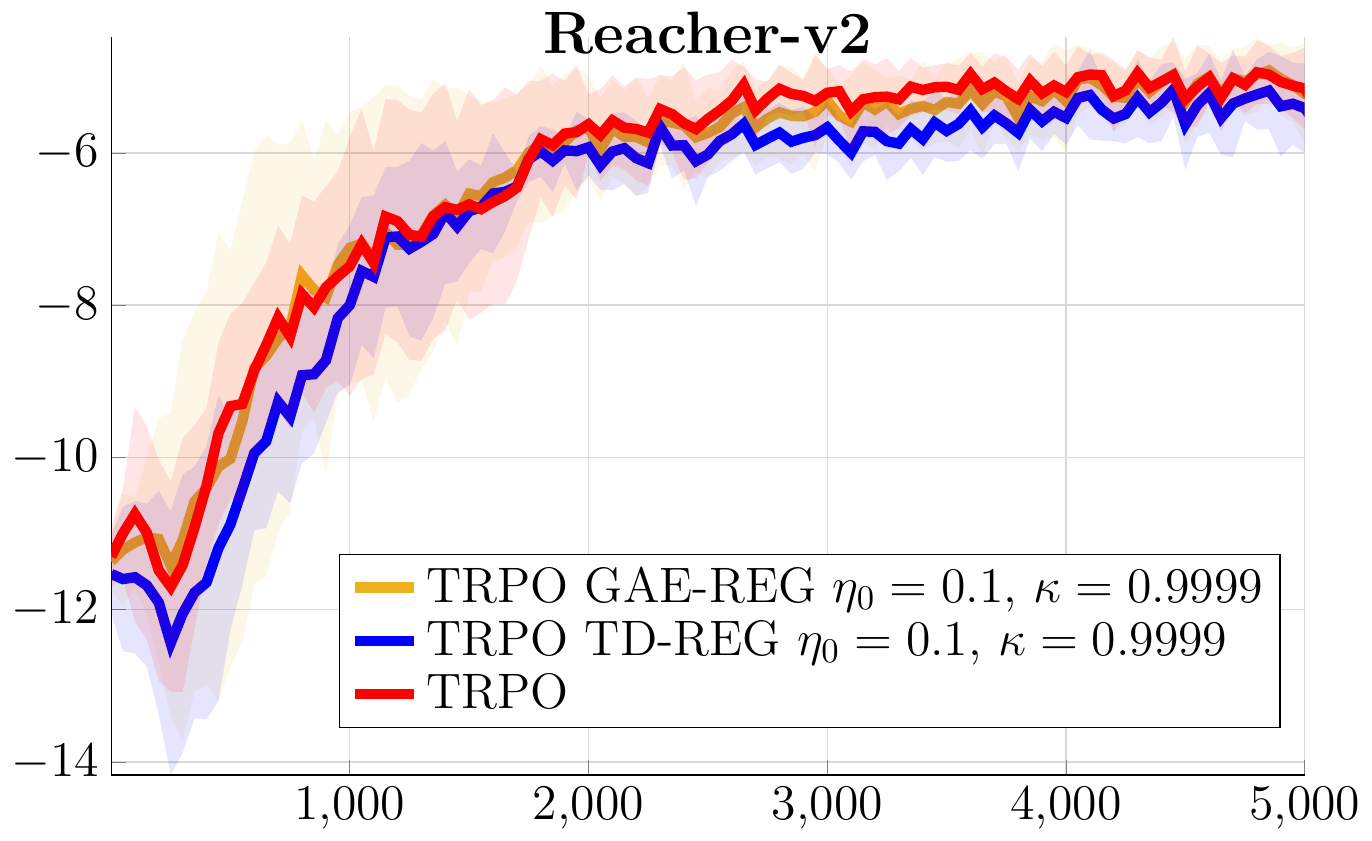}
		\end{subfigure}
		\\
		\begin{subfigure}[t]{.5\linewidth}
			\centering
			\includegraphics[width=\textwidth]{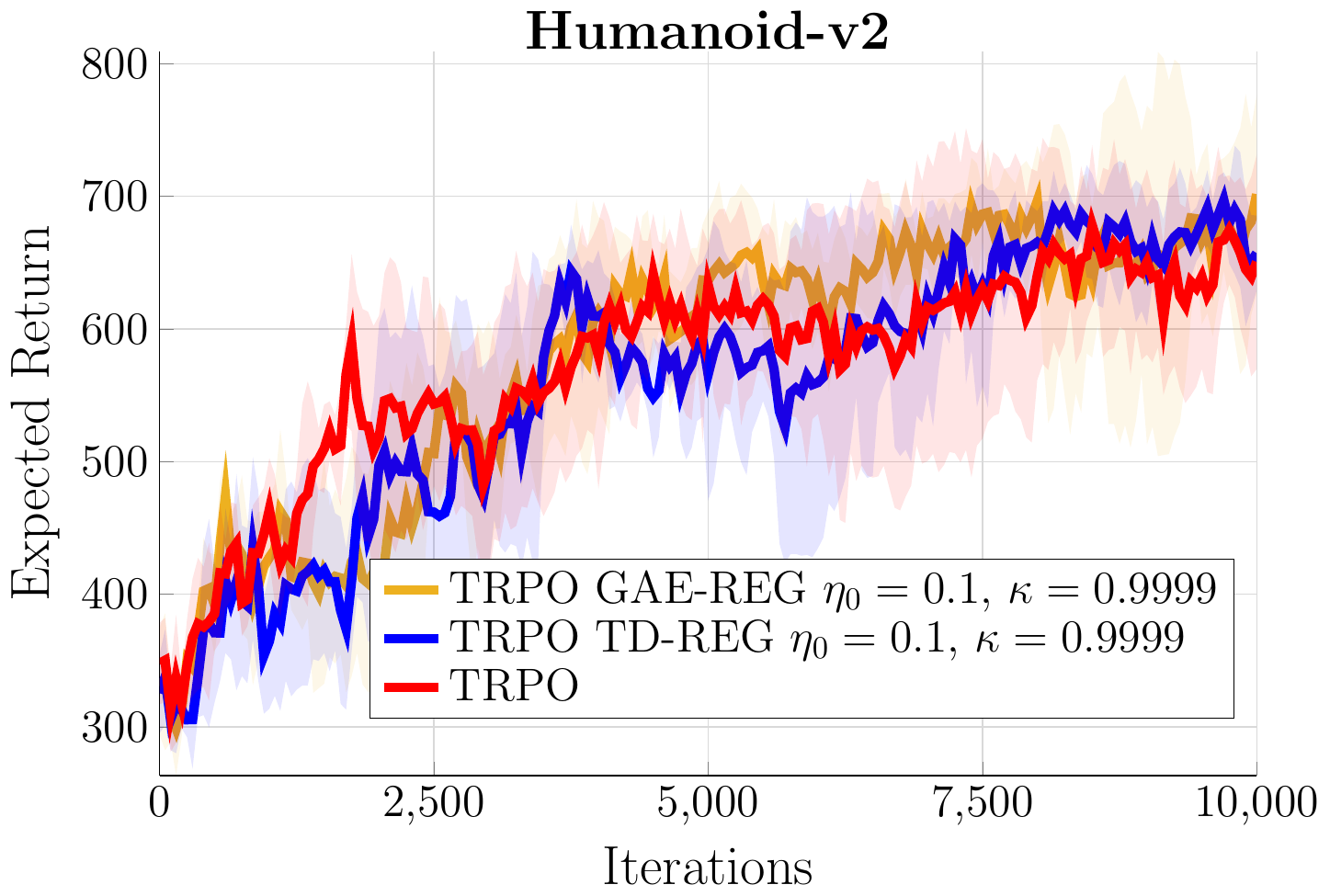}
		\end{subfigure}
		\hfill
		\begin{subfigure}[t]{.49\linewidth}
			\centering
			\includegraphics[width=\textwidth]{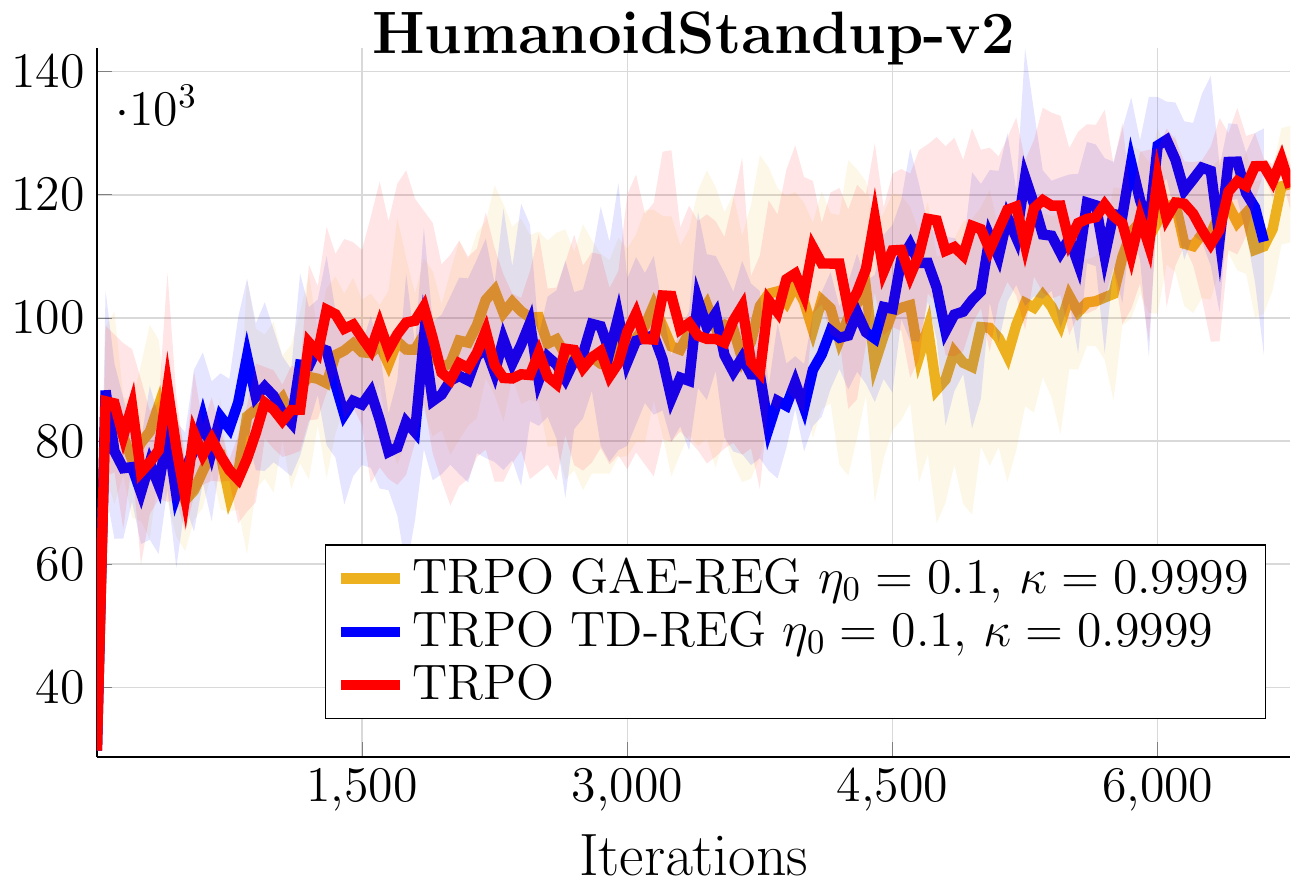}
		\end{subfigure}
		\caption{\label{fig:trpo} Results averaged over five runs, shaded areas denote 95\% confidence interval.}
	\end{minipage}
\end{figure}

\subsection{MuJoCo Continuous Control Tasks}
We perform continuous control experiments using OpenAI Gym~\citep{brockman2016openai} with the MuJoCo physics simulator \citep{todorov2012mujoco}. 
For all algorithms, the advantage function is estimated by GAE.
For TRPO, we consider a deep RL setting where the actor and the critic are two-layer neural networks with 128 hyperbolic tangent units in each layer. For PPO, the units are 64 in each layer. In both algorithms, both the actor and the critic gradients are optimized by ADAM~\citep{kingma2014adam}. For the policy update, both the advantage and the TD error estimates are standardized.
More details of the hyperparameters are given in Appendix \ref{app:mujoco}.

From the evaluation on the pendulum tasks, it emerged that the value of the regularization coefficient $\tdcoeff$ strongly depends on the magnitude of the advantage estimator and the TD error. In fact, for TD-REG we had to decrease $\tdcoeff_0$ from 1 to 0.1 in the double-pendulum task because the TD error was larger. 
For this reason, for both PPO and TRPO we tested different initial regularization coefficients $\tdcoeff_0$ and decay factors $\kappa$, choosing among all combinations of $\tdcoeff_0 = 0.1, 1$ and $\kappa = 0.99, 0.9999$.

Figure~\ref{fig:ppo} shows the expected return against training iteration for PPO. On Ant, HalfCheetahm, Walker2d, Humanoid and HumanoidStandup both TD-REG and GAE-REG performed substantially better, especially on Ant-v2, where PPO performed very poorly. 
On Swimmer and Hopper, TD-REG and GAE-REG also outperformed vanilla PPO, but the improvement was less substantial. On Reacher all algorithms performed the same. This behavior is expected, since Reacher is the easiest of MuJoCo tasks, followed by Swimmer and Hopper.
On Ant and Walker, we also notice the ``slow start'' of TD-regularized algorithms already experienced in the LQR. For the first 1,000 iterations, in fact, vanilla PPO expected return increased faster than PPO TD-REG and PPO GAE-REG, but then it also plateaued earlier.

Results for TRPO (Figure~\ref{fig:trpo}) are the same except for Humanoid and HumanoidStandup. TD-REG and GAE-REG always outperformed or performed as well as TRPO, and the ranking of the algorithm (first, second and third best performing) is the same of Figure~\ref{fig:ppo}. On Ant, HalfCheetah and Walker2d, GAE-REG performed better than TD-REG, while on Swimmer GAE-REG surpassed TD-REG.
On Hopper, Reacher, Humanoid, and HumanoidStandup, all algorithms performed the same. 

It is also interesting to notice that both GAE-REG and TD-REG shared the best performing $\tdcoeff_0$ and $\kappa$. For instance, with PPO on Ant they both performed best with $\tdcoeff_0 = 1.0$ and $\kappa = 0.99999$. Only on HalfCheetah (for PPO) and Swimmer (for TRPO) we had to use different $\tdcoeff_0$ and $\kappa$ between GAE-REG and TD-REG. On the contrary, the same values do not work for both PPO and TRPO, as for instance with TRPO on Ant the best performing values were $\tdcoeff_0 = 0.1$ and $\kappa = 0.99$.

From this evaluation it emerged that both TD-REG and GAE-REG can substantially improve the performance of TRPO and PPO. However, as for any other method based on regularization, the tuning of the regularization coefficient is essential for their success.

\clearpage

\section{Conclusion}
Actor-critic methods often suffer from instability. A major cause is the function approximation error in the critic. 
In this paper, we addressed the stability issue taking into account the relationship between the critic and the actor.
We presented a TD-regularized approach penalizing the actor for breaking the critic Bellman equation, in order to perform policy updates producing small changes in the critic.
We presented practical implementations of our approach and combined it together with existing methods stabilizing the critic. Through evaluation on benchmark tasks, we showed that our TD-regularization is complementary to already successful methods, such as Retrace, and allows for more stable updates, resulting in policy updates that are less likely to diverge and improve faster.

Our method opens several avenues of research.
In this paper, we only focused on direct TD methods. In future work, we will consider the Bellman-constrained optimization problem and extend the regularization to residual methods, as they have stronger convergence guarantees even when nonlinear function approximation is used to learn the critic \citep{baird1995residual}. 
We will also study equivalent formulations of the constrained problem with stronger guarantees. For instance, the approximation of the integral introduced by the expectation over the Bellman equation constraint could be addressed by using the representation theorem.
Furthermore, we will also investigate different techniques to solve the constrained optimization problem. For instance, we could introduce slack variables or use different penalty functions. Another improvement could address techniques for automatically tuning the coefficient $\tdcoeff$, which is crucial for the success of TD-regularized algorithms, as emerged from the empirical evaluation.
Finally, it would be interesting to study the convergence of actor-critic methods with TD-regularization, including cases with tabular and compatible function approximation, where convergence guarantees are available.


\bibliographystyle{plainnat}
\bibliography{my_bib}

\clearpage

\appendix 

\setlength{\belowcaptionskip}{5pt}
\captionsetup[subfigure]{aboveskip=5pt}
\captionsetup[subfigure]{belowskip=0pt}

\section{2D Linear-Quadratic Regulator Experiments}
\label{app:lqr}
The LQR problem is defined by the following discrete-time dynamics
\begin{equation*}
	\state' = A \state +B \action + \gaussian(0,0.1^2) ,\quad\quad\quad\quad
	a = K\state, \quad\quad\quad\quad
	\rmodel = -\state^{\T} X\state - {\action}^{\T} Y \action,
\end{equation*}
where $A,B,X,Y \in \realspace^{d \times d}$, $X$ is a symmetric positive semidefinite matrix, $Y$ is a symmetric positive definite matrix, and $K \in \realspace^{d \times d}$ is the control matrix. The policy parameters we want to learn are $\params = \text{vec}(K)$.
For simplicity, dynamics are not coupled, i.e., $A$ and $B$ are identity matrices, and both $X$ and $Y$ are identity matrices as well.

Although it may look simple, the LQR presents some challenges. First, the system can become easily unstable, as the control matrix $K$ has to be such that the matrix $(A+BK)$ has eigenvalues of magnitude smaller than one. Therefore, policy updates cannot be too large, in order to prevent divergence.
Second, the reward is unbounded and the expected return can be very large, especially at the beginning with an initial random policy. As a consequence, the initial TD error can be very large as well. Third, states and actions are unbounded and cannot be normalized in [0,1], a common practice in RL.

However, we can compute in closed form both the expected return and the Q-function, being able to easily assess the quality of the evaluated algorithms. More specifically, the Q-function is quadratic in the state and in the action, i.e.,
\begin{equation*}
	Q^\pi(\state,\action) = Q_0 + \state^{\T} Q_{\state\state}\state + \action^{\T} Q_{\action\action}\action + \state^{\T} Q_{\state\action}\action,
\end{equation*}
where $Q_0, Q_{\state\state}, Q_{\action\action}, Q_{\state\action}$ are matrices computed in closed form given the MDP characteristics and the control matrix $K$. It should be noted that the linear terms are all zero.

In the evaluation below, we use a 2-dimensional LQR, resulting in four policy parameters. The Q-function is approximated by $\widehat Q(\state,\action;\params) = \phi(\state,\action)^{\T} \paramsQ$, where $\phi(\state,\action)$ are features. We evaluate two different features: polynomial of second degree (\textit{quadratic features}) and polynomial of third degree (\textit{cubic features}). We know that the true Q-function is quadratic without linear features, therefore quadratic features are sufficient. By contrast, cubic features could overfit. Furthermore, quadratic features result in 15 parameters $\paramsQ$ to learn, while the cubic one has 35.

In our experiments, the initial state is uniformly drawn in the interval $[-10,10]$.
The Q-function parameters are initialized uniformly in $[-1,1]$. The control matrix is initialized as $K = -K_0^{\T} K_0$ to enforce negative semidefiniteness, where $K_0$ is drawn uniformly in [-0.5, -0.1].

Along with the expected return, we show the trend of two mean squared TD errors (MSTDE) of the critic $\widehat Q(\state,\action;\paramsQ)$: one is estimated using the TD error, the other is computed in closed form using the true $Q^\pi(\state,\action)$ defined above. It should be noticed that $Q^\pi(\state,\action)$ is not the optimal Q-function (i.e., of the optimal policy), but the true Q-function with respect to the current policy.
We also show the learning of the diagonal entries of $K$ in the policy parameter space. These parameters, in fact, are the most relevant because the optimal $K$ is diagonal as well, due to the reward and transition functions characteristics ($A=B=X=Y=I$).

All results are averaged over 50 trials. In all trials, the random seed is fixed and the initial parameters are the same (all random).
In expected return plots, we bounded the expected return to $-10^3$ and the MSTDE to $3\mathord{\cdot}10^5$ for the sake of clarity, since in the case of unstable policies (i.e., when the matrix $(A+BK)$ has eigenvalues of magnitude greater than or equal to one) the expected return and the TD error are $-\infty$.

\subsection{DPG Evaluation on the LQR}
\label{app:lqr_dpg}
In this section, we evaluate five versions of deterministic policy gradient~\citep{silver2014deterministic}. In the first three, the learning of the critic happens in the usual actor-critic fashion. The Q-function is learned independently from the policy and a target Q-function of parameters $\bar{\paramsQ}$, assumed to be independent from the critic, is used to improve stability, i.e.,
\begin{align}
	\delta_Q(\state,\action,\state';\params, \paramsQ, \textcolor{red}{\bar{\paramsQ}}) = r + \gamma{\widehat Q(\state',\pi(\state';\params); \textcolor{red}{\bar{\paramsQ}})} - \widehat Q(\state,\action;\paramsQ). \label{eq:td_err}
\end{align}
Under this assumption, the critic is updated by following the SARSA gradient
\begin{equation}
\medmuskip=0.99mu
\thinmuskip=0.99mu
	\nabla_\paramsQ \EVV{\substack{\mu_\beta(\state),\beta(\action|\state),\prob(\state'|\state,\action)}}{ \delta_Q(\state,\action,\state';\params,\paramsQ,\bar{\paramsQ})^2} = \EVV{\substack{\mu_\beta(\state),\beta(\action|\state), \prob(\state'|\state,\action)}}{\delta_Q(\state,\action,\state';\params, \paramsQ, \bar{\paramsQ}) \nabla_\paramsQ \widehat Q(\state,\action;\paramsQ)},
\end{equation}
where $\beta(a|s)$ is the behavior policy used to collect samples. 
In practice, $\bar{\paramsQ}$ is a copy of $\paramsQ$. We also tried a soft update, i.e., $\bar{\paramsQ}_{t+1} = \tau_\paramsQ\paramsQ_t + (1-\tau_\paramsQ)\bar{\paramsQ}_t$, with $\tau_\paramsQ \in (0,1]$, as in DDPG~\citep{lillicrap2015continuous}, the deep version of DPG. However, the performance of the algorithms decreased (TD-regularized DPG still outperformed vanilla DPG). We believe that, since for the LQR we do not approximate the Q-function with a deep network, the soft update just restrains the convergence of the critic.

These three versions of DPG differ in the policy update. The first algorithm (\textbf{DPG}) additionally uses a target actor of parameters $\bar{\params}$ for computing the Q-function targets, i.e., 
\begin{align}
	\delta_Q(\state,\action,\state';\textcolor{blue}{\bar\params}, \paramsQ, \textcolor{red}{\bar{\paramsQ}}) = r + \gamma{\widehat Q(\state',\pi(\state';\textcolor{blue}{\bar\params}); \textcolor{red}{\bar{\paramsQ}})} - \widehat Q(\state,\action;\paramsQ), \label{eq:td_err_target}
\end{align}
to improve stability. The policy is updated softly at each iteration, i.e., $\bar{\params}_{t+1} = \tau_\paramsQ\params_t + (1-\tau_\params)\bar{\params}_t$, with $\tau_\params \in (0,1]$
The second algorithm (\textbf{DPG TD-REG}) applies the penalty function $G(\params)$ presented in this paper and does not use the target policy, i.e., 
\begin{align}
	\delta_Q(\state,\action,\state';\textcolor{blue}{\params}, \paramsQ, \textcolor{red}{\bar{\paramsQ}}) = r + \gamma {\widehat Q(\state',\pi(\state';\textcolor{blue}{\params}); \textcolor{red}{\bar{\paramsQ}})} - \widehat Q(\state,\action;\paramsQ), \label{eq:td_err_notarget}
\end{align}
in order to compute the full derivative with respect to $\params$ for the penalty function (Eq.~\eqref{eq:dpg_tdreg}).
%
The third algorithm (\textbf{DPG NO-TAR}) is like DPG, but also does not use the target policy. The purpose of this version is to check that the benefits of our approach do not come from the lack of the target actor, but rather from the TD-regularization.

The last two versions are twin delayed DPG (\textbf{TD3})~\citep{fujimoto2018addressing}, which achieved state-of-the-art results, and its TD-regularized counterpart (\textbf{TD3 TD-REG}). TD3 proposes three modifications to DPG. First, in order to reduce overestimation bias, there are two critics. Only the first critic is used to update the policy, but the TD target used to update both critics is given by the minimum of their TD target. Second, the policy is not updated at each step, but the update is delayed in order to reduce per-update error. Third, since deterministic policies can overfit to narrow peaks in the value estimate (a learning target using a deterministic policy is highly
susceptible to inaccuracies induced by function approximation
error) noise is added to the target policy.
The resulting TD error is 
\begin{align}
	\delta_{Q_i}(\state,\action,\state';\textcolor{blue}{\bar\params}, \paramsQ, \textcolor{red}{\bar{\paramsQ}_1, \bar{\paramsQ}_2}) = r + \gamma \min_{j=1,2} {\widehat Q(\state',\pi(\state';\textcolor{blue}{\bar\params}) + \xi; \textcolor{red}{\bar{\paramsQ}_j})} - \widehat Q(\state,\action;\paramsQ_i), \label{eq:td3}
\end{align}
where the noise $\xi = \texttt{clip}(\gaussian(0,\tilde \sigma), -c, c)$ is clipped to keep the target close to the original action. Similarly to DPG TD-REG, \textbf{TD3 TD-REG} removes the target policy (but keeps the noise $\xi$) and adds the TD-regularization to the policy update. Since TD3 updates the policy according to the first critic only, the TD-regularization considers the TD error in Eq. \eqref{eq:td3} with $i=1$.

\paragraph*{Hyperparameters}
\begin{itemize}
	\setlength{\itemsep}{0em}
	\item Maximum number of steps per trajectory: 150.
	\item Exploration: Gaussian noise (diagonal covariance matrix) added to the action. The standard deviation $\sigma$ starts at 5 and decays at each step according to $\sigma_{t+1} = 0.95\sigma_t$.
	\item Discount factor: $\gamma = 0.99$.
	\item Steps collected before learning (to initialize the experience replay memory): $100$.
	\item Policy and TD errors evaluated every 100 steps.
	\item At each step, all data collected (state, action, next state, reward) is stored in the experience replay memory, and a mini-batch of 32 random samples is used for computing the gradients.
	\item DPG target policy update coefficient: $\tau_\params = 0.01$ (DPG NO-TAR is like DPG with $\tau_\params = 1$). With $\tau_\params = 0.1$ results were worse. With $\tau_\params = 0.001$ results were almost the same. 
	\item ADAM hyperparameters for the gradient of $\paramsQ$: $\alpha = 0.01$, $\beta_1 = 0.9$, $\beta_2 = 0.999$, $\epsilon = 10^{-8}$. With higher $\alpha$ all algorithms were unstable, because the critic was changing too quickly.
	\item ADAM hyperparameters for the gradient of $\params$: $\alpha = 0.0005$, $\beta_1 = 0.9$, $\beta_2 = 0.999$, $\epsilon = 10^{-8}$. Higher $\alpha$ led all algorithms to divergence, because the condition for stability (magnitude of the eigenvalues of $(A+BK)$ smaller than one) was being violated.
	\item Regularization coefficient: $\tdcoeff_0 = 0.1$ and then it decays according to $\tdcoeff_{t+1} = 0.999\tdcoeff_t$.
	\item In TD3 original paper, $\xi \sim \gaussian(0,0.2)$ and is clipped in [-0.5,0.5]. However, the algorithm was tested on tasks with action bounded in [-1,1]. In the LQR, instead, the action is unbounded, therefore we decided to use $\xi \sim \gaussian(0,2)$ and to clip it in $[-\sigma_t/2, \sigma_t/2]$, where $\sigma_t$ is the current exploration noise. We also tried different strategies, but we noticed no remarkable differences. The noise is used only for the policy and critics updates, and it is removed for the evaluation of the TD error for the plots.
	\item In TD3 and TD3 TD-REG, the second critic parameters are initialized uniformly in $[-1,1]$.
	\item In TD3 and TD3 TD-REG, the policy is updated every two steps, as in the original paper.
	\item The learning of all algorithms ends after $12,000$ steps. 
\end{itemize}

\begin{figure}[h]
	\begin{center}
		\bfseries DPG on LQR - Quadratic Features \vspace*{-0.5em}
	\end{center}
	\begin{minipage}[t]{\textwidth}
		\begin{subfigure}[t]{.325\linewidth}
			\centering
			\includegraphics[width=\textwidth]{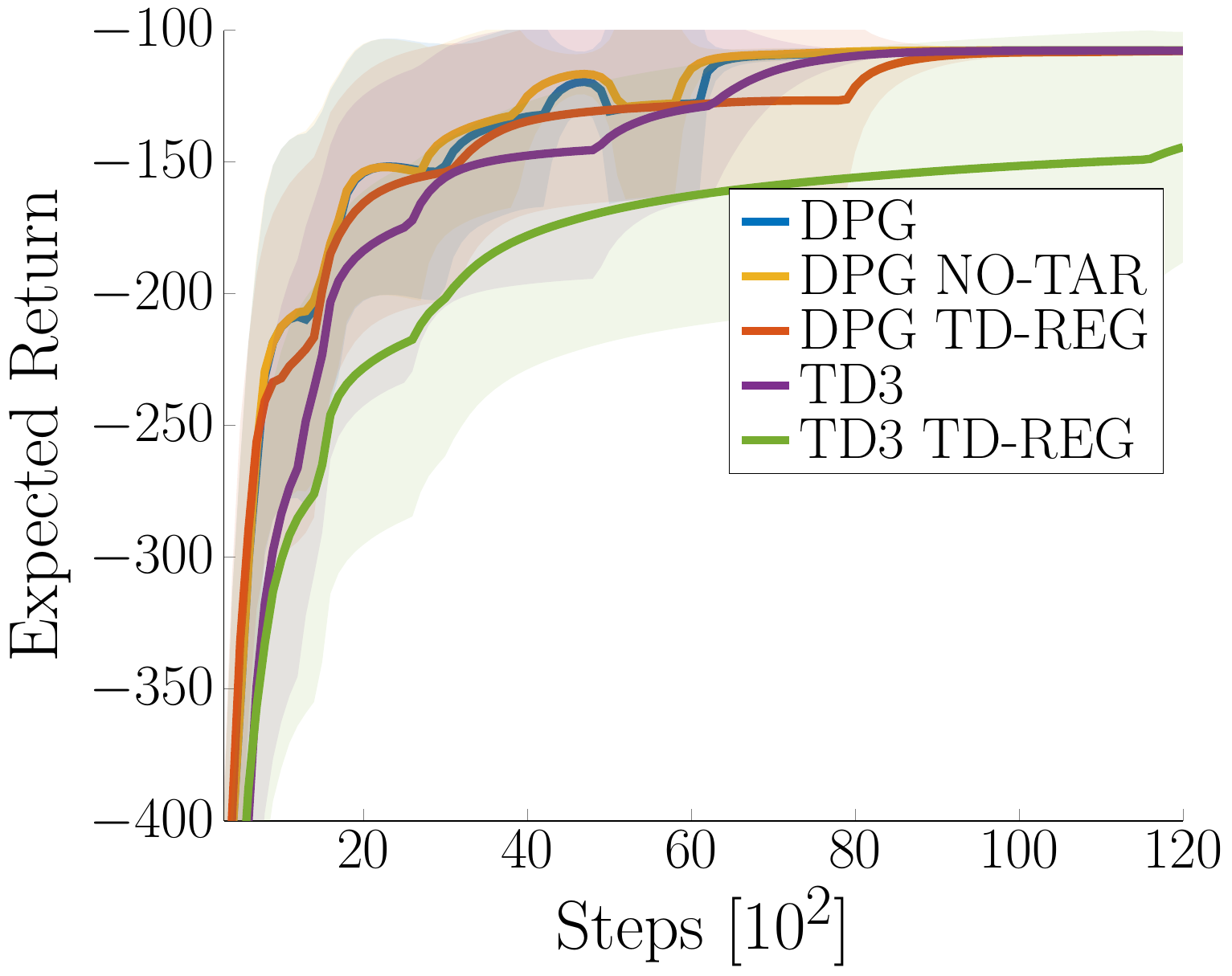}
		\end{subfigure}
		\hfill
		\begin{subfigure}[t]{.325\linewidth}
			\centering
			\includegraphics[width=\textwidth]{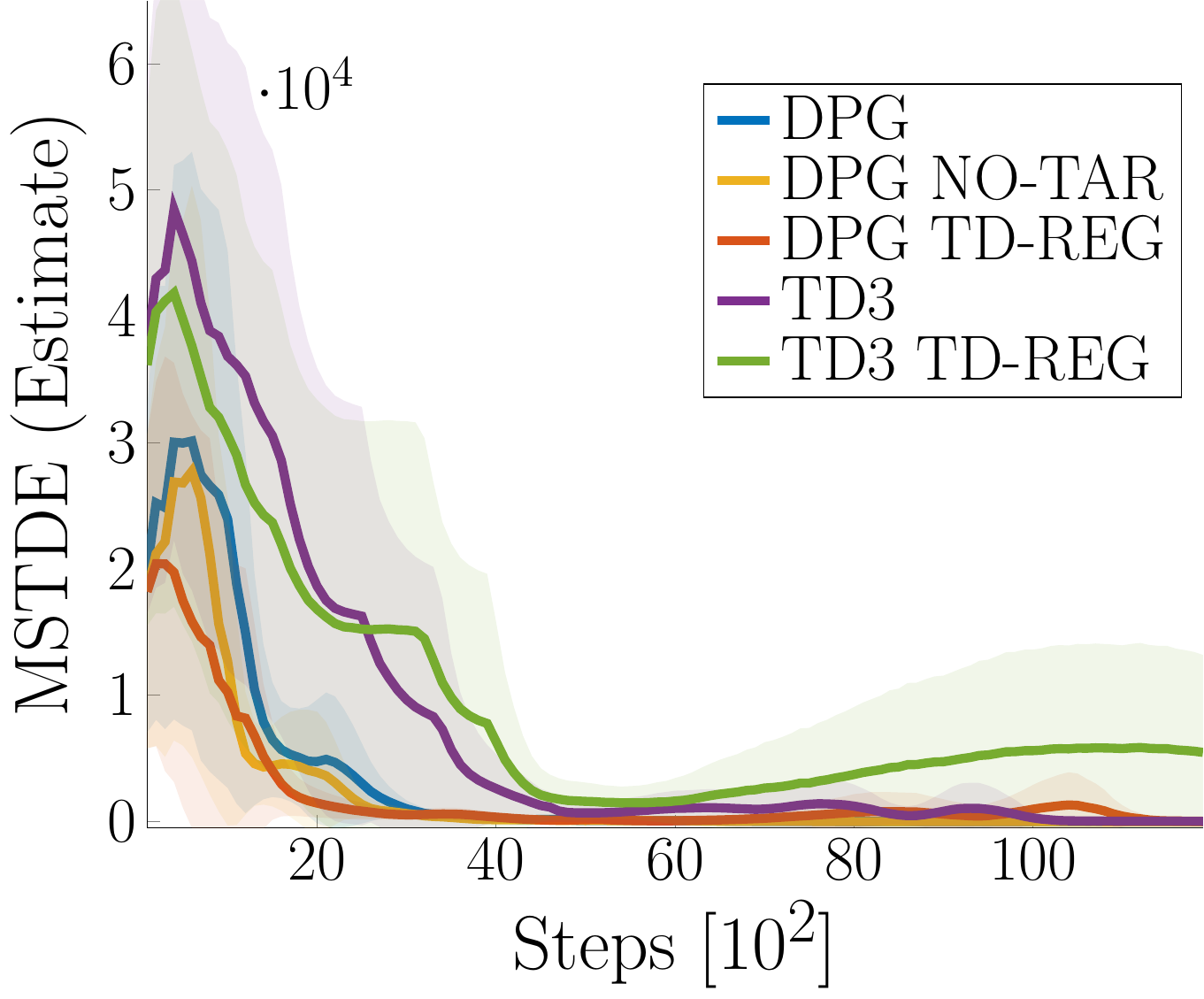}
		\end{subfigure}
		\hfill
		\begin{subfigure}[t]{.325\linewidth}
			\centering
			\includegraphics[width=\textwidth]{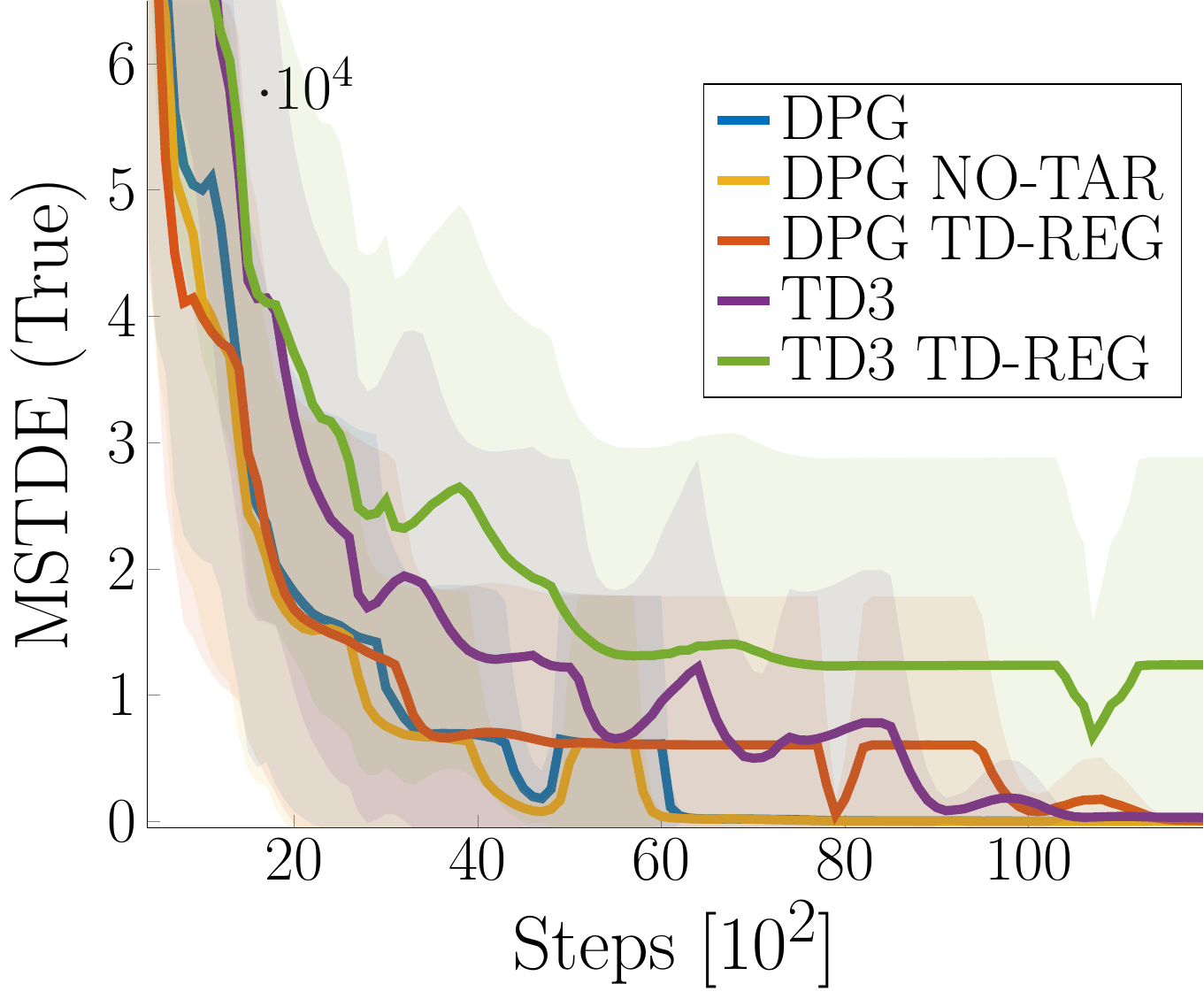}
		\end{subfigure}
		\caption{\label{fig:lqr_squared_plots} All algorithms perform similarly (DPG and DPG NO-REG almost overlap), because quadratic features are sufficient to approximate the true Q-function. Only TD3 TD-REG did not converge within the time limit in two runs, but it did not diverge either.}
	\end{minipage}
\end{figure}

\begin{figure}[h]
	\begin{center}
		\bfseries DPG on LQR - Cubic Features \vspace*{-0.5em}
	\end{center}
	\begin{minipage}{\textwidth}
		\begin{subfigure}[t]{.325\linewidth}
			\centering
			\includegraphics[width=\textwidth]{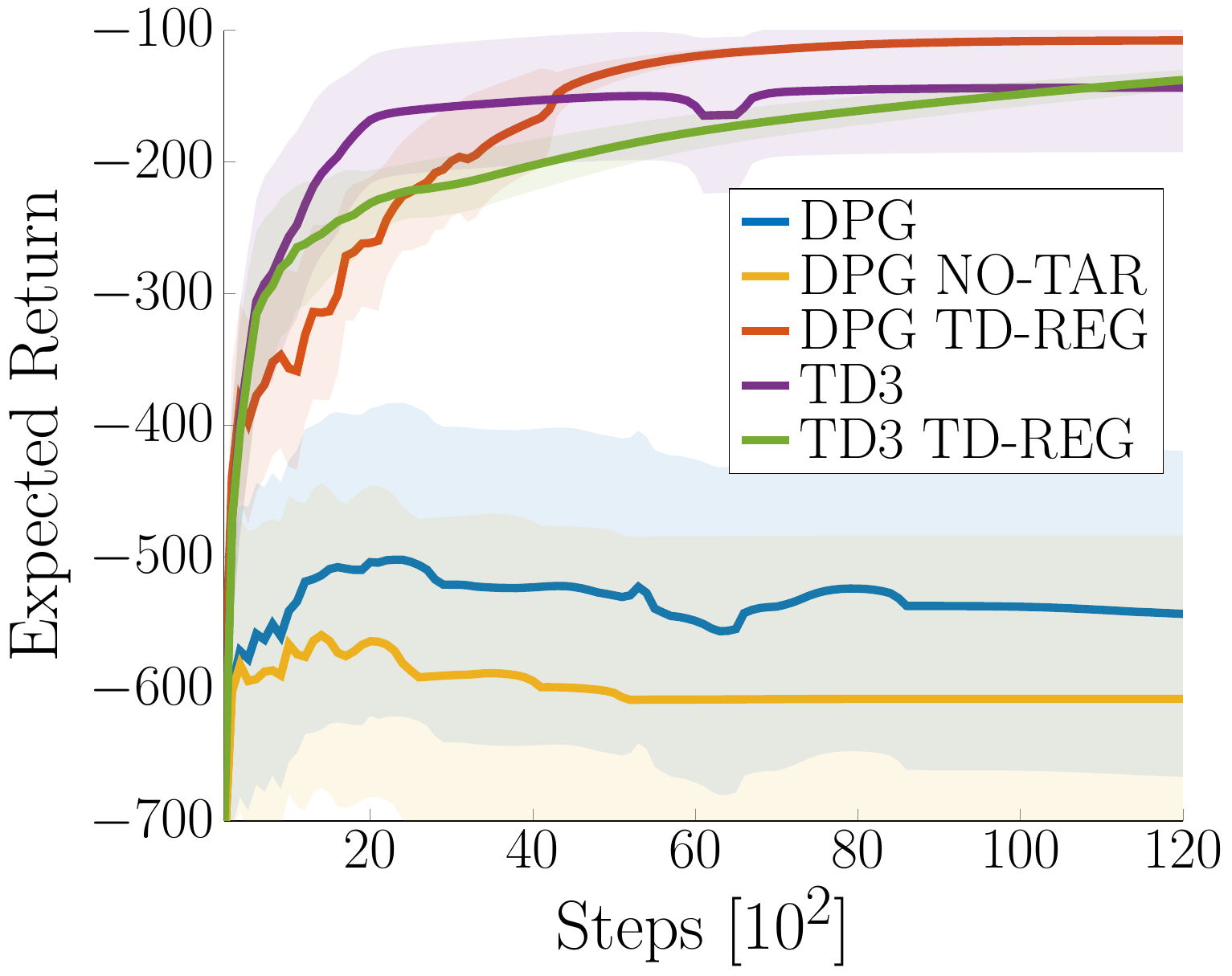}
		\end{subfigure}
		\hfill
		\begin{subfigure}[t]{.325\linewidth}
			\centering
			\includegraphics[width=\textwidth]{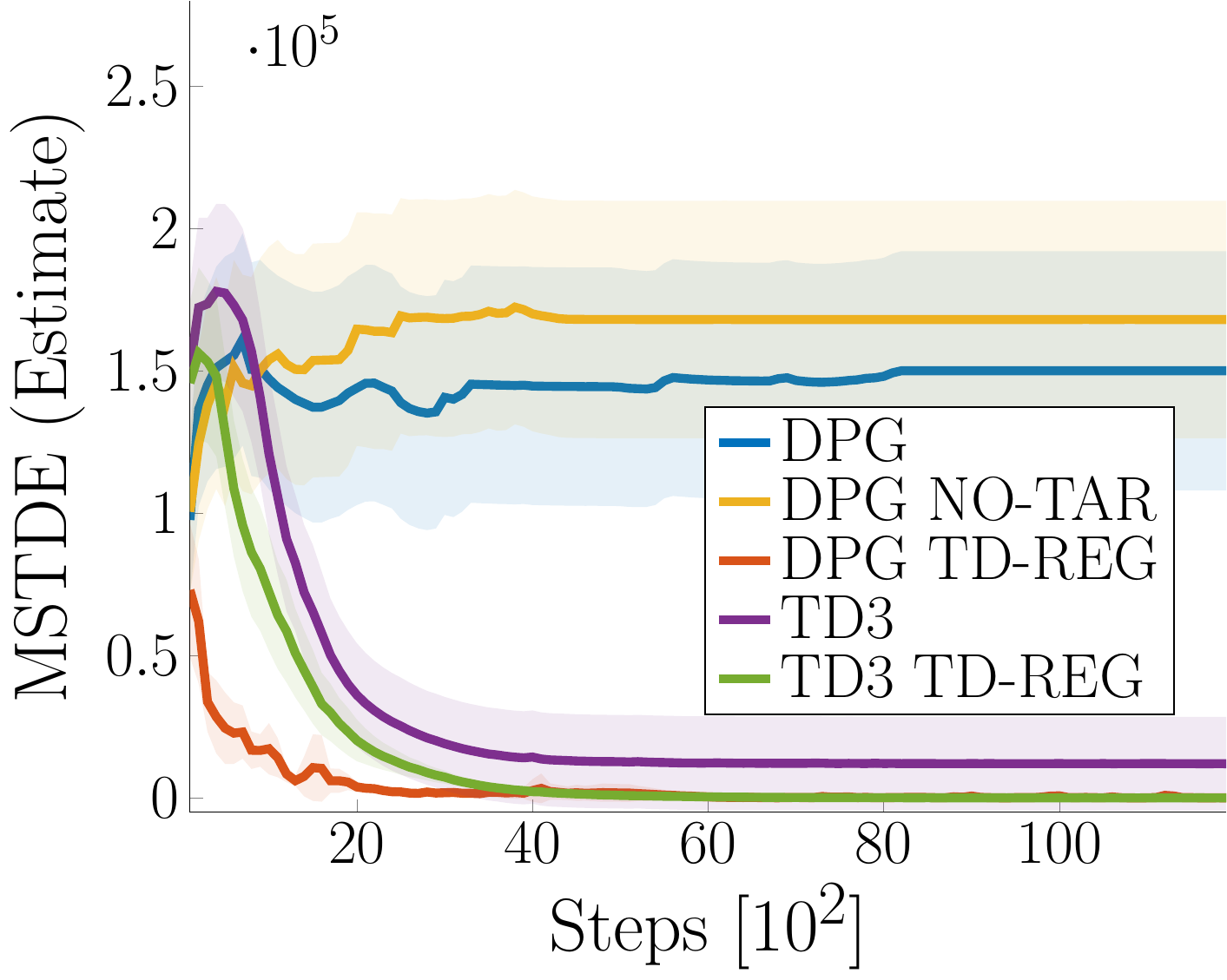}
		\end{subfigure}
		\hfill
		\begin{subfigure}[t]{.325\linewidth}
			\centering
			\includegraphics[width=\textwidth]{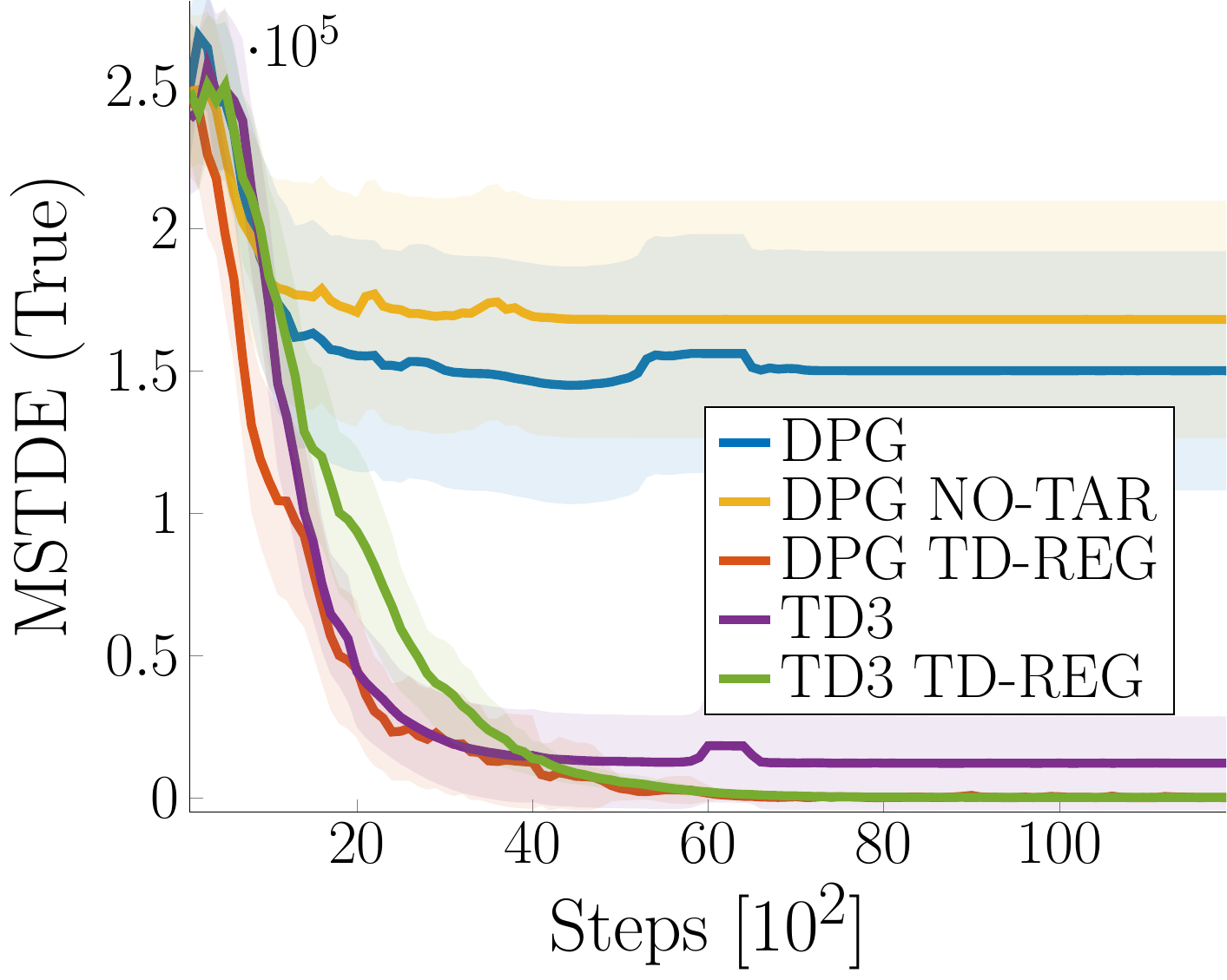}
		\end{subfigure}
		\caption{\label{fig:lqr_cubic_plots} By contrast, with cubic features the critic is prone to overfit, and DPG and DPG NO-TAR diverged 24 and 28 times, respectively. TD3 performed better, but still diverged two times. TD-regularized algorithms, instead, never diverged, and DPG TD-REG always learned the true Q-function and the optimal policy within the time limit. Similarly to Figure~\ref{fig:lqr_squared_plots}, TD3 TD-REG is the slowest, and in this case it never converged within the time limit, but did not diverge either (see also Figure~\ref{fig:lqr_3_1_path}).}
	\end{minipage}
	\begin{center}
		\bfseries Path of the Learned Policy Parameters $\params$
	\end{center}
	\begin{minipage}{\textwidth}
		\begin{subfigure}[t]{.49\linewidth}
			\centering
			\includegraphics[width=\textwidth]{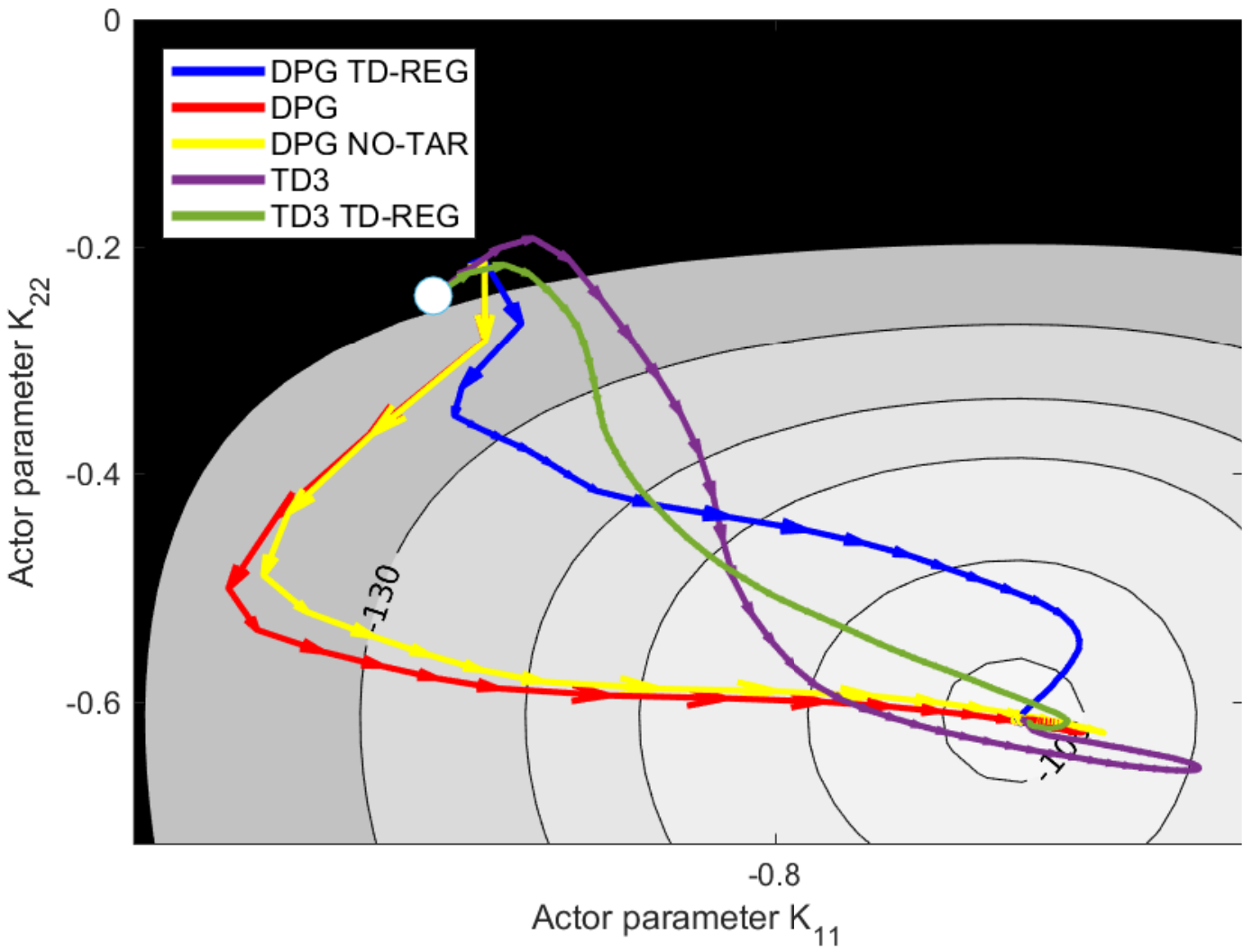}
			\caption{\label{fig:lqr_2_1_path} Quadratic features.}
		\end{subfigure}
		\hfill
		\begin{subfigure}[t]{.49\linewidth}
			\centering
			\includegraphics[width=\textwidth]{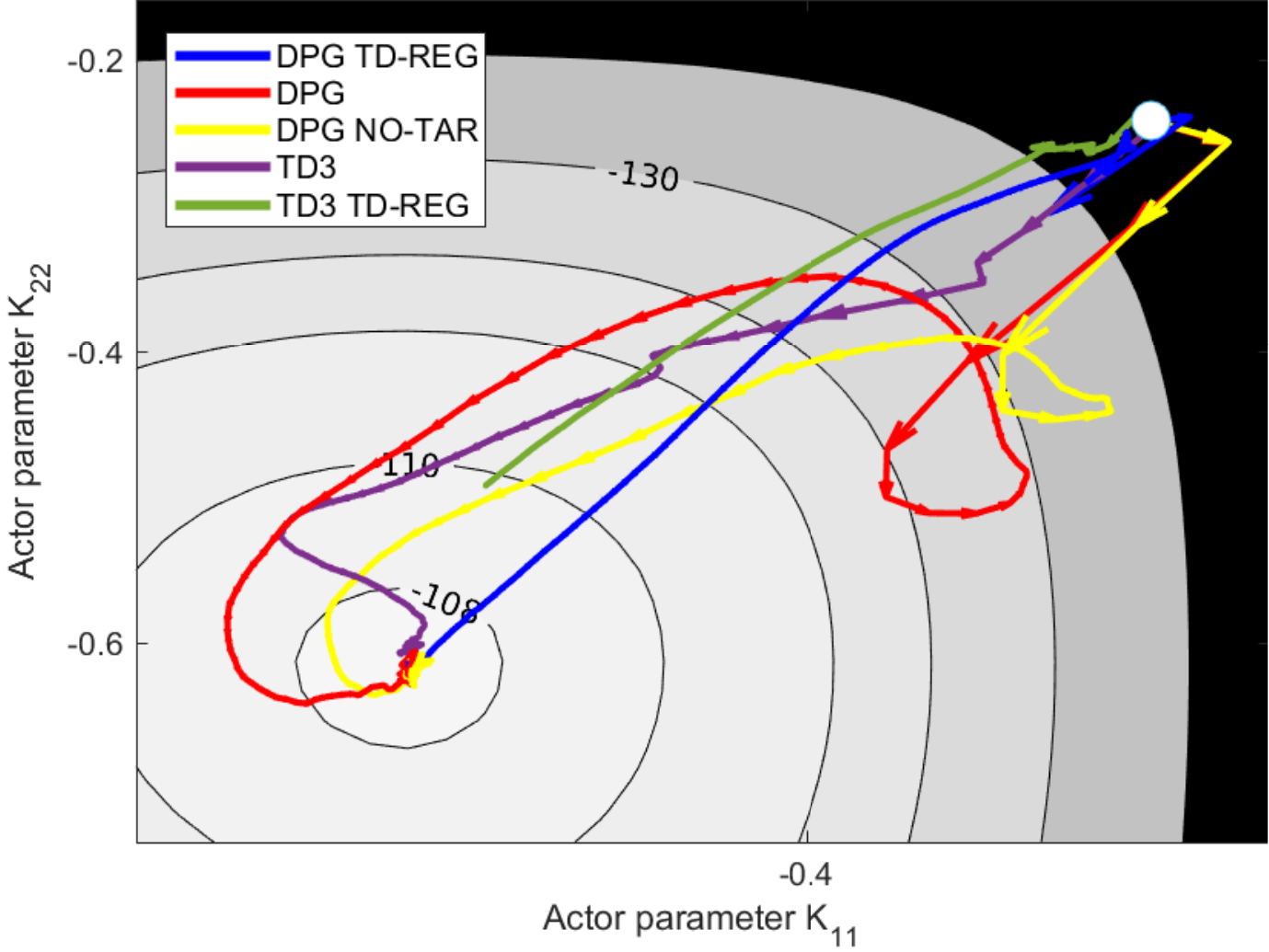}
			\caption{\label{fig:lqr_3_1_path} Cubic features.}
		\end{subfigure}
		\caption{\label{fig:lqr_paths} Paths followed by the policy parameters during runs in which no algorithm diverged. Each arrow represents 100 update steps. Contour denotes the expected return magnitude. The initial policy parameter vector $\params$ is denoted by the white circle. The TD-regularization enables safer and more stable trajectories, leading the policy parameters more straightforwardly to the optimum. We can also see that {blue} and {green} arrows are initially shorter, denoting smaller update steps. In fact, due to the initial inaccuracy of the critic, the TD-regularization gradient ``conflicts'' with vanilla gradient at early stages of the learning and avoids divergence. However, in the case of TD3 TD-REG (green arrows) the use of both two critics, delayed policy updates and TD-regularization slows down the learning up to the point that the algorithm never converged within the time limit, as in Figure~\ref{fig:lqr_3_1_path}.}
	\end{minipage}
\end{figure}

\paragraph{Results\\}
Figures~\ref{fig:lqr_squared_plots} and~\ref{fig:lqr_2_1_path} show the results using quadratic features. 
Since these features are very unlikely to overfit, all algorithms perform almost the same. However, from Figure~\ref{fig:lqr_2_1_path} we can clearly see that the TD-regularization keeps the policy parameters on a more straight path towards the optimum, avoiding detours. Both the TD-regularization and TD3 also result in smaller, but safer, update steps. This behavior is reflected by the learning curves in Figure~\ref{fig:lqr_squared_plots}, as DPG and DPG NO-TAR converge slightly faster. However, using both TD3 and TD-REG at the same time can result in excessively slow learning. The green arrows in Figure~\ref{fig:lqr_2_1_path} are, in fact, the smallest, and in Figure~\ref{fig:lqr_squared_plots} TD3 TD-REG did not converge within the time limits in two runs (but it did not diverge either). 
\\
Looking at the TD error plots, it is interesting to notice that the estimated TD error is always smaller than the true TD error, meaning that the critic underestimates the TD error.  This is a normal behavior, considering the stochasticity of the behavior policy and of the environment. 
It is also normal that this overestimation bias is less prominent in TD3, thanks to the use of two critics and to delayed policy updates.
However, TD3 TD-REG is surprisingly the only algorithm increasing the estimated TD error around mid-learning. We will further investigate this behavior at the end of this section.

Results are substantially different when cubic features are used (Figure~\ref{fig:lqr_cubic_plots}). In this case, many features are irrelevant and the model is prone to overfit. As a consequence, the TD error shown in Figure~\ref{fig:lqr_cubic_plots} is much larger than the one shown in Figure~\ref{fig:lqr_squared_plots}, meaning that it is harder for the critic to correctly approximate the true Q-function. The problem is prominent for DPG and DPG NO-TAR, which cannot learn 24 and 28 times, respectively, out of 50 (thus, the very large confidence interval). Similarly to the previous case, the true TD error is underestimated. Initially their critics estimate a TD error of $10^5$ but the true one is $2.5 \mathord{\cdot} 10^5$. This large error guides the actors incorrectly, inevitably leading to divergence. Then, after approximately 3,000 steps, the two TD errors match at $1.5 \mathord{\cdot} 10^5$.
Among the two algorithms, DPG NO-TAR performs worse, due to the lack of the target policy. 
\\
TD3 performs substantially better, but still diverges two times. 
By contrast, TD-REG algorithms never diverges. Figure~\ref{fig:lqr_3_1_path} shows the benefits of the TD-regularization in the policy space. Initially, when the TD error is large, the policy ``moves around'' the starting point. Then, as the critic accuracy increases, the policy goes almost straightforwardly to the goal. This ``slow start'' behavior is also depicted in Figure~\ref{fig:lqr_cubic_plots}, where DPG TD-REG expected return initially improves more slowly compared to TD3 and TD3 TD-REG. Finally, we notice once again that the combination of TD3 and TD-REG results in the slowest learning: unlike TD3, TD3 TD-REG never diverged, but it also never converged within the time limit. This behavior is also depicted in Figure~\ref{fig:lqr_3_1_path}, where green arrows (TD3 TD-REG policy updates) are so small that the algorithm cannot reach the goal in time.

\clearpage

\begin{figure}[h]
	\begin{center}
		\bfseries TD3 on LQR - Quadratic Features, No Policy Update Delay \vspace*{-0.5em}
	\end{center}
	\begin{minipage}[t]{\textwidth}
		\begin{subfigure}[t]{.325\linewidth}
			\centering
			\includegraphics[width=\textwidth]{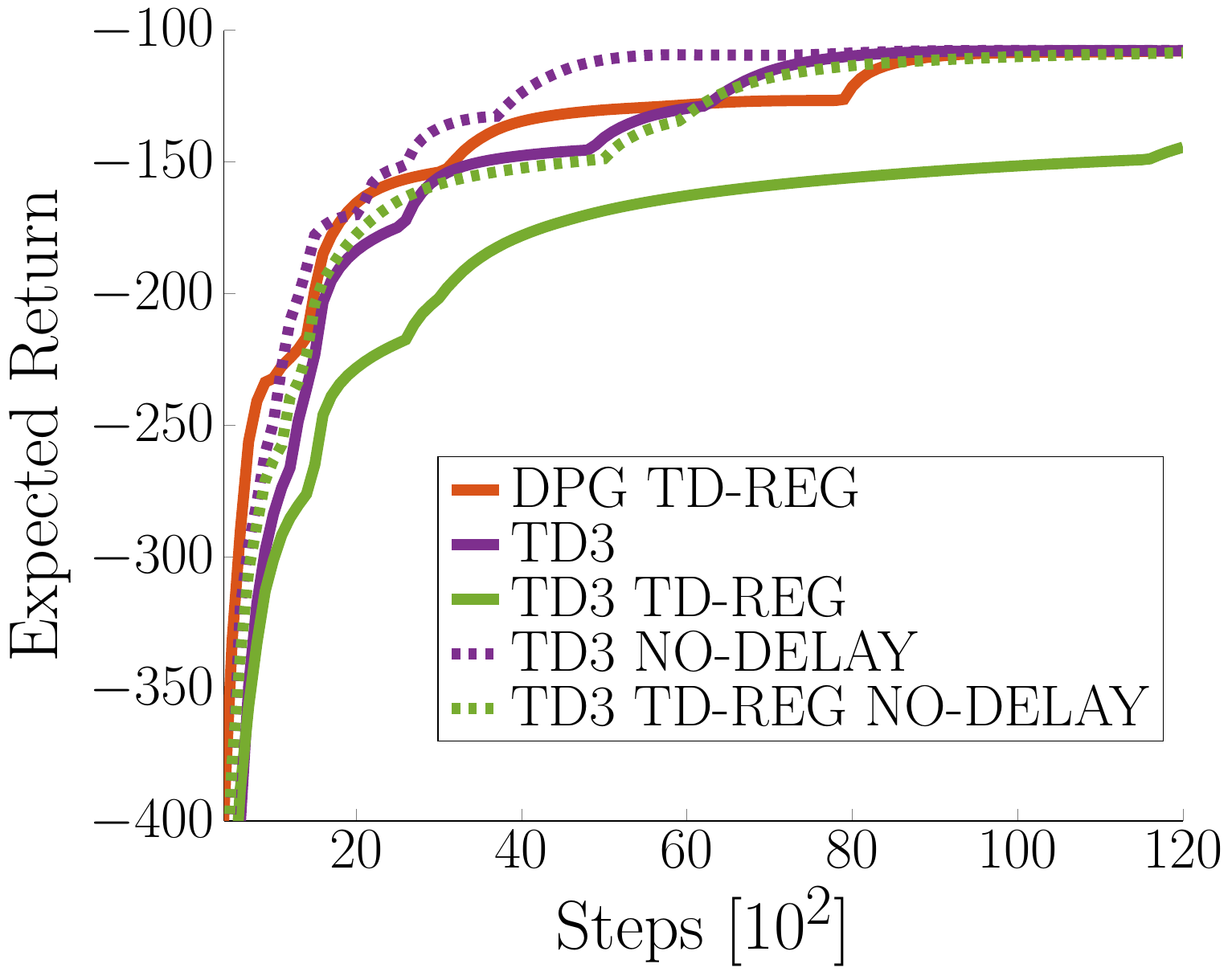}
		\end{subfigure}
		\hfill
		\begin{subfigure}[t]{.325\linewidth}
			\centering
			\includegraphics[width=\textwidth]{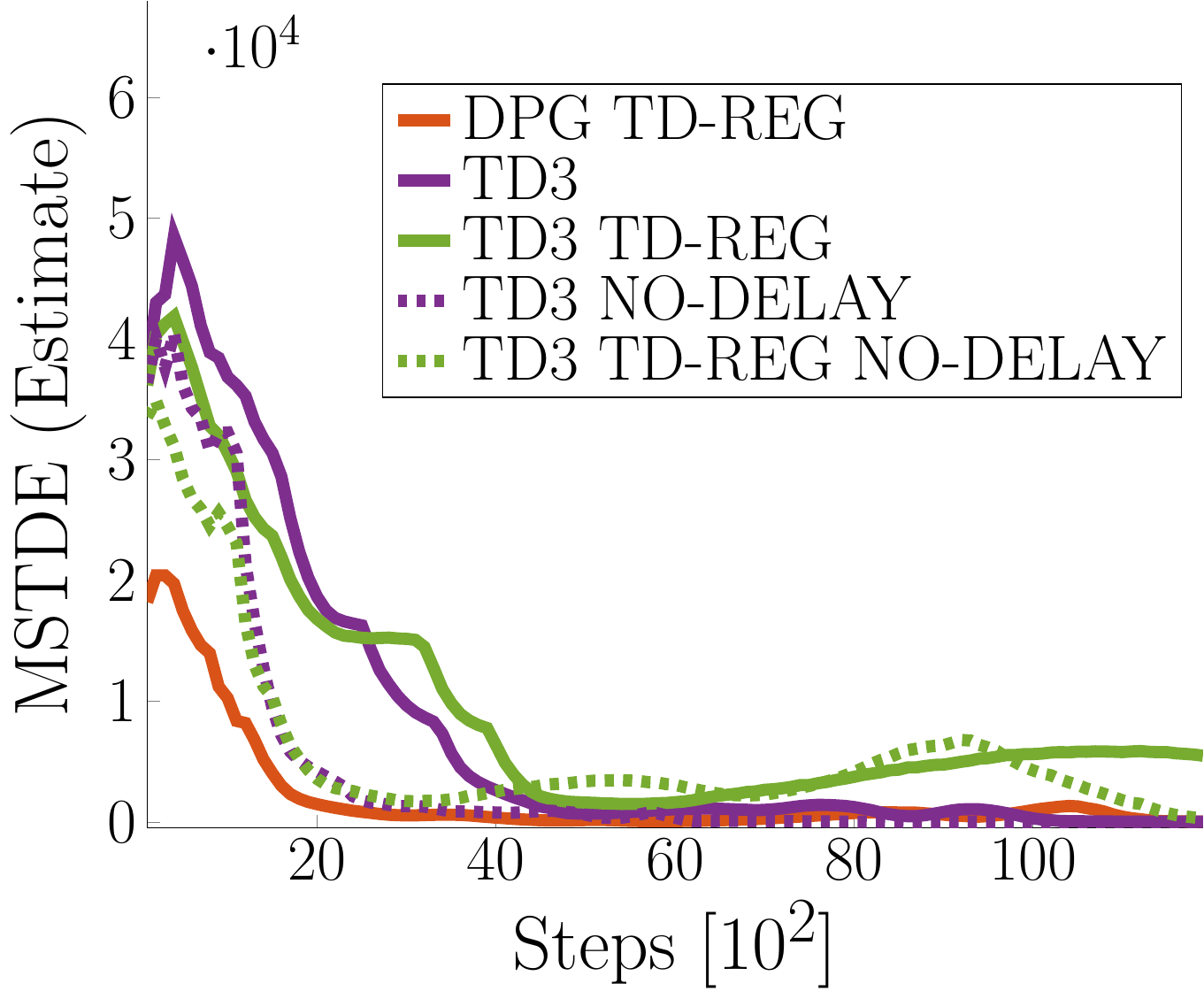}
		\end{subfigure}
		\hfill
		\begin{subfigure}[t]{.325\linewidth}
			\centering
			\includegraphics[width=\textwidth]{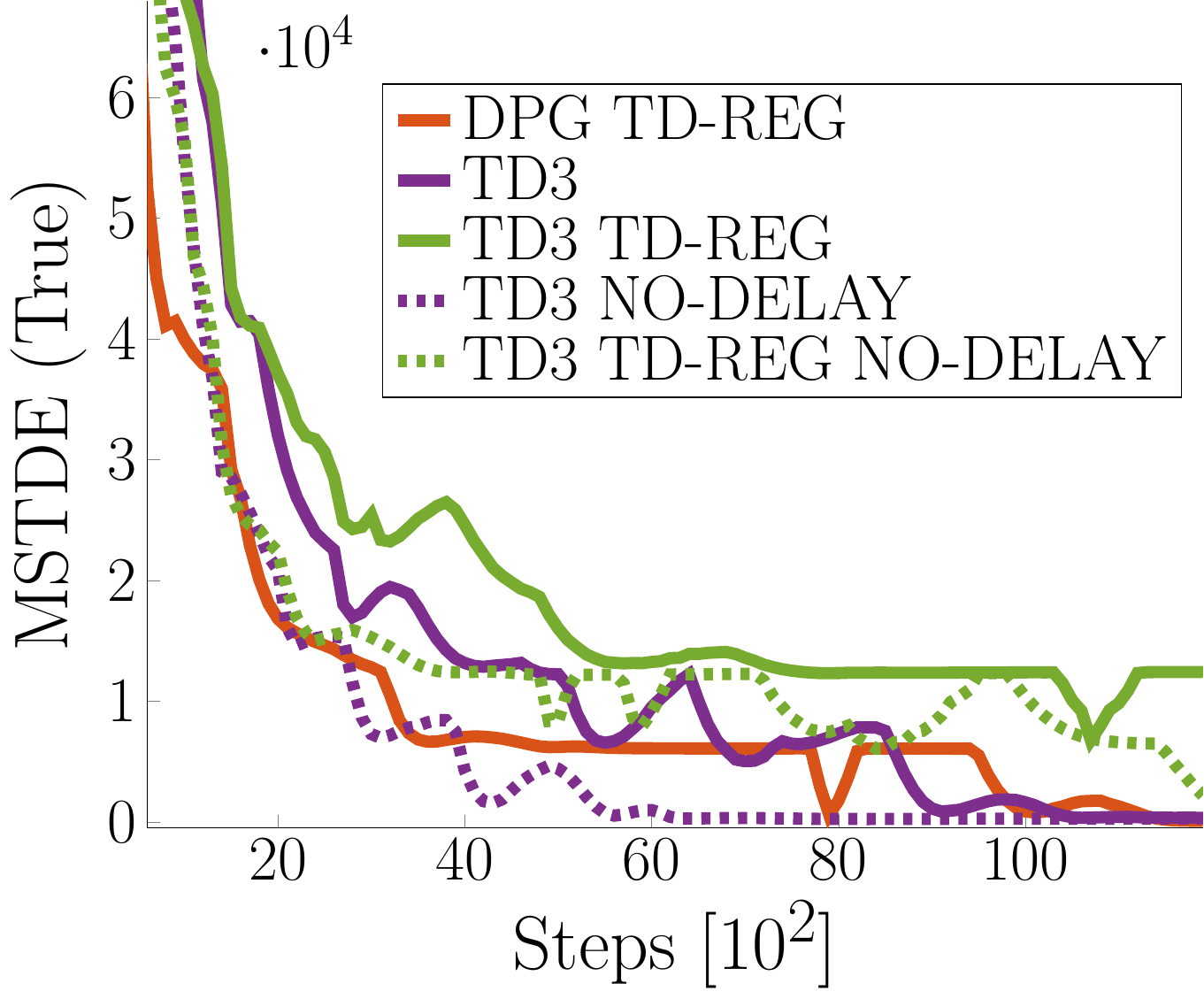}
		\end{subfigure}
		\caption{\label{fig:lqr_squared_plots_nodelay} Without policy update delays, both TD3 and TD3 TD-REG converge faster.}
	\end{minipage}
	\begin{center}
		\bfseries TD3 on LQR - Cubic Features, No Policy Update Delay \vspace*{-0.5em}
	\end{center}
	\begin{minipage}{\textwidth}
		\begin{subfigure}[t]{.325\linewidth}
			\centering
			\includegraphics[width=\textwidth]{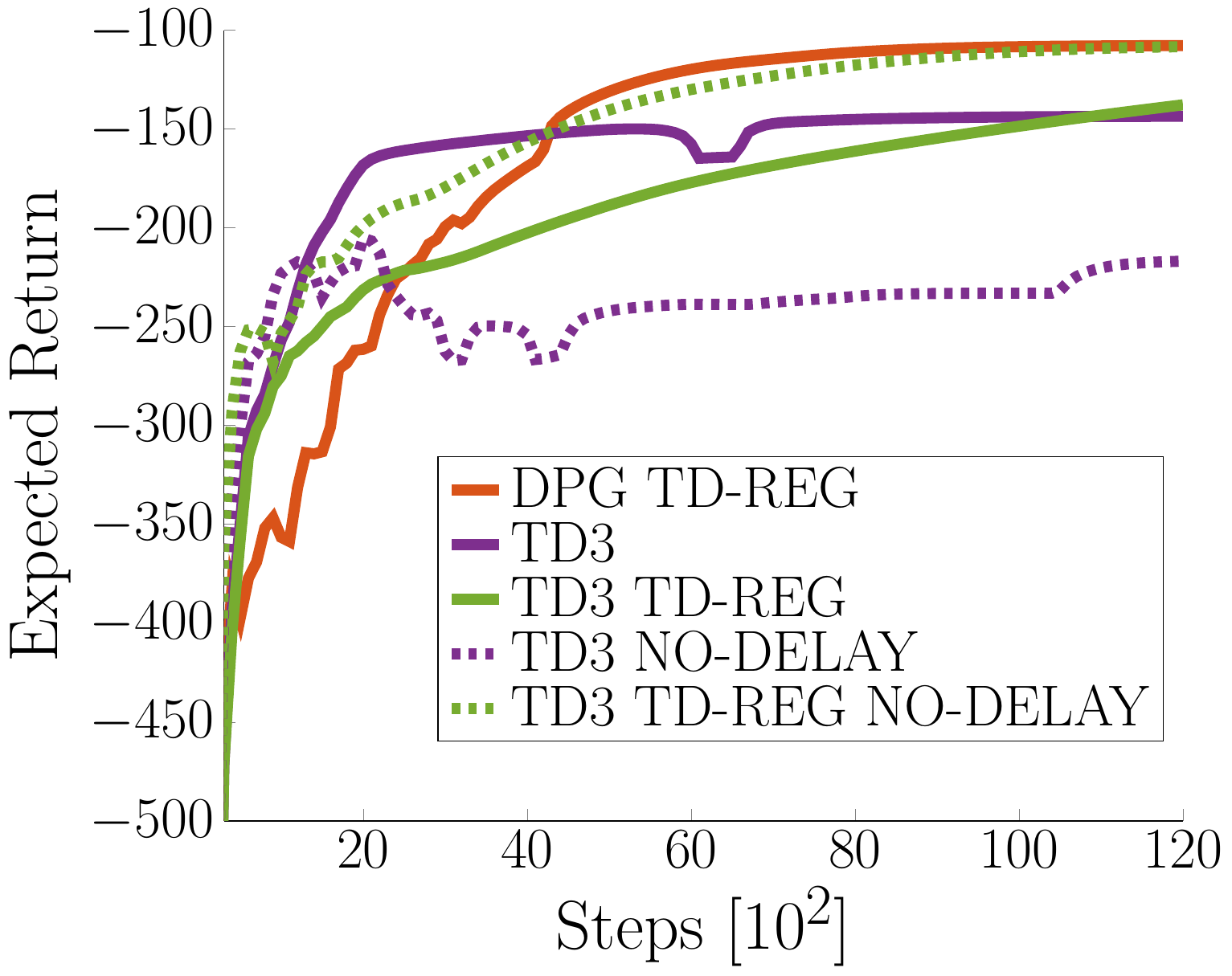}
		\end{subfigure}
		\hfill
		\begin{subfigure}[t]{.325\linewidth}
			\centering
			\includegraphics[width=\textwidth]{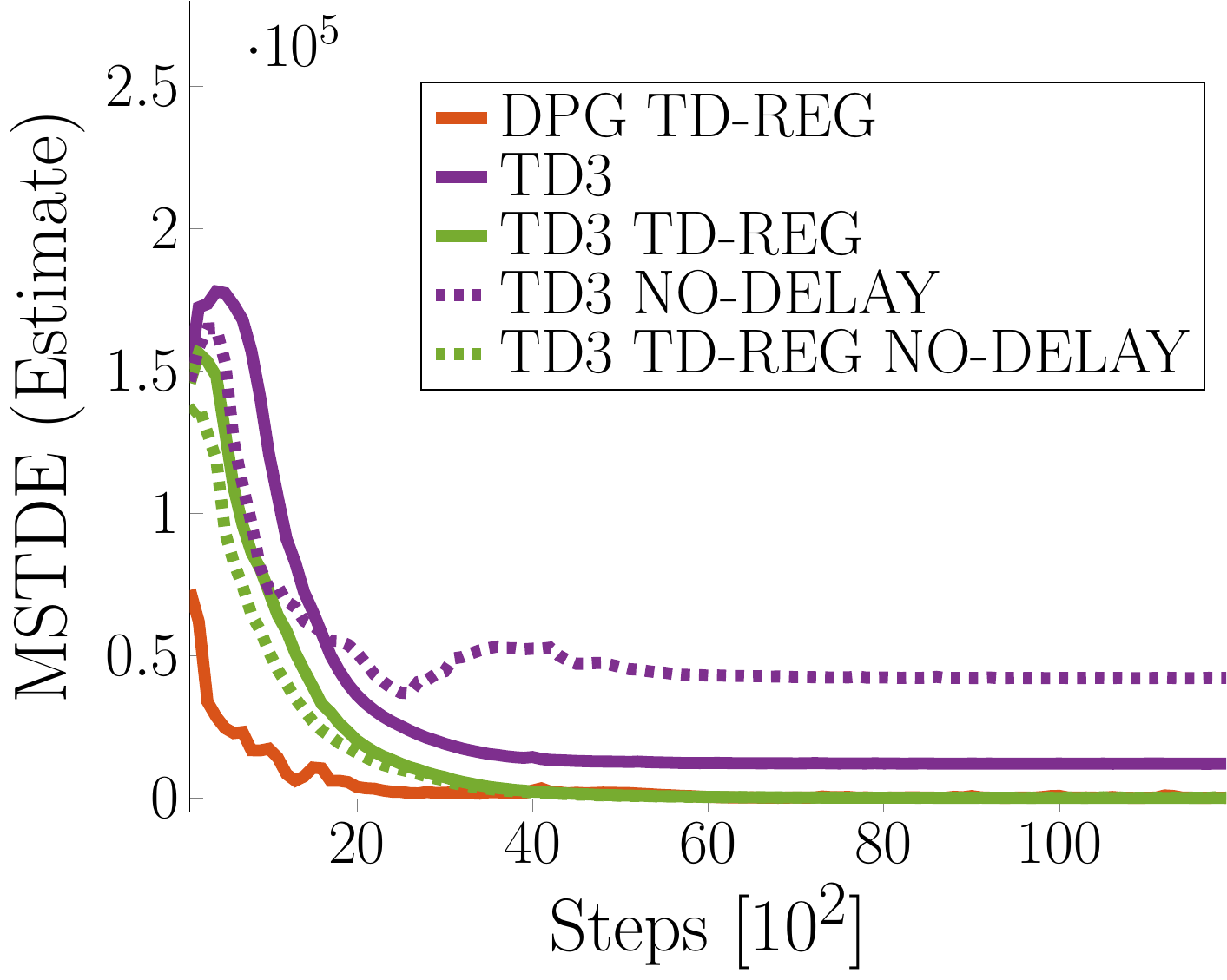}
		\end{subfigure}
		\hfill
		\begin{subfigure}[t]{.325\linewidth}
			\centering
			\includegraphics[width=\textwidth]{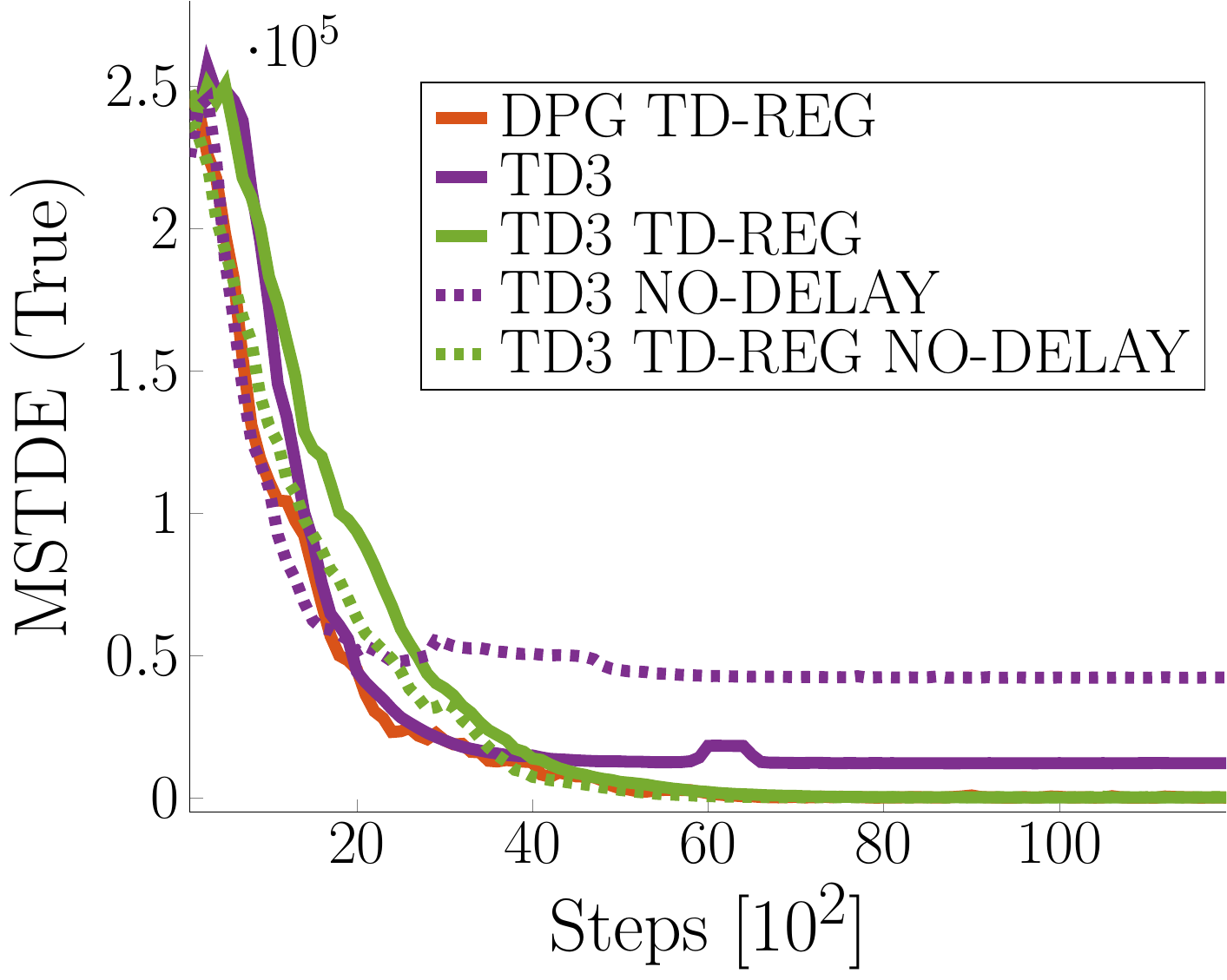}
		\end{subfigure}
		\caption{\label{fig:lqr_cubic_plots_nodelay} With cubic features, TD3 NO-DELAY performs worse than TD3, diverging six times instead of two. By contrast, TD3 TD-REG NO-DELAY performs better than TD3 TD-REG and it always converges within the time limit.}
	\end{minipage}
\end{figure}

Figure~\ref{fig:lqr_cubic_plots} hints that the ``slow learning'' behavior of TD3 TD-REG may be due to the delayed policy update, as both the estimated and the true TD error are close to zero by mid-learning. To further investigate this behavior, we performed the same experiments without delayed policy updates for both TD3 and TD3-REG. For the sake of clarity, we report the results on separates plots without DPG and DPG NO-TAR and without shaded areas for confidence interval. In the case of quadratic features (Figure~\ref{fig:lqr_squared_plots_nodelay}) both TD3 and TD3 TD-REG gain from the removal of the policy update delay. However, in the case of cubic features (Figure~\ref{fig:lqr_cubic_plots_nodelay}), TD3 NO-DELAY performs worse than TD3, as it diverges six times instead of two. By contrast, TD3 TD-REG NO-DELAY performs better than TD3 TD-REG, and the expected return curve shows traits of both TD3 and TD-REG: initially it improves faster, like TD3, and then it always converges to optimality, like DPG TD-REG.
We can conclude that delaying policy updates is not necessary when appropriate features are used and overfitting cannot happen. Otherwise, the combination of two critics and TD-regularization yields the best results, with the TD-regularization providing the most benefits.

\subsection{SPG Evaluation on the LQR}
\label{app:lqr_spg}
DPG is an on-line algorithm, i.e., it performs a critic/actor update at each time step, using mini-batches of previously collected samples. Instead, stochastic policy gradient (SPG) collects complete trajectories with a stochastic policy before updating the critic/actor.
In this section, we evaluate SPG, SPG TD-REG, and REINFORCE~\citep{williams1992simple}. The first two maximize Q-function estimates given by a learned critic. SPG TD-REG additionally applies the TD-regularization presented in the paper. REINFORCE, instead, maximizes Monte Carlo estimates of the Q-function, i.e., 
\begin{equation}
\widehat Q^\pi(\state_t,\action_t) = {\sum_{i=t}^T \gamma^{i-t} r_{i}}.
\end{equation}
The policy is Gaussian, i.e., $\pi(\action|\state;\params) = \gaussian(K\state, \Sigma)$, where $\Sigma$ is a diagonal covariance matrix. $K$ is initialized as for DPG. The diagonal entries of $\Sigma$ are initialized to five. Six policy parameters are learned, four for $K$ and two for $\Sigma$. 
For SPG TD-REG, the expectation  $\EVV{\pi(\action'|\state';\params)}{\widehat Q(\state',\action';\paramsQ)}$ is approximated with the policy mean Q-value, i.e., $\smash{\widehat Q(\state', K\state';\paramsQ)}$.
For SPG and SPG TD-REG, $\widehat Q(\state,\action;\paramsQ)$ is learned by Matlab \texttt{fminunc} optimizer using the samples collected during the current iteration.

\paragraph*{Hyperparameters}
\begin{itemize}
\setlength{\itemsep}{0em}
	\item Trajectories collected per iteration: 1 or 5.
	\item Steps per trajectory: 150.
	\item Discount factor: $\gamma = 0.99$.
	\item Policy and TD error evaluated at every iteration.
	\item No separate target policy or target Q-function are used for learning the critic, but we still consider $\nabla_\paramsQ \widehat Q(\state',\action';\paramsQ) = 0$.
	\item Policy update learning rate: 0.01. The gradient is normalized if its norm is larger than one.
	\item The policy update performs only one step of gradient descent on the whole dataset.
	\item Regularization coefficient: $\tdcoeff_0 = 0.1$ and then it decays according to $\tdcoeff_{t+1} = 0.999\tdcoeff_t$.
\end{itemize}

\begin{figure}[h]
	\begin{center}
		\bfseries SPG on LQR - One Episode per Iteration
	\end{center}
	\begin{minipage}[t]{\textwidth}
		\begin{subfigure}[t]{.325\linewidth}
			\centering
			\includegraphics[width=\textwidth]{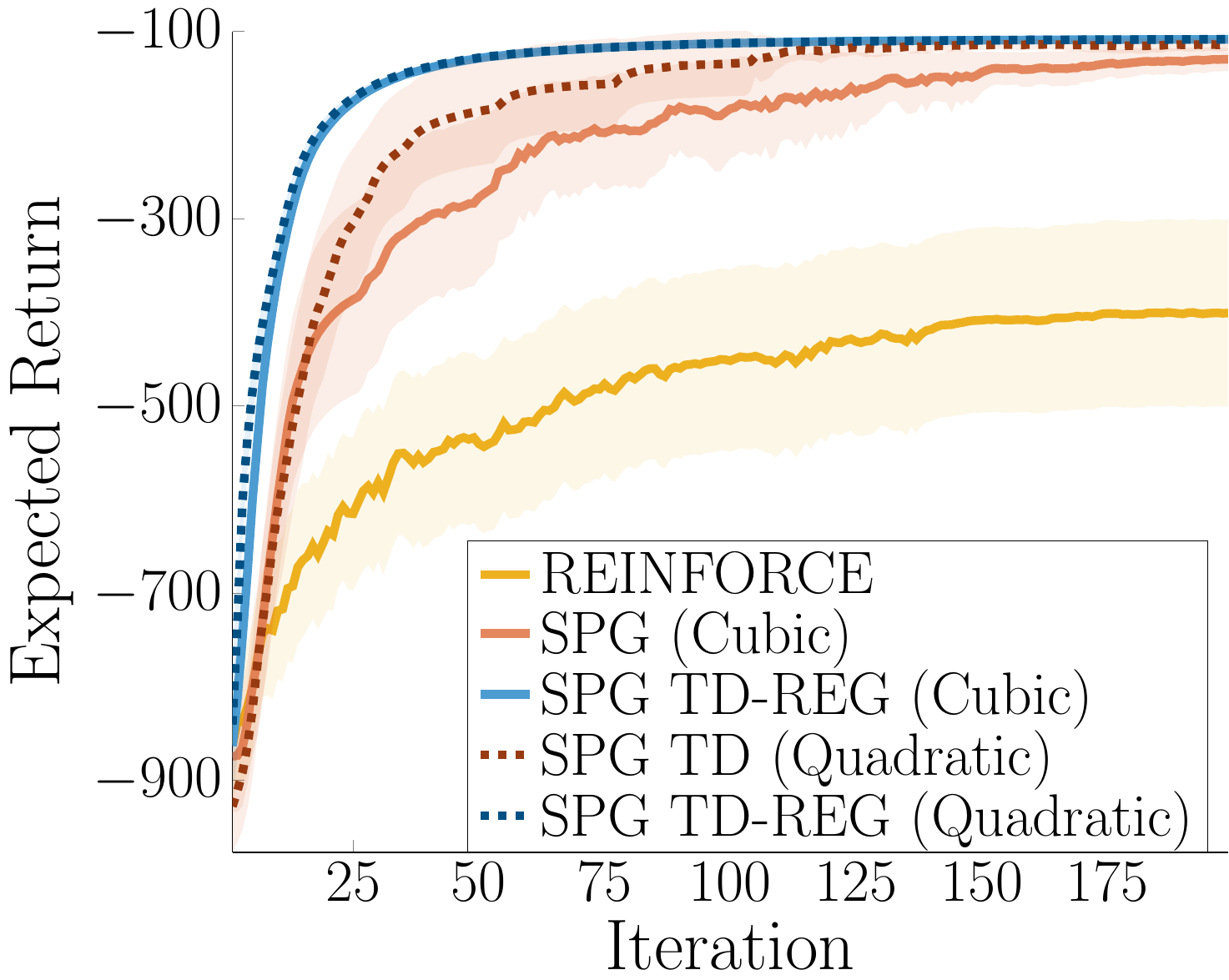}
			\caption{\label{fig:spg_app_1_ret}}
		\end{subfigure}
		\hfill
		\begin{subfigure}[t]{.325\linewidth}
			\centering
			\includegraphics[width=\textwidth]{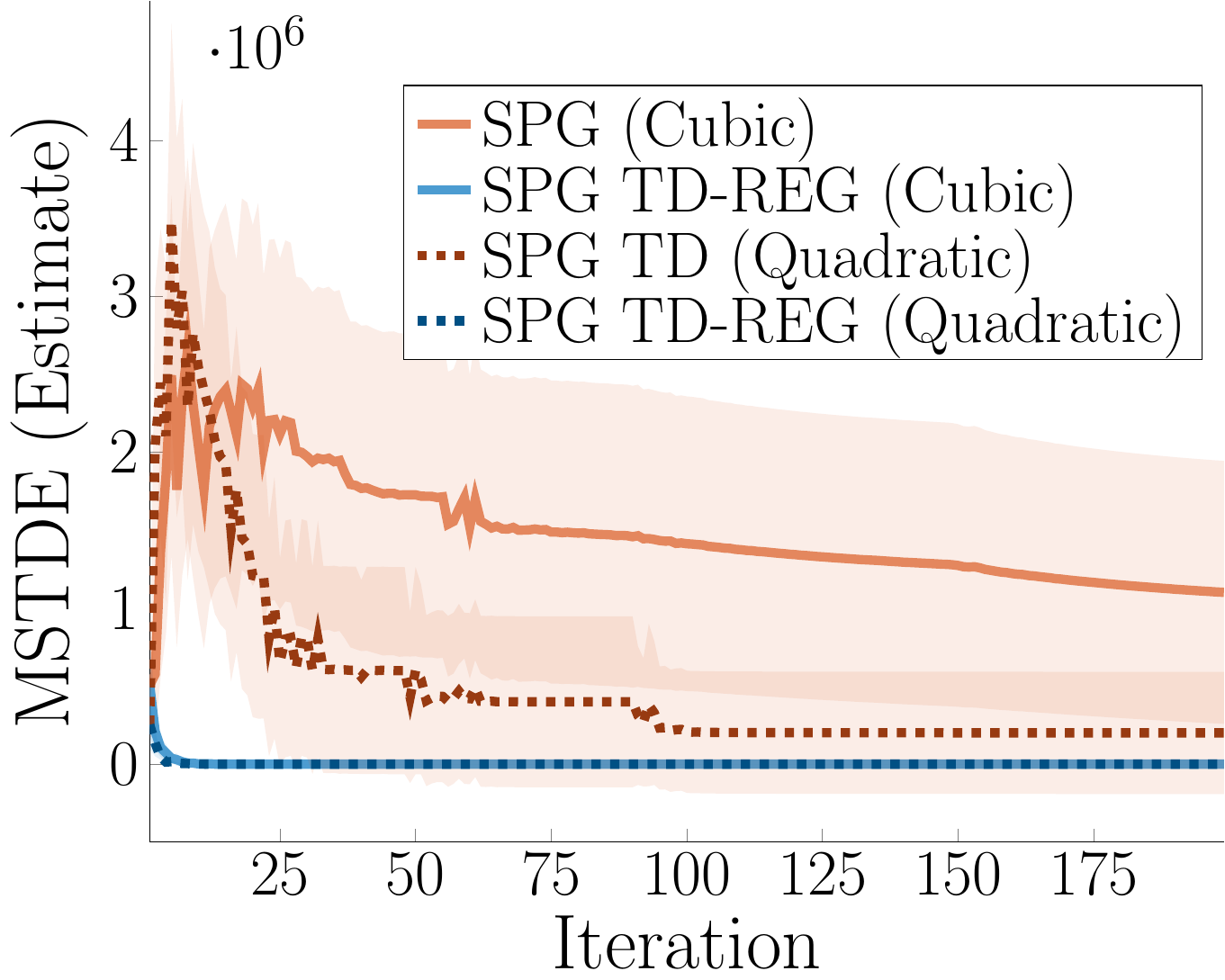}
			\caption{\label{fig:spg_app_1_td}}
		\end{subfigure}
		\hfill
		\begin{subfigure}[t]{.325\linewidth}
			\centering
			\includegraphics[width=\textwidth]{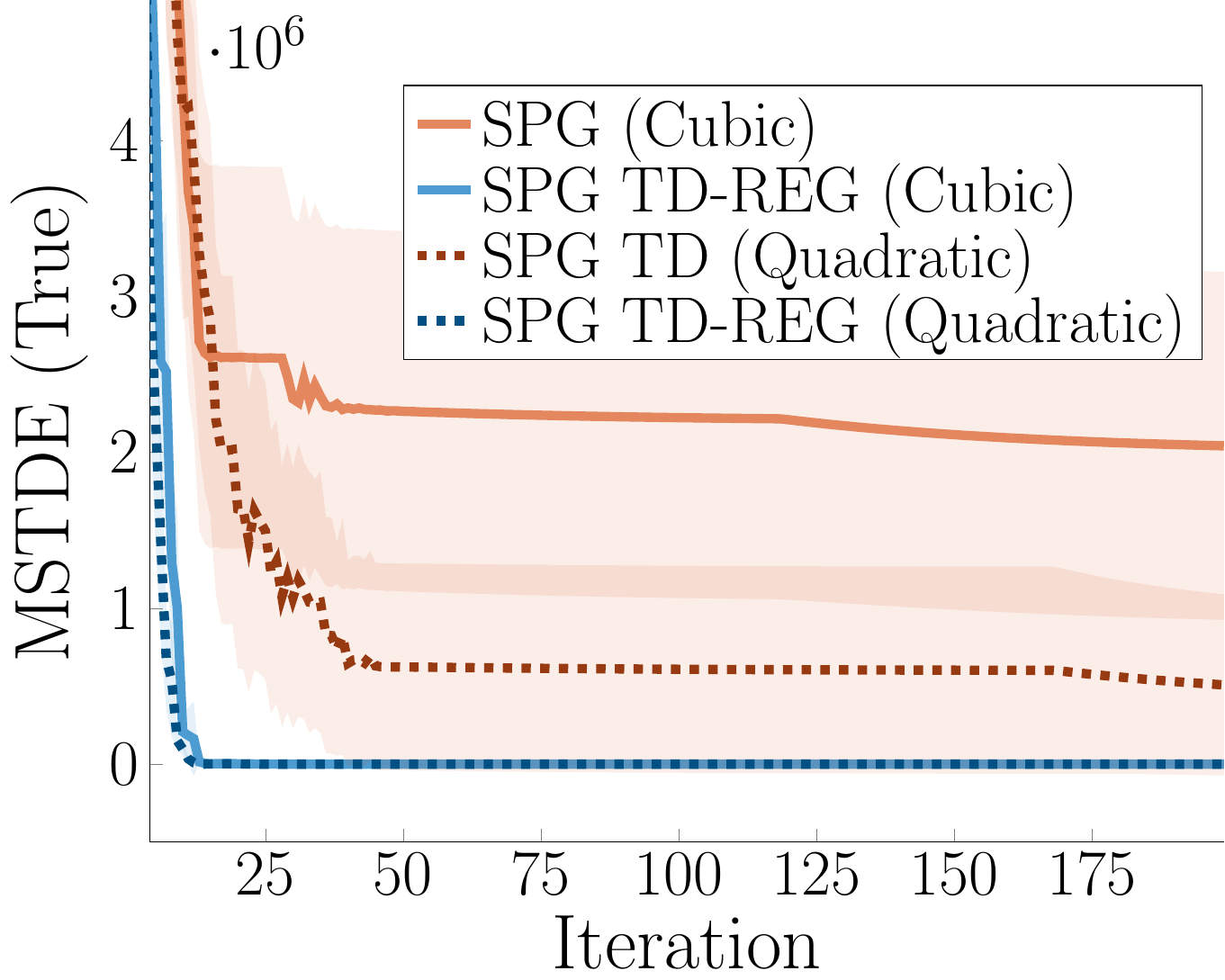}
			\caption{\label{fig:spg_app_1_tdtrue}}
		\end{subfigure}
		\caption{\label{fig:spg_app_1} Results averaged over 50 runs, shaded areas denote 95\% confidence interval. Only one episode is collected to update the critic and the policy. SPG TD-REG did not suffer from the lack of samples, and it always learned the optimal critic and policy within few iterations, both with quadratic and cubic features (the two blue plots almost overlap). By contrast, vanilla SPG is much more sensitive to the number of samples. Even if it never diverged, its convergence is clearly slower. REINFORCE, instead, diverged 13 times, due to the high variance of Monte Carlo estimates.}
	\end{minipage}
\end{figure}

\begin{figure}[h]
	\centering
	\begin{center}
		\bfseries SPG on LQR - Five Episodes per Iteration
	\end{center}
	\begin{minipage}[t]{\textwidth}
		\begin{subfigure}[t]{.325\linewidth}
			\centering
			\includegraphics[width=\textwidth]{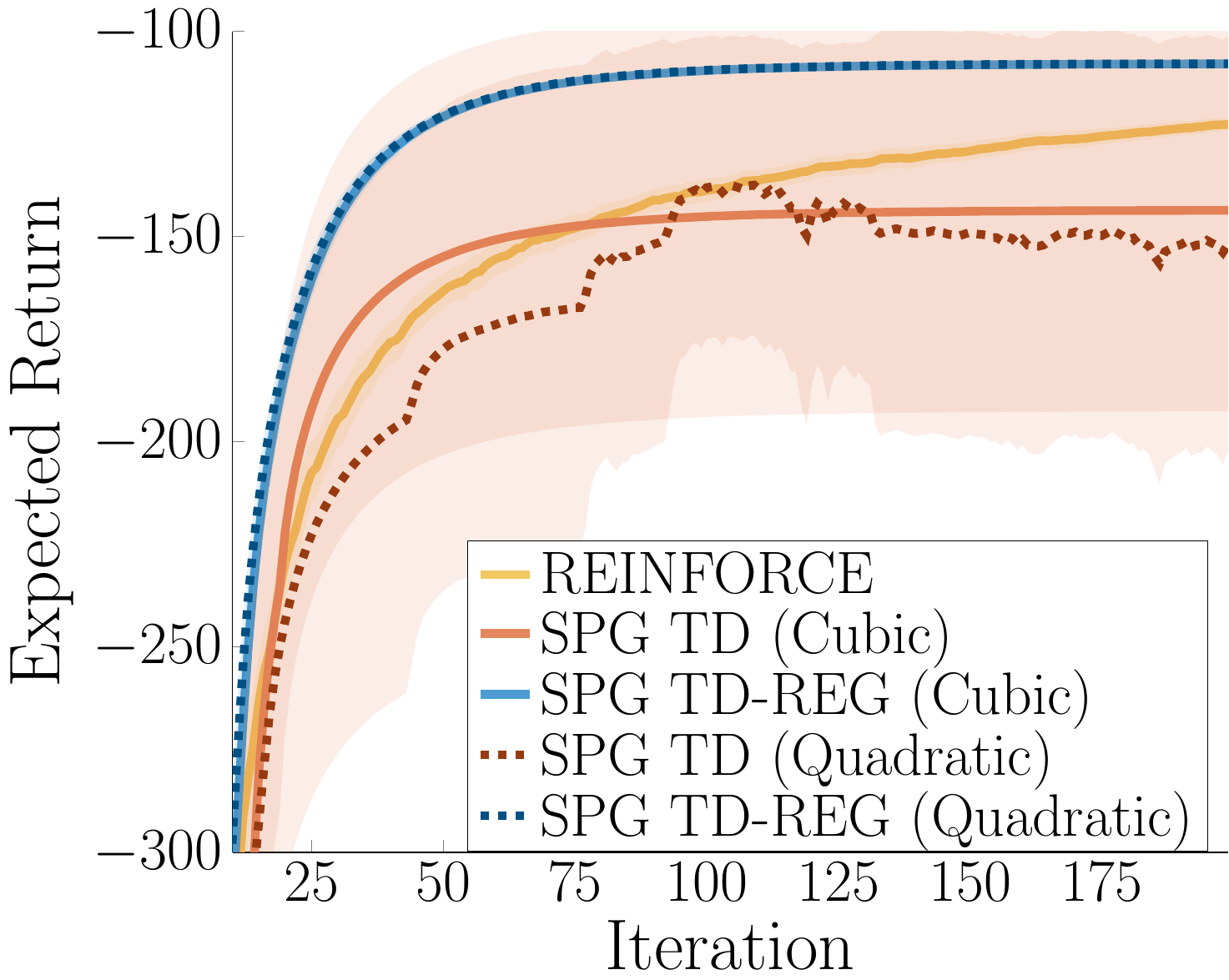}
			\caption{\label{fig:spg_app_5_ret}}
		\end{subfigure}
		\hfill
		\begin{subfigure}[t]{.325\linewidth}
			\centering
			\includegraphics[width=\textwidth]{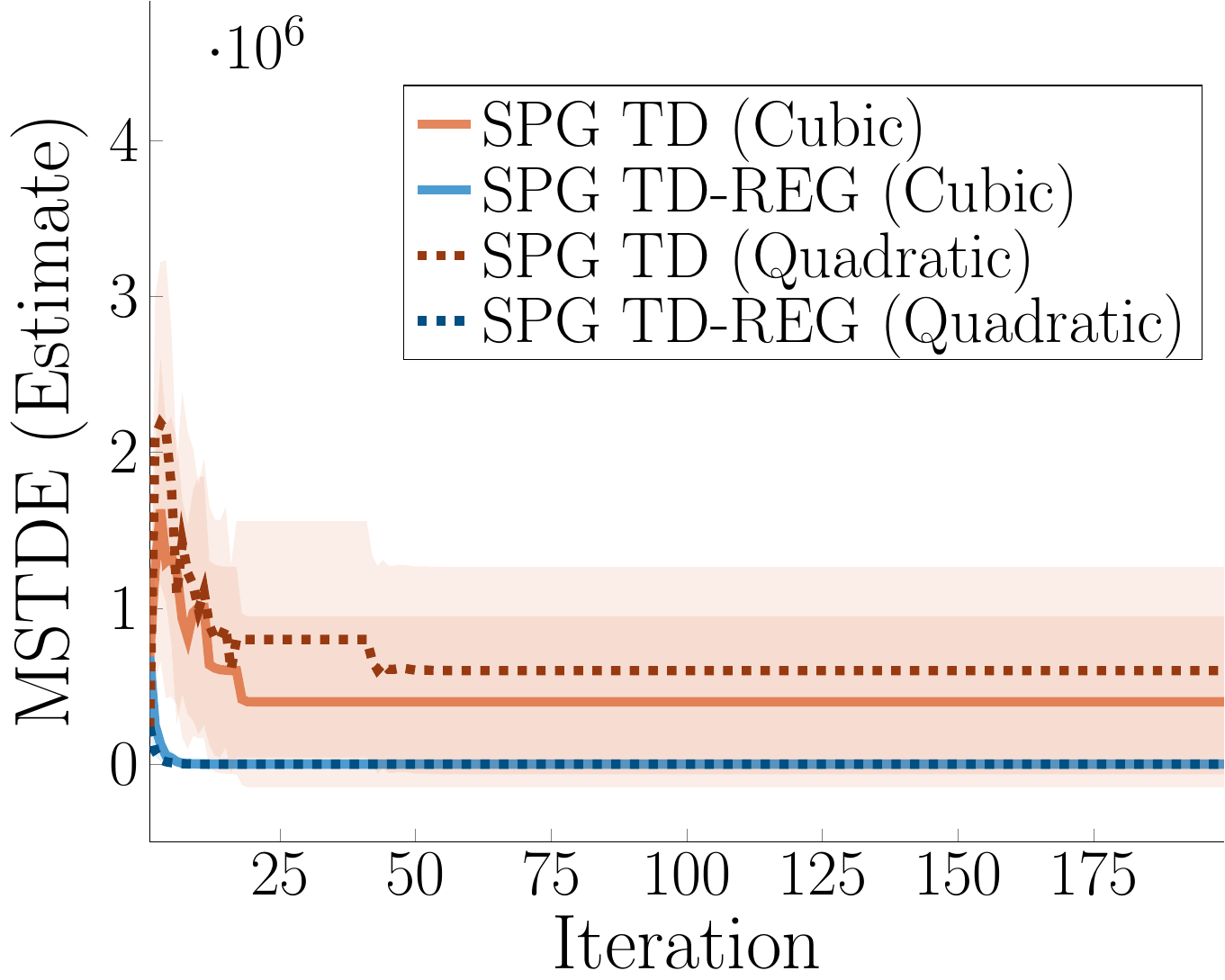}
			\caption{\label{fig:spg_app_5_td}}
		\end{subfigure}
		\hfill
		\begin{subfigure}[t]{.325\linewidth}
			\centering
			\includegraphics[width=\textwidth]{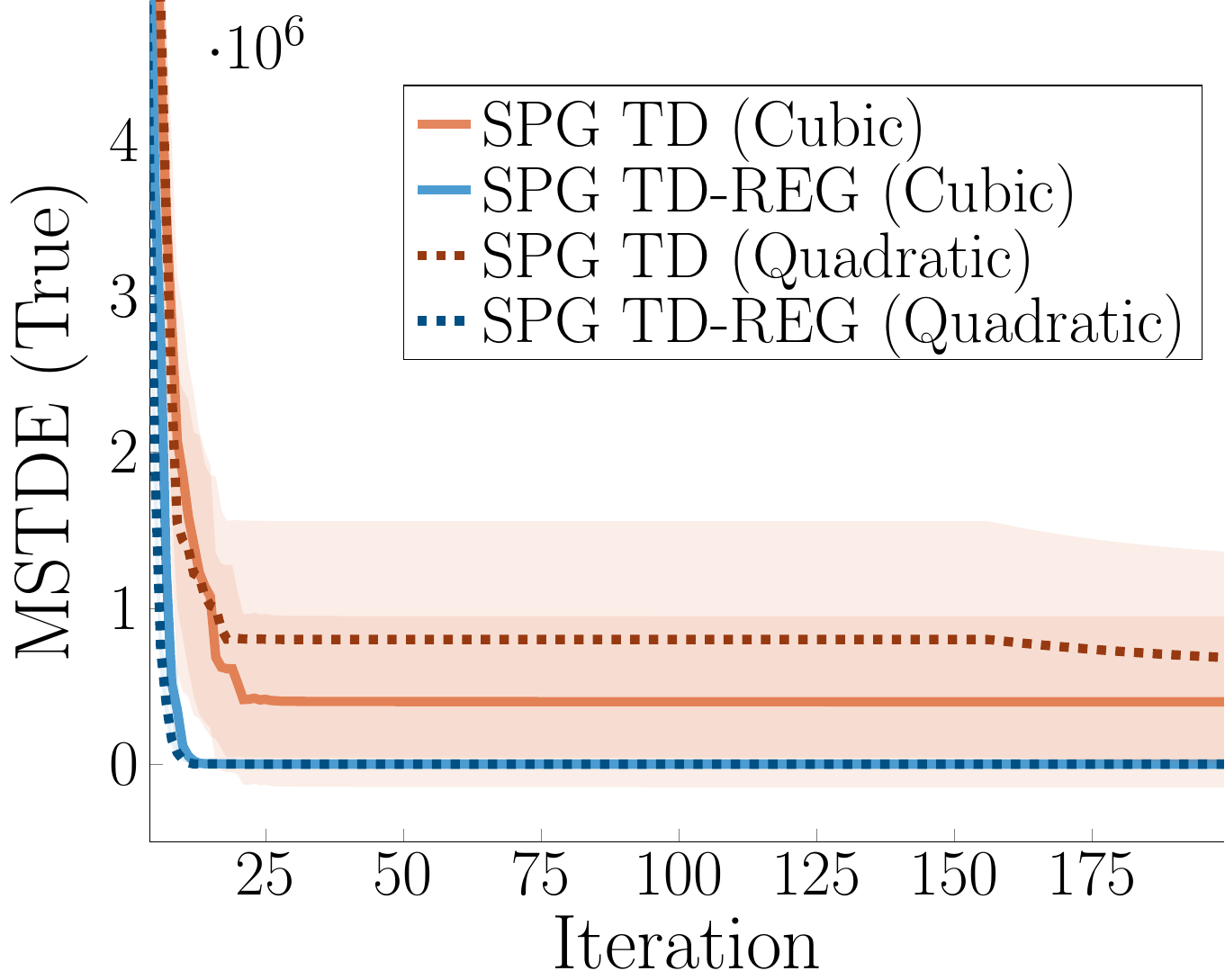}
			\caption{\label{fig:spg_app_t_tdtrue}}
		\end{subfigure}
		\caption{\label{fig:spg_app_5} With more samples per iteration REINFORCE was able to converge, as Monte Carlo estimates have lower variance. However, it still converged more slowly than SPG TD-REG. Surprisingly, vanilla SPG diverged two times, thus explaining its slightly larger confidence interval (in Figure~\ref{fig:spg_app_5_ret} the y-axis starts at -300 for the sake of clarity). In the other 48 runs, it almost matched SPG TD-REG.}
	\end{minipage}
	\\[1em]
	\includegraphics[width=\textwidth]{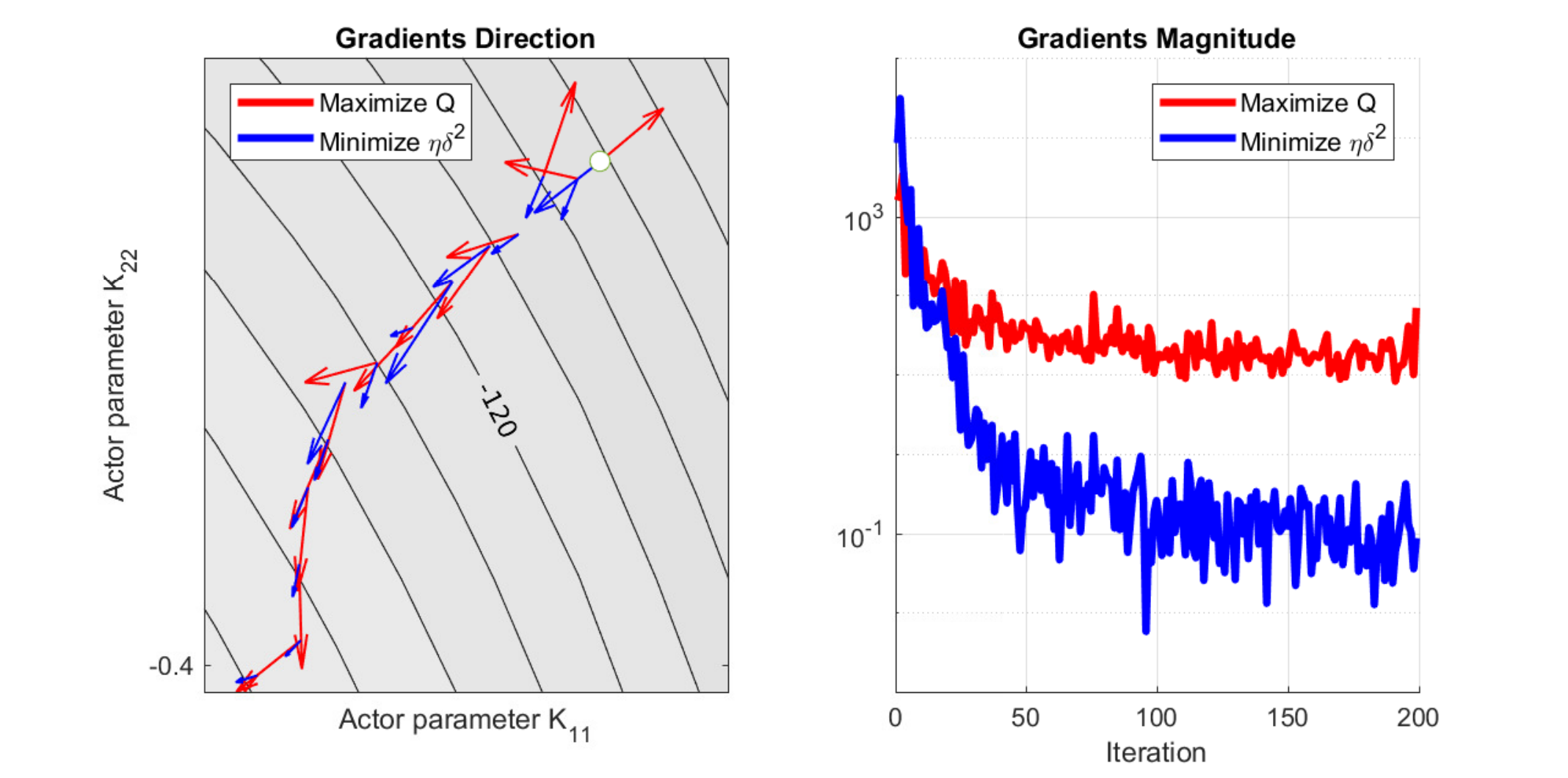}
	\caption{\label{fig:spg_gradients} On the left, beginning of the path followed by SPG TD-REG in the policy parameter space during one trial, when samples from only one episode are collected per iteration. The initial policy parameter vector $\params$ is denoted by the white circle. Each arrow represents one iteration. Red and blue arrows denote the direction of the gradient $\nabla_\params \EV[\widehat Q(\state,\action;\paramsQ)]$ and $\nabla_\params \EV[\tdcoeff\delta_Q(\state,\action,\state';\params,\paramsQ)^2]$, respectively. Contour denotes the expected return magnitude. The magnitude of the arrows has been scaled for better visualization, and the true one is shown on the right. Initially, as the critic estimate is highly inaccurate (see the TD error in Figures \ref{fig:spg_app_1_td} and \ref{fig:spg_app_1_tdtrue}), red arrows point to the wrong direction. However, blue arrows help keeping the policy on track.}
\end{figure}

\paragraph{Results\\}
Figures \ref{fig:spg_app_1} and \ref{fig:spg_app_5} show the results when one or five episodes, respectively, are used to collect samples during one learning iteration. REINFORCE performed poorly in the first case, due to the large variance of Monte Carlo estimates of the Q-function. In the second case, it performed better but still converged slowly. 
SPG performed well except for two runs in Figure~\ref{fig:spg_app_5}, in which it diverged already from the beginning. In both cases, its performance is coupled with the approximator used, with quadratic features yielding more stability. By contrast, SPG TD-REG never failed, regardless of the number of samples and of the function approximator used, and despite the wrong estimate of the true TD error. Similarly to what happened in DPG, in fact, the critic always underestimates the true TD error, as shown in Figures~\ref{fig:spg_app_1_td} and~\ref{fig:spg_app_5_td}.

Finally, Figure~\ref{fig:spg_gradients} shows the direction and the magnitude of the gradients. We clearly see that initially the vanilla gradient (which maximizes Q-function estimates, red arrows) points towards the wrong direction, but thanks to the gradient of the TD-regularization (blue arrows) the algorithm does not diverge. As the learning continues, red arrows point towards the optimum and the magnitude of blue arrows decreases, because 1) the critic becomes more accurate and the TD error decreases, and 2) the regularization coefficient $\tdcoeff$ decays. The same behavior was already seen in Figure~\ref{fig:lqr_paths} for DPG, with the penalty gradient dominating the vanilla gradient at early stages of the learning.

\clearpage

\section{Pendulum Swing-up Tasks Experiments}
\label{app:pendulum}
\begin{figure}[h]
	\centering
	\begin{minipage}[t]{\textwidth}
		\begin{subfigure}[t]{.49\linewidth}
			\centering
			\includegraphics[width=\linewidth]{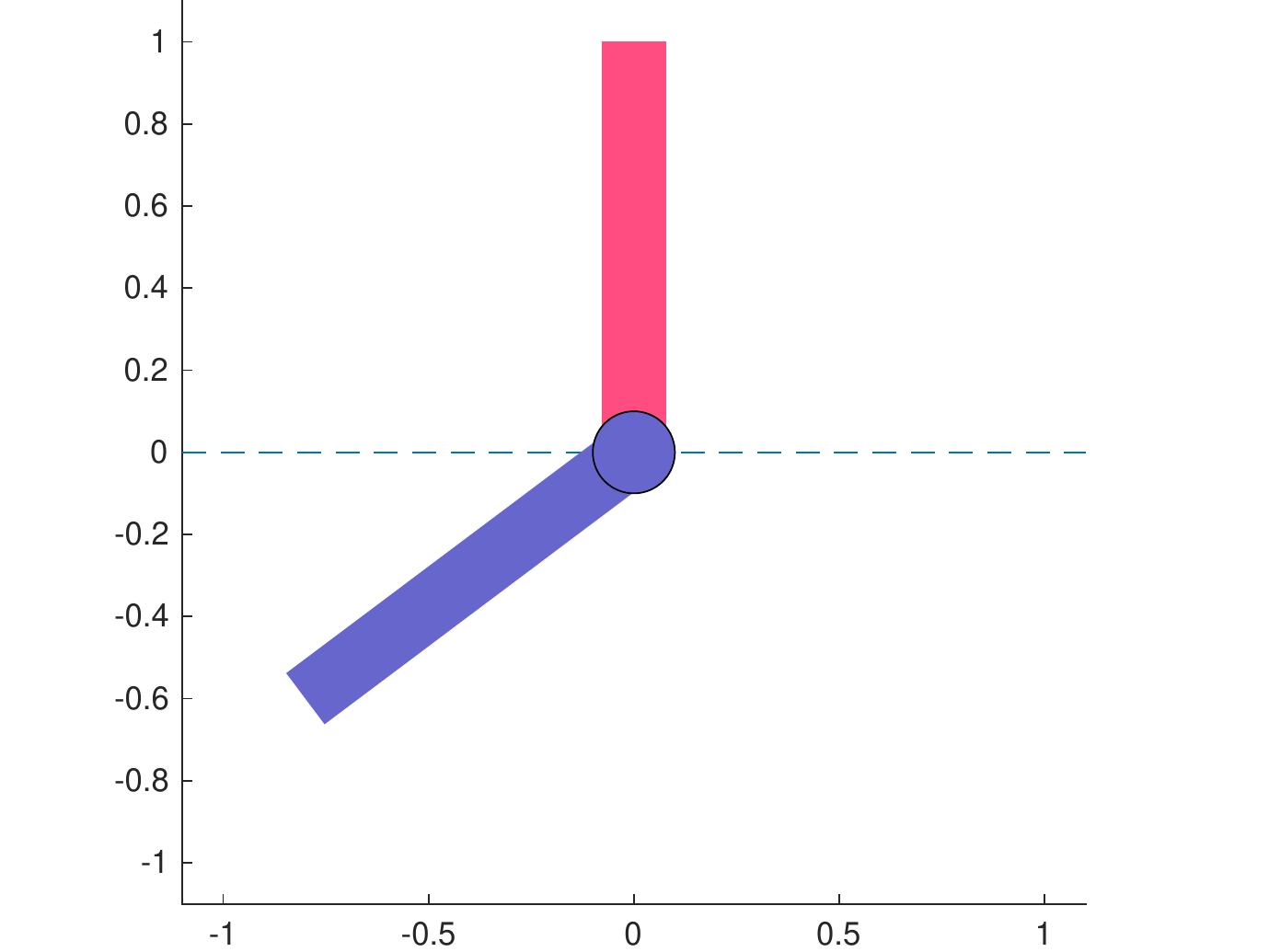}
			\caption{Single-pendulum.\label{fig:pend_env}}
		\end{subfigure}
		\hfill
		\begin{subfigure}[t]{.49\linewidth}
			\centering
			\includegraphics[width=\linewidth]{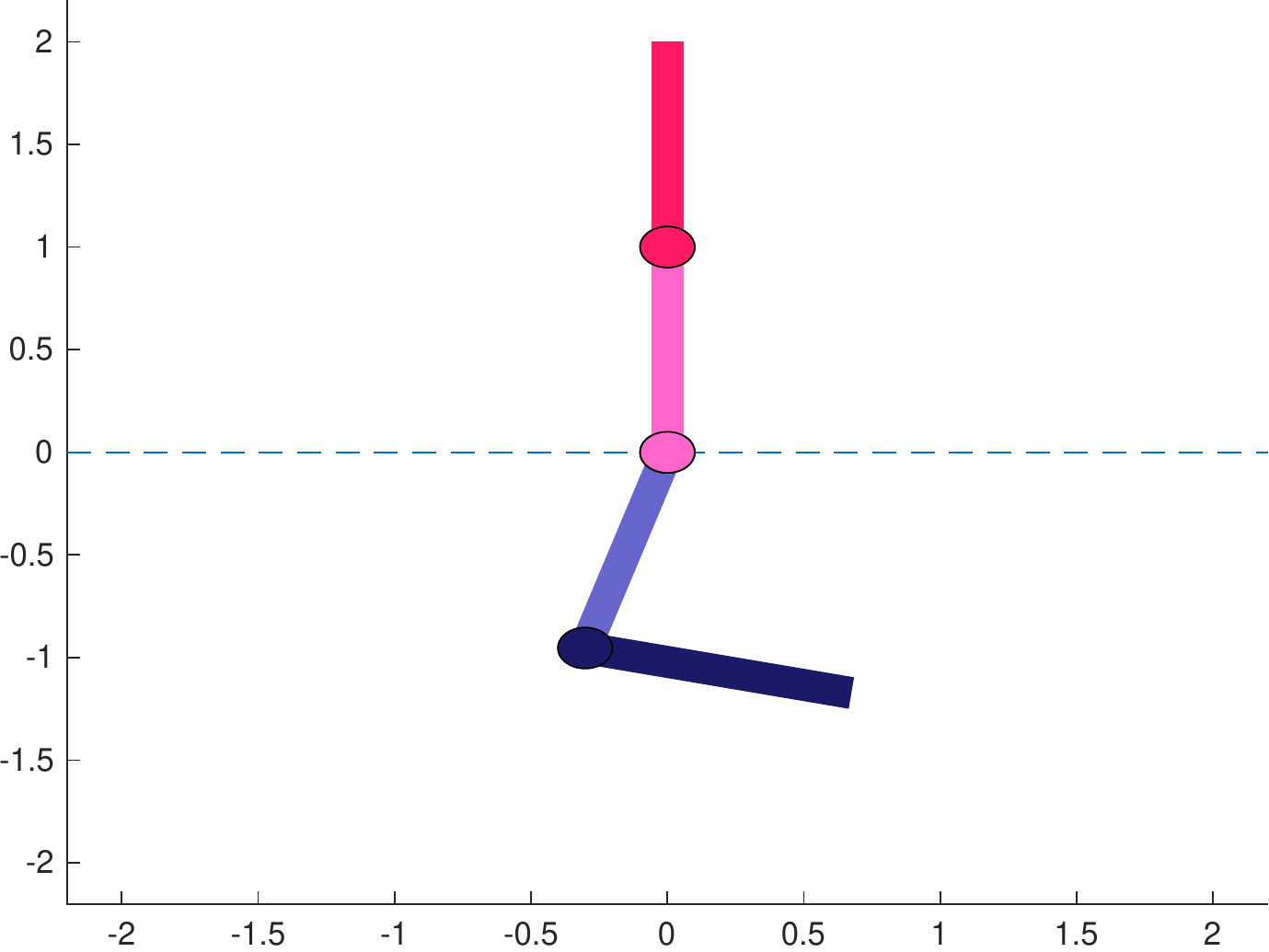}
			\caption{Double-pendulum.\label{fig:pend2_env}}
		\end{subfigure}
		\caption{\label{fig:pend_app_env}The two pendulum swing-up tasks, with an example of the current state (in blue) and the goal state (in red).}
	\end{minipage}
\end{figure}

\paragraph{Single-pendulum.} This task (Figure~\ref{fig:pend_env}) follows the equations presented in OpenAI Gym \citep{brockman2016openai}. The state consists of the angle position $q$ and velocity $\dot q$ of the link. The former is bounded in $[-\pi, \pi]$ and is initialized uniformly in $[-\pi, \pi]$. The latter is bounded in $[-8,8]$ and is initialized in $[-1,1]$. As the pendulum is underactuated, the action is bounded in $[-2,2]$. The transition function is
\begin{align*}
	\dot q_{t+1} &= \dot q_{t} - \frac{3g}{2l}\sin(q+\pi) + \frac{3}{ml^2}a\delta_t,
	\\
	q_{t+1} &= q_t  + \dot q_{t+1}\delta_t,
\end{align*}
where $g = 10$ is the gravitational acceleration, $m = 1$ is the link mass, $l = 1$ is the link length, and $\delta_t = 0.05$ is the timestep.
The goal state is $q = 0$ and the reward function is
\begin{equation*}
	r_t = -q_t^2 - 0.1\dot q_t^2 - 0.001 a_t^2.
\end{equation*}

For learning, at each iteration only 10 episodes of 50 steps are collected. For the expected return evaluation, the policy is tested over 1,000 episodes of 150 steps.
\\
The critic approximates the V-function with a linear function $\smash{\widehat{V}(\state;\paramsQ) = \phi(\state)^{\T}\paramsQ}$ where $\phi(\state)$ are 100 random Fourier features, as presented in~\citep{rajeswaran2017towards}. The bandwidths are computed as suggested by~\citep{rajeswaran2017towards} as the average pairwise distances between 10,000 state observation vectors. These states are collected only once before the learning and are shared across all 50 trials. The phase and the offset of the Fourier features are instead random and different for all trials.
\\
The policy is Gaussian, i.e., $\pi(\action|\state;\params) = \gaussian(b + K\phi(\state), \sigma^2)$. The same Fourier features of the critic are used for the policy, for a total of 102 policy parameters to learn. $K$ and $b$ are initialized to zero, while $\sigma$ to four. For the expected return evaluation, the noise of the policy is zeroed.
\\
The critic parameters are initialized uniformly in $[-1,1]$ and are learned with Matlab \texttt{fminunc} optimizer such that they minimize the mean squared TD($\elicoeff$) error (since we use GAE). GAE hyperparameters are $\gamma = 0.99$ and $\elicoeff = 0.95$.
\\
The policy is learned with TRPO with a KL bound of 0.01. For computing the natural gradient, both the advantage, the TD error and the regularization are standardized, i.e., $\hat y \leftarrow (\hat y - \mu_\textsub{y}) / \sigma_\textsub{y}$, where $\hat y$ is either the advantage estimator, the TD error estimator, or the regularization ($\delta^2$ or $\smash{\widehat{A}^2}$, depending if we use TD-REG or GAE-REG), $\mu_\textsub{y}$ is the mean of the estimator and $\sigma_\textsub{y}$ its standard deviation.
The conjugate gradient is computed with Matlab \texttt{pcg}. Additionally, since TRPO uses a quadratic approximation of the KL divergence, backtracking line search is performed to ensure that the KL bound is satisfied.
The starting regularization coefficient is $\tdcoeff_0 = 1$ and then decays according to $\tdcoeff_{t+1} = \kappa\tdcoeff_t$ with $\kappa = 0.999$.

\paragraph{Double-pendulum.} This task (Figure~\ref{fig:pend2_env}) follows the equations presented by \citet{yoshikawa1990foundations}. The state consists of the angle position $[q_1, q_2]$ and velocity $[\dot q_1, \dot q_2]$ of the links. Both angles are bounded in $[-\pi, \pi]$ and initialized uniformly in $[-\pi, \pi]$, while both velocities are bounded in $[-50,50]$ and initialized in $[-1,1]$. The agent, however, observes the six-dimensional vector $[\sin(q), \cos(q), \dot{q}]$. As the pendulum is underactuated, the action on each link is bounded in $[-10,10]$. The transition function is            
\begin{align*}
	\ddot{q}_{t+1} &= M_t^{-1} (a_t - f_{gt} - f_{ct} - f_{vt}),
	\\
	\dot q_{t+1} &= \dot q_{t} - \ddot{q}_{t+1}\delta_t,
	\\
	q_{t+1} &= q_t  + \dot q_{t+1}\delta_t,
\end{align*}
where $\delta_t = 0.02$ is the timestep, $M$ is the inertia matrix, $f_g$ is gravitational, $f_c$ the Coriolis force, and $f_v$ the frictional force. The entries of $M$ are
\begin{align*}
	M_{11} &= m_1\left(\frac{l_1}{2}\right)^2 + I_1 + m_2\left(l_1^2 + \left(\frac{l_2}{2}\right)^2 + 2l_1\left(\frac{l_2}{2}\right)^2 \cos(q_2)\right) + I_2,
	\\
	M_{12} = M_{21} &= m_2 \left(\left(\frac{l_2}{2}\right)^2 + l_1 \frac{l_2}{2} \cos(q_2)\right) + I_2,
	\\
	M_{22} &= m_2 \left(\frac{l_2}{2}\right)^2 + I_2,
\end{align*}
where $l_i = 1$ is the length of a link, $m_i = 1$ is the mass of the a link, and $I_i = (1 + 0.0001) / 3.0$ is the moment of inertia of a link. Gravitational forces are
\begin{align*}
	f_{g1} &= m_1 g \frac{l_1}{2} \cos(q_1) + m_2 g \left(l_1 \cos(q_1) + \frac{l_2}{2} \cos(q_1+q_2)\right),
	\\
	f_{g2} &= m_2 g \frac{l_2}{2} \cos(q_1+q_2),
\end{align*}
where $g = 9.81$ is the gravitational acceleration.
Coriolis forces are
\begin{align*}
	f_{c1} &= -m_2 l_1 \frac{l_2}{2} \sin(q_2) (2 \dot{q}_1 \dot{q}_2 + \dot{q}_2^2),
	\\
	f_{c2} &= m_2 l_1 \frac{l_2}{2} \sin(q_2) \dot{q}_1^2.
\end{align*}
Frictional forces are
\begin{align*}
	f_{v1} &= v_1 \dot{q}_1,
	\\
	f_{v2} &= v_1 \dot{q}_2,
\end{align*}
where $v_i = 2.5$ is the viscous friction coefficient.
The goal state is $q = [\pi/2, 0]$ and the reward function is
\begin{equation*}
	r_t = -||\pi - \texttt{abs}(\texttt{abs}(q-q_\textsub{goal}) - \pi)||_2^2 - 0.001 ||a||_2^2,
\end{equation*}
where the first term is the squared distance between the current angles and the goal position $q_\textsub{goal} = [\pi/2, 0]$, wrapped in $[-\pi,\pi]$.
Note that, compared to the single-pendulum, the frame of reference is rotated of $90^\circ$. The first angle $\q_1$ is the angle between the base of the pendulum and the first link. The second angle $q_2$ is the angle between the two links.

For learning, at each iteration only 6 episodes of 500 steps are collected, because the double-pendulum needs more steps to swing up than the single-pendulum. For the expected return evaluation, the policy is tested over 1,000 episodes of 500 steps. 
The same actor-critic setup of the single-pendulum is used, except for 
\begin{itemize}
	\item The Fourier features, which are 300,
	\item The policy, which has a full covariance matrix. Its diagonal entries are initialized to 200 and we learn its Cholesky decomposition, for a total of 605 parameters to learn, and
	\item The initial regularization coefficient for TD-REG, which is $\tdcoeff_0 = 0.1$.
\end{itemize}

\clearpage

\section{MuJoCo Continuous Control Tasks Experiments}
\label{app:mujoco}
The continuous control experiments are performed using OpenAI Gym~\citep{brockman2016openai} with MuJoCo physics simulator. All environments are version 2 (v2).
We use the same hyperparameters and neural network architecture for all tasks. 
The policy is a Gaussian distribution with a state-independent diagonal covariance matrix. 
Each algorithm is evaluated over five trials using different random seeds with common seeds for both methods.
For TRPO, the policy was evaluated over 20 episodes without exploration noise. For PPO, we used the same samples collected during learning, i.e., including exploration noise.

The actor mean and the critic networks are two-layer neural networks with hyperbolic tangent activation function. Because all tasks actions are bounded in $[-1,1]$, the actor network output has an additional hyperbolic tangent activation.

At each iteration, we collect trajectories until there are at least 3,000 transition samples in the training batch. The maximum trajectory length is 1,000 steps.
Then, we compute the GAE estimator as in Eq.~\eqref{eq:gae_eq_tdl} and train the critic to minimize the mean squared TD($\elicoeff$) error (Eq.~\eqref{eq:tde_lambda}) using ADAM~\citep{kingma2014adam}\footnote{TRPO original paper proposes a more sophisticated method to solve the regression problem. However, we empirically observed that batch gradient descent is sufficient for good performance.}.
The actor is then trained using the same training batch, following the algorithm policy update (PPO, TRPO, and their respective regularized versions). For the policy update, both the advantage, the TD error and the regularization are standardized, i.e., $\hat y \leftarrow (\hat y - \mu_\textsub{y}) / \sigma_\textsub{y}$, where $\hat y$ is either the advantage estimator, the TD error estimator, or the regularization ($\delta^2$ or $\smash{\widehat{A}^2}$, depending if we use TD-REG or GAE-REG), $\mu_\textsub{y}$ is the mean of the estimator and $\sigma_\textsub{y}$ its standard deviation.

\paragraph{TRPO Hyperparameters}
\begin{itemize}
	\setlength{\itemsep}{0em}
	\item Hidden units per layer 128.
	\item Policy initial standard deviation 1.
	\item GAE hyperparameters: $\elicoeff = 0.97$ and $\gamma = 0.995$
	\item ADAM hyperparameters: 20 epochs, learning rate $0.0003$, mini-batch size 128.
	\item Policy KL divergence bound $\epsilon = 0.01$. 
	\item We also add a damping factor $0.1$ to the diagonal entries of the Fisher information matrix for numerical stability
	\item The maximum conjugate gradient step is set to 10.
\end{itemize}

\paragraph{PPO Hyperparameters}
\begin{itemize}
	\setlength{\itemsep}{0em}
	\item Hidden units per layer 64.
	\item Policy initial standard deviation 2.
	\item GAE hyperparameters: $\elicoeff = 0.95$ and $\gamma = 0.99$
	\item ADAM hyperparameters (for both the actor and the critic): 20 epochs, learning rate $0.0001$, mini-batch size 64.
	\item Policy clipping value $\varepsilon = 0.05$.
\end{itemize}

\end{document}